\documentclass{article}
\usepackage{iclr2027_conference,times}

\usepackage{amsmath,amssymb}
\usepackage{graphicx}
\usepackage{booktabs}
\usepackage{float}
\usepackage{array}
\usepackage{multirow}
\usepackage{fontawesome5}
\usepackage{microtype}   
\usepackage{siunitx}     
\usepackage{subcaption}  
\usepackage[shortlabels]{enumitem}
\DeclareTextCommand{\k}{OT1}[1]{{\fontencoding{T1}\selectfont\k{#1}}}
\usepackage{xcolor}
\usepackage{url}
\usepackage{hyperref}
\usepackage{tabularx}
\usepackage{placeins}
\usepackage{etoc}        
\definecolor{linkblue}{rgb}{0.08,0.12,0.45}
\definecolor{gray50}{gray}{0.50}
\definecolor{bcol}{RGB}{74,128,228}

\newcommand{\srcfig}[1]{%
 \includegraphics[width=0.88in,height=0.88in,keepaspectratio]
 {example-fig-mol/#1}}

\newcommand{\gtfig}[1]{%
\includegraphics[width=0.88in,height=0.88in,keepaspectratio]
 {molfigs/#1.pdf}}

\newcommand{\ourfig}[1]{%
    \includegraphics[width=0.88in,height=0.88in,keepaspectratio]
    {molfigs/#1.pdf}}
\hypersetup{colorlinks=true,linkcolor=linkblue,citecolor=linkblue,urlcolor=linkblue}  

\newcommand{\todo}[2][]{\textcolor{red}{\footnotesize~[TODO: #2]}}
\newcommand{\figbench}{\textsc{ChemRxn-Bench}}
\newcommand{\agentname}{\textsc{RxnMiner}}   
\newcommand{\corpusfigures}{10{,}107}
\newcommand{\corpusreactions}{88{,}202}

\iclrfinalcopy
\renewcommand{\todo}[2][]{}

\title{\agentname{}: A Multimodal Agent for Structured Reaction Data
Extraction from Chemistry Figures}

\author{%
Jinbo Hu$^{1,2,*}$, Songren Zhang$^{1,2,*}$, Chuanxun Fu$^{1,2,*}$, Jinrui Huang$^{1,2,*}$, Zhen Hao Wong$^{1,2,*}$,\\
\bf Dongxu Cui$^{1,*}$, Haoran Fan$^{1,*}$, Muhao Ge$^{1}$, Zhen Jian$^{1}$, Jian Zhou$^{1}$, Chengxiang Luo$^{1}$,\\
\bf Han Lyu$^{1}$, Haoyi Tao$^{1}$, Chaozheng Huang$^{1}$, Hanzheng Li$^{1}$, ZiChen Wang$^{1,2}$, Hengtian Ma$^{1,3}$,\\
\bf Jinyang Dai$^{1,4}$, Nan Wang$^{1,\dagger}$, Yanhui Hong$^{1}$, Xi Fang$^{1,\dagger,\ddagger}$, Guolin Ke$^{1,\ddagger}$\\[4pt]
\normalfont $^{1}$DP Technology \quad $^{2}$Peking University \quad
$^{3}$The Hong Kong University of Science and Technology\\
\normalfont $^{4}$School of Mathematical Sciences, Peking University\\
\normalfont $^{*}$Equal contribution. \quad $^{\dagger}$Project lead. \quad $^{\ddagger}$Corresponding authors: \texttt{fangxi@dp.tech}.
}

\begin{document}
\maketitle
\lhead{Preprint. Under review.}

\begin{abstract}
Reaction data underpin machine learning for reaction prediction,
condition recommendation, and retrosynthetic planning, yet extracting
these data from literature figures still requires substantial manual
effort. Many figures encode multiple experiments through shared
templates, substituent tables, and product grids. Recovering concrete
reaction records therefore requires molecular \emph{recognition},
\emph{association} of evidence with individual experimental entries,
and \emph{reconstruction} of indirectly specified participants.
We present \agentname{}, a multimodal agent system that combines
an expert-informed extraction procedure with a chemistry toolbox to
coordinate these operations. The agent reads molecular structures,
resolves shared and entry-specific evidence, and instantiates
participants from templates to assemble source-linked reaction records.
We also introduce \figbench{}, a benchmark of 603 reaction figures
with 8,547 manually annotated and reviewed reaction records, evaluated
under a common protocol for exact molecular identity and
reactant--product pairing. On this benchmark, with Qwen3.8-Flash-Next as the backbone, \agentname{}
raises stereochemistry-aware reaction macro-F1 from 21.0\% under direct
prediction to 71.0\%, compared with 25.0\% for the strongest existing
extraction system. Across five backbones, it also outperforms the same
base agent without the extraction procedure and chemistry toolbox by
9.4 to 25.5 points.
Applied without human intervention to 10,107 literature figures,
\agentname{} produces a silver corpus of 88,202 source-linked reaction
records.

\vspace{6pt}
\noindent{\small
\faHome\ \textbf{Homepage:} \href{https://rxnminer.github.io/RxnMiner/}{\texttt{rxnminer.github.io/RxnMiner}}\\[2pt]
\faGithub\ \textbf{Code:} \href{https://github.com/rxnminer/RxnMiner}{\texttt{github.com/rxnminer/RxnMiner}}\\[2pt]
\faDatabase\ \textbf{Benchmark:} \href{https://huggingface.co/datasets/UniParser/RxnMiner-Benchmark}{\texttt{hf.co/datasets/UniParser/RxnMiner-Benchmark}}\\[2pt]
\faDatabase\ \textbf{Dataset:} \href{https://huggingface.co/datasets/UniParser/RxnMiner-Dataset}{\texttt{hf.co/datasets/UniParser/RxnMiner-Dataset}}
}
\end{abstract}


\section{Introduction}
\label{sec:intro}
Chemical reaction data underpin data-driven chemistry. Models for reaction outcome prediction, condition recommendation, and retrosynthetic planning are all learned from records of reactions that chemists have already carried out~\citep{segler2018planning,schwaller2019molecular,gao2018conditions,coley2018casp}. Many of these records, however, remain embedded in the chemical literature, written for human readers rather than machines. Text mining of patents has yielded large reaction collections~\citep{lowe2017uspto}, and open repositories gather contributed experimental records~\citep{kearnes2021ord}, but extracting reactions from journal figures still requires substantial manual effort. The cost of that transcription is visible in AbSynth~\citep{czuba2026absynth}, which turns nearly a century of total syntheses into a machine-readable resource: its curation relied on the broad engagement of undergraduate students, who over several years assembled more than 3,000 total syntheses spanning close to 60,000 individual steps.

Such curated resources support both reaction modeling and the study of synthesis. Multi-step resources such as AbSynth make it possible to analyze the strategic logic of human synthesis~\citep{czuba2026absynth}, and expert-encoded reaction knowledge has enabled computers to plan syntheses of complex natural products~\citep{szymkuc2016planning,mikulak2020natural}. Building these resources depends in part on recovering the individual reactions reported in the literature. Automating this step would reduce the manual curation bottleneck and expand the data available for reaction prediction, retrosynthesis, and condition recommendation. Doing so requires recovering a concrete reaction record for each experimental entry represented in a figure: fully instantiated reactant and product molecular structures, correctly paired according to the depicted transformation and linked to the available experimental information.

\begin{figure}[t]
\centering
\includegraphics[width=\linewidth]{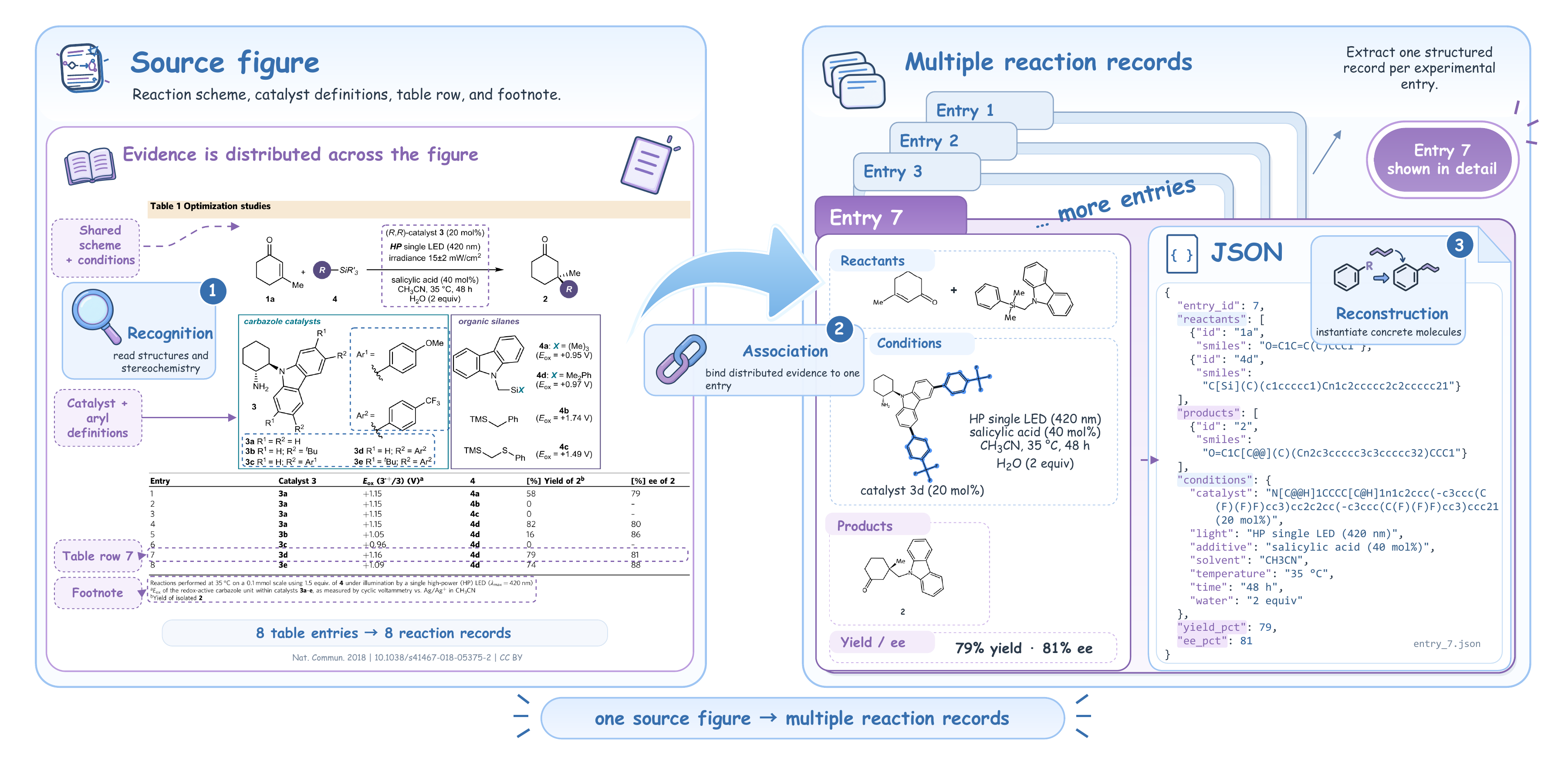}
\caption{\textbf{From compositional evidence to concrete reaction records.} An optimization table specifies eight experimental entries through a shared reaction scheme, catalyst definitions, table rows, and footnotes. The numbered steps indicate recognition, association, and reconstruction. Entry~7 is expanded to illustrate the target record, including its molecular participants, conditions, outcomes, and JSON representation.
}
\label{fig:compositional}
\end{figure}

Many reaction figures are \emph{compositional}: several
experimental entries share information that is drawn or stated
only once. In Figure~\ref{fig:compositional}, a shared reaction scheme, catalyst definitions, table rows, and footnotes jointly specify multiple experimental entries. Recovering a concrete reaction record for each entry therefore requires combining evidence across the figure; a collection of correctly recognized molecules alone does not fully specify that record. These dependencies give rise to three requirements.
\emph{Recognition} recovers molecular structures and printed text. \emph{Association} links these elements to the experimental entries they describe, resolving shared information and entry-specific assignments. \emph{Reconstruction} instantiates concrete molecules from templates, substituents, and the depicted transformation. Recognition supplies the visible components, while association and reconstruction determine how those components specify concrete reactions. Evaluation must therefore assess molecular identity and reactant--product pairing at the level of individual experimental entries.

Existing methods address these requirements to varying degrees. Molecular recognition methods recover individual structures~\citep{molscribe2023,molparser}, while reaction-diagram parsers and integrated extraction systems also recover roles, associate information, or process shared templates~\citep{rxnscribe2023,rxncaption2026,openchemie2024,chemeagle2024}. However, their reported evaluations differ in data collections, output targets, and matching rules. These differences make it difficult to compare how reliably systems recover the full set of concrete reaction records represented in a figure. Together, these considerations motivate a coordinated extraction approach and a common protocol for evaluating the recovered reaction records.

To address these compositional demands, we introduce \agentname{}, an expert-informed multimodal agent that coordinates molecular recognition, evidence association, and template instantiation. Its read-first workflow grounds drawn structures in optical recognition outputs, uses multimodal reasoning to associate evidence with experimental entries, and derives undrawn reactants with a deterministic R-group engine. Explicit provenance distinguishes structures read from the figure, derived from templates, or written by the model, supporting inspection of the resulting records against the source.

Our contributions are as follows:
\begin{enumerate}[leftmargin=*,itemsep=2pt]
\item \textbf{\agentname{}}, a multimodal agent that combines expert-informed instructions with chemical tools for reaction extraction. Entry-specific plans coordinate read-first molecular recognition, evidence association, and deterministic template instantiation, producing source-linked records with explicit structure provenance (Section~\ref{sec:agent}). With a locally deployed Qwen3.8-Flash-Next backbone, it achieves a
stereochemistry-aware reaction F1 of 71.0\%, compared with 21.0\% for the same model under vanilla prediction (Section~\ref{sec:results}).
\item \textbf{\figbench{}}, a benchmark of 603 reaction figures with 8,547 manually annotated and expert-reviewed concrete reaction records. It defines a common figure-level extraction task and ranks systems by macro-averaged reaction F1, requiring exact agreement on reactant and product molecular identities, including stereochemistry, within each matched record. Conditions and outcomes are assessed separately by a multimodal judge (Section~\ref{sec:benchmark}).
\item \textbf{A silver reaction corpus} of 88,202 source-linked reaction records extracted without human intervention from 10,107 figures, containing 3.6 times as many concrete reaction records as the original annotations of the same figures (Section~\ref{sec:corpus}).
\item \textbf{Open-source resources.}
We release the code for \agentname{} through
\href{https://github.com/rxnminer/RxnMiner/tree/main}{GitHub}
and our \href{https://rxnminer.github.io/RxnMiner/}{project website},
\figbench{} on Hugging Face (\href{https://huggingface.co/datasets/UniParser/RxnMiner-Benchmark}{RxnMiner-Benchmark}) with access granted on request, and the silver
reaction corpus openly on Hugging Face (\href{https://huggingface.co/datasets/UniParser/RxnMiner-Dataset}{RxnMiner-Dataset}), to support reproducible evaluation and further research.
\end{enumerate}

\section{Related Work}
\label{sec:related}

\paragraph{Optical chemical structure recognition.}
Optical chemical structure recognition (OCSR) converts a drawn molecule into a machine-readable structure. Rule-based and deep OCSR readers recover structures from literature and patent images~\citep{osra2009,rajan2020ocsrreview,decimer2021,img2mol2021,swinocsr2022,molscribe2023,molgrapher2023,molnextr2024,molparser}; page-level and Markush-aware extensions locate structures at document scale and resolve variable groups~\citep{decimerai2023,chemscraper2024,moleculeminer2025,markushgrapher2025,markushgrapher2_2026,patcid2024,rgreco2026}. These tiers identify molecules but do not determine reaction membership, roles, or conditions. \agentname{} therefore uses OCSR within a broader procedure that binds entries and reconstructs their reaction participants; detailed reader capabilities are deferred to the appendix.

\paragraph{Reaction extraction from chemical literature.}
Existing systems span several capability tiers. Diagram parsers recover reaction regions and scheme layout~\citep{rxnscribe2023,rde2023}; figure-level multimodal models add role, condition, and identifier prediction~\citep{rxnim2025,rxncaption2026,idtvp2026}; and document-level or agentic systems further combine structure recognition with text, template, and R-group reasoning~\citep{mermaid2025,openchemie2024,chemeagle2024,mineruchem2026}. Some recent systems can resolve R-groups or reconstruct concrete reaction participants from shared templates, but these capabilities are evaluated under heterogeneous datasets, schemas, and matching rules, and are not consistently isolated under a common reaction-level exact-match protocol. General-purpose vision--language models can also extract reactions directly. In contrast, \agentname{} binds evidence per experimental entry, instantiates shared templates, derives undrawn reactants deterministically, and preserves stereochemical evidence, yielding one concrete reactant--product record per entry. Section~\ref{sec:results} compares these systems and direct models under a common protocol; detailed capability differences are deferred to Appendix~\ref{app:related-work}.

\paragraph{Benchmarks for reaction understanding.}
RxnBench and ReactBench evaluate selected aspects of reaction-figure understanding through questions~\citep{rxnbench2026,reactbench2026}, while extraction systems generally use task-specific datasets, schemas, and matching rules. \figbench{} instead evaluates concrete reactant--product pairings across shared templates, product-only scopes, optimization tables, and multi-step schemes, making exhaustive reaction recovery the prediction target.

\section{\figbench{}}
\label{sec:benchmark}
\figbench{} connects the compositional requirements of Section~\ref{sec:intro} to a common extraction target. Its three components are a figure-level task, expert-reviewed records, and a scoring protocol that evaluates molecular identity within each recovered reaction.

\subsection{Task and Record Definition}
\label{sec:task}
We define reaction extraction as recovering the concrete reaction records represented in a reaction figure by identifying the molecular structures of reactants and products and their reaction-level associations.

\paragraph{Input Scope.}
The input is a manually checked image crop containing a reaction figure, its caption, footnotes, and associated table text. Figures identified during annotation as lacking sufficient information to recover the reference reactions from the provided crop are excluded. All primary systems receive the same crop, without access to the article body or supporting information.

\paragraph{Output representation.}

Each concrete reaction record specifies fully instantiated reactant and product molecular structures, encoded as  Extended SMILES (E-SMILES) in MolParser~\cite{molparser}, with no unresolved R-group placeholders. Shared templates are instantiated for individual experimental entries, and multi-step schemes are split into separate reaction records. Distinct experimental entries remain separate even when their molecular participants are identical. Records also retain conditions, outcomes, and provenance. The primary evaluation measures molecular identity and reactant--product pairing; conditions and outcomes are evaluated separately by the secondary judge described in Section~\ref{sec:protocol}. The full record schema and annotation conventions are provided in Appendix~\ref{app:records}.

\subsection{Sources, Annotation, and Coverage}

\figbench{} comprises 603 reaction figures annotated with 8,547 concrete reaction records. The figures are drawn from the chemical literature, including Synfacts, and the ChemEAGLE and OpenChemIE collections~\citep{chemeagle2024,openchemie2024}.  Annotation and review are performed by undergraduate students with a chemistry background. Annotators encode molecular structures and templates using the E-SMILES and assemble the reactants and products associated with each experimental entry. An initial annotation round is followed by iterative review and revision until approval. Source composition, input preparation, annotation guidelines, and review procedures are detailed in Appendix~\ref{app:benchmark}.

\begin{figure}[t]
\centering
\includegraphics[width=\linewidth]{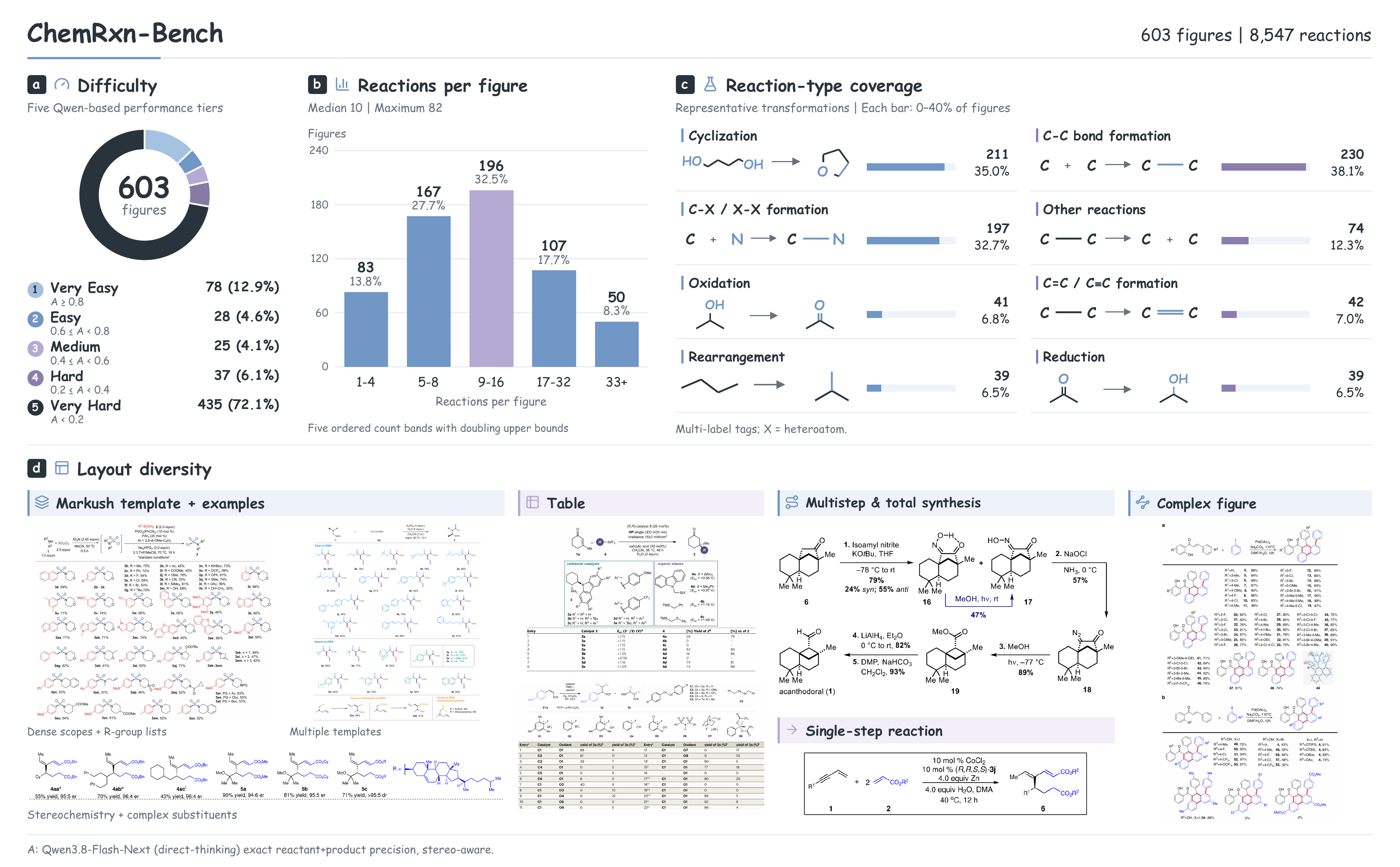}
\caption{\textbf{Composition of \figbench{}.}
\textbf{(a)} Difficulty tiers defined by Qwen3.8-Flash-Next performance.
\textbf{(b)} Distribution of concrete reaction records per figure.
\textbf{(c)} Coverage of eight reaction types with representative
transformations. Percentages use all figures as the denominator;
multiple tags may apply to each figure. X denotes a heteroatom.
\textbf{(d)} Representative layout examples.
Tier definitions and source attributions, including licenses and
modifications to the examples in (d), are provided in
Appendix~\ref{app:taxonomy}.}
\label{fig:benchmark-overview}
\end{figure}

The benchmark spans single-reaction schemes, Markush tables, substrate scopes, optimization tables, and multi-step schemes. We characterize this coverage using layout tags, reaction-type tags, and the number of reference reaction records per figure. These tags are fixed before evaluation. We additionally define performance-based difficulty tiers using a direct vision--language baseline; these tiers reflect the baseline's performance rather than intrinsic figure difficulty. Figure~\ref{fig:benchmark-overview} summarizes these distributions, and Appendix~\ref{app:taxonomy} provides the tagging procedures and tier definitions.

\subsection{Evaluation Protocol}
\label{sec:protocol}

We apply the same normalization procedure to predicted and reference molecular structures. Reaction records are then matched one to one within each figure, with records sharing identical molecular participants retained as separate instances. A match requires exact agreement on both reactant and product sets, including stereochemistry. These matches determine per-figure reaction precision; reaction recall is the fraction of reference products recovered, and F1 combines the two, with zero scores assigned to figures with no predictions. We then average each metric across figures and rank systems by macro-averaged reaction F1. Repeating the evaluation without stereochemistry provides a complementary diagnostic. Normalization, matching, and failure-handling rules are detailed in Appendix~\ref{app:scorer}.

As a complementary assessment, we use a multimodal LLM judge, Claude Opus 5.5, to evaluate molecular participants, conditions, and outcomes under a fixed rubric. Judge scores are reported separately and do not affect the primary ranking. In an expert audit of sampled judge assessments, more than 90\% were judged correct. Appendix~\ref{app:judge} provides the judge configuration, scoring procedure, and validation limitations.


\section{\agentname{}: From Visual Evidence to Concrete Reaction Records}
\label{sec:agent}

\begin{figure}[tb]
\centering
\includegraphics[width=0.95\linewidth]{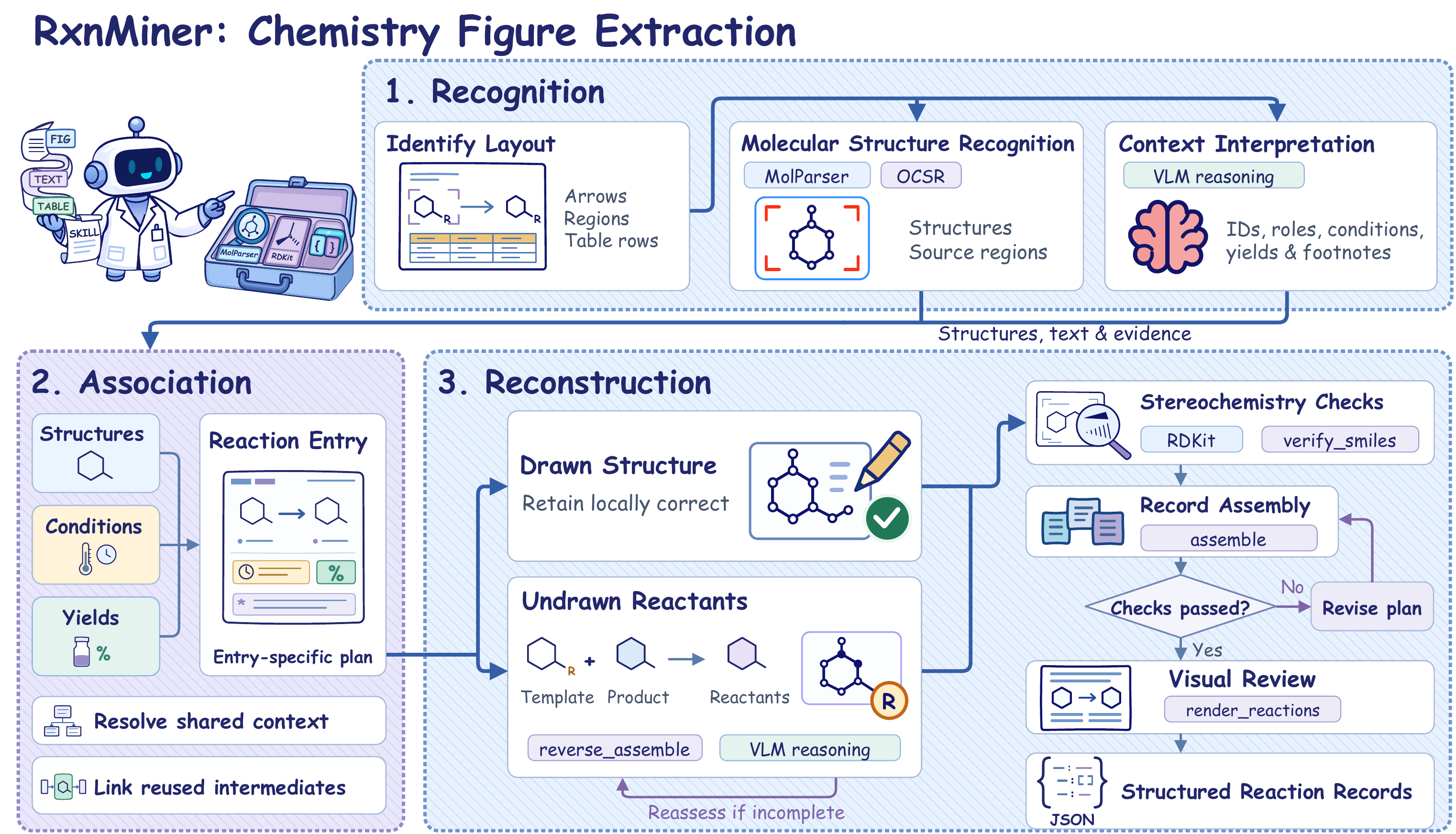}
\caption{\textbf{Workflow of \agentname{}.}
The agent combines visual reasoning and chemical tools for recognition,
entry-level association, and template-based reconstruction.
Recovered participants are assembled into source-linked reaction records,
with stereochemical checks and visual review supporting validation.}
\label{fig:agent}
\end{figure}

\agentname{} is built on the Pi agent framework~\citep{zechner2026pi} (hereafter Pi), an open-source agent toolkit that provides the agent loop, the model interface, and tool calling. On top of this base agent, \agentname{} adds an expert-informed extraction procedure and a chemistry toolbox to recover concrete reaction records. We encode expert extraction practices as structured instructions that guide the backbone to read molecular structures, associate evidence with experimental entries, and reconstruct indirectly specified participants. Following these instructions, the vision--language backbone constructs an entry-specific plan, selects tool calls, and interprets their outputs. The tools provide molecular recognition, template instantiation, and record assembly. Together, these components support recognition, association, and reconstruction, as illustrated in Figure~\ref{fig:agent}.

\paragraph{Recognition.}
Following a read-first principle, the agent applies MolParser~\citep{molparser} to obtain molecular structures and templates in Extended SMILES, together with bounding boxes and confidence scores. These readings are cached with their source regions and reused throughout extraction. When the full-figure pass misses a structure, the backbone selects a crop for an additional recognition call. It also reads compound identifiers, substituent assignments, conditions, and footnotes. Recognition supplies the molecular representations and printed evidence needed to recover individual experimental entries.

\paragraph{Association.}
The backbone organizes this evidence into an entry-specific plan. For each entry, the plan identifies the molecular readings assigned to reactant and product roles, the applicable shared templates, and the bindings between substituent labels and template positions. Conditions and outcomes are associated with the same entry through compound identifiers, table rows, and footnotes. This intermediate representation connects interpretation to chemical operations: reconstruction uses the selected templates and bindings to instantiate participants, while assembly uses their entry assignments to preserve reactant--product pairing.

\paragraph{Reconstruction.}
Explicitly drawn concrete participants retain their tool readings. Participants specified indirectly are instantiated by a deterministic R-group engine using explicit entry bindings, product-template matching, or both. When an entry specifies its substituents, the engine fills the corresponding template positions directly. When a product scope leaves reactants undrawn, it matches the recognized product to its associated product template, recovers substituent bindings, and applies the relevant bindings to the reactant templates.

Product-derived bindings are combined with printed assignments because the product may not retain all reactant information. For example, if a reactant template contains a leaving-group variable X, an entry annotation such as X = I specifies iodine at that position. Once this group is replaced during the reaction, its identity cannot be recovered from the product alone and must be read from the associated entry. Reconstruction therefore instantiates supplied templates from associated evidence rather than predicting a retrosynthetic route. If the engine reports a derivation failure, the backbone may provide a model-written fallback, explicitly distinguished from tool-read and template-derived structures.

Structures with stereochemical evidence are checked and rendered with RDKit for comparison against the source. The assembly tool combines the molecular participants and their entry associations into concrete reaction records, retaining conditions, outcomes, and provenance. Each participant references its source region, its derivation template and bindings, or its model-written fallback designation. The assembled reactions are rendered for final inspection. These checks support error detection but do not guarantee chemical correctness. Appendix~\ref{app:agent} provides the tool interfaces, verification procedures, and execution configuration.

\section{Experiments and Analysis}
\label{sec:results}
\label{sec:evaluation}

We investigate four research questions.
\textbf{RQ1:} How well do existing extraction systems and vanilla LLMs recover concrete reaction records on \figbench{}?
\textbf{RQ2:} Can \agentname{} enable a locally deployed model to outperform existing extraction systems and vanilla LLMs?
\textbf{RQ3:} What is the joint contribution of the expert-informed extraction procedure and chemistry toolbox?
\textbf{RQ4:} How do extraction performance and gains vary across backbones?

\subsection{Experimental Setup}

All primary comparisons use the same 603 figure-level inputs and the matching protocol in Section~\ref{sec:protocol}, retaining failed runs in the evaluation. We evaluate three configurations. Vanilla LLMs generate reaction records from the figure, task instruction, and output schema without external tools. Bare Pi is the base agent: the unmodified Pi framework (Section~\ref{sec:agent}) without the extraction procedure and chemistry toolbox. \agentname{} extends that framework with expert-informed extraction procedure and a chemistry toolbox, allowing us to evaluate the added design as a package. We report results for both \agentname{} and Bare Pi across five backbones. Baseline versions, execution configurations, budgets, and output adapters are documented in Appendix~\ref{app:comparison}.

\subsection{End-to-End Reaction Extraction}

\begin{table*}[t]
\centering
\small
\caption{End-to-end reaction extraction on \figbench{}.
Scores are percentages, macro-averaged over all 603 figures.
Precision, Recall, and F1 include stereochemistry; the final column omits it. The best results are highlighted in \textbf{bold}.
The vanilla LLM reference is our primary locally deployed backbone.}
\label{tab:extraction-systems}
\begin{tabular}{lrrrr}
\toprule
\shortstack{System / \\Configuration} & Precision & Recall & \shortstack{F1} & \shortstack{F1 w/o \\stereochemistry} \\
\midrule
\multicolumn{5}{l}{\textit{Existing extraction systems}} \\
ChemEAGLE & 25.1 & 49.7 & 25.0 & 35.0 \\
OpenChemIE & 9.8 & 8.1 & 6.0 & 9.0 \\
MERMaid & 5.8 & 15.2 & 6.0 & 8.0 \\
RxnScribe & 6.1 & 3.4 & 2.0 & 3.0 \\
MinerU.Chem & 7.1 & 2.6 & 2.0 & 3.0 \\
RDE2 & 2.0 & 1.8 & 1.0 & 1.0 \\
RxnCaption & 0.0 & 0.0 & 0.0 & 0.0 \\
RxnID & 0.0 & 0.0 & 0.0 & 0.0 \\
\midrule
\multicolumn{5}{l}{\textit{Vanilla LLM}} \\
Qwen3.8-Flash-Next & 20.0 & 36.3 & 21.0 & 33.0 \\
\midrule
\agentname{} (Qwen3.8-Flash-Next) & \textbf{69.9} & \textbf{89.2} & \textbf{71.0} & \textbf{78.0} \\
\bottomrule
\end{tabular}
\end{table*}

Table~\ref{tab:extraction-systems} compares eight existing extraction systems with representative vanilla LLM and \agentname{} configurations. Output adapters convert predictions to the common record format without reference-dependent repair or additional reconstruction. Qwen3.8-Flash-Next is our primary locally deployed backbone.

\noindent\textbf{RQ1:}
Even the strongest existing extraction system, ChemEAGLE, achieves only 25.0 reaction macro-F1, indicating that extensive manual checking remains necessary before its outputs can support reliable downstream data curation. Our primary locally deployed Qwen model reaches 21.0 as a vanilla LLM. Kimi K3 reaches 40.0 reaction macro-F1 as a vanilla LLM
(Table~\ref{tab:main-results}), outperforming the existing
extraction systems. This highlights the potential of general-purpose models while leaving substantial room for more reliable record extraction.

\noindent\textbf{RQ2:}
With the locally deployed Qwen backbone, \agentname{} reaches
71.0 reaction macro-F1, the highest score among the evaluated
configurations. It surpasses all evaluated existing extraction
systems and vanilla LLMs, demonstrating the effectiveness of
the workflow with a locally deployed backbone.

The improvement extends to both reaction precision and recall
(Table~\ref{tab:extraction-systems}), indicating broader recovery
of reference products together with a higher proportion of
exactly matched predicted reactions. Under the exact-match protocol,
each match requires agreement on the fully instantiated reactant
and product sets, including stereochemistry. These gains therefore
provide end-to-end evidence of improved recovery of molecular
participants and their reaction-level associations, directly
addressing the extraction target motivated by compositional figures.

\subsection{Effect of the Chemistry-Specific Design}

\begin{table*}[t]
\centering
\small
\setlength{\tabcolsep}{4pt}
\caption{Backbone-matched comparison of Vanilla LLM, Bare Pi, and \agentname{}.
Scores are stereochemistry-aware macro-F1 (\%) on all 603 figures.
The best results are highlighted in \textbf{bold}.
$\Delta_{\mathrm{Pi}}$ and $\Delta_{\mathrm{Vanilla}}$ denote gains over Bare Pi and Vanilla LLM, respectively, in percentage points.}
\label{tab:main-results}

\begin{tabular}{lrrrrr}
\toprule
& \multicolumn{3}{c}{Reaction F1 (\%)} & \multicolumn{2}{c}{$\Delta$ Reaction F1 (\%)} \\
\cmidrule(lr){2-4}\cmidrule(lr){5-6}
Backbone & Vanilla LLM & Bare Pi & \agentname{} & $\Delta_{\mathrm{Pi}}$ & $\Delta_{\mathrm{Vanilla}}$ \\
\midrule
Qwen3.8-Flash-Next & 21.0 & \textbf{59.6} & \textbf{71.0} & +11.4 & \textbf{+50.0} \\
DeepSeek V4.1 Flash & 16.0 & 54.6 & 64.0 & +9.4 & +48.0 \\
GLM 5.3 Flash & 31.0 & 15.5 & 41.0 & \textbf{+25.5} & +10.0 \\
Kimi K3 & \textbf{40.0} & 57.7 & 69.0 & +11.3 & +29.0 \\
Doubao Seed 2.1 Pro & 34.0 & 26.5 & 47.0 & +20.5 & +13.0 \\
\bottomrule
\end{tabular}
\end{table*}

\noindent\textbf{RQ3:}
Table~\ref{tab:main-results} shows that \agentname{} outperforms Bare Pi on all five available backbone pairs, with reaction macro-F1 gains of 9.4--25.5 percentage points. This comparison assesses the expert-informed extraction procedure and chemistry toolbox as a joint extension of the Pi framework. Bare Pi alone underperforms vanilla LLM prediction on GLM and Doubao, whereas \agentname{} improves over both alternatives. The consistent advantage supports the effectiveness of the combined chemistry-specific design.

\subsection{Effect of Backbone Choice}

\noindent\textbf{RQ4:}
Table~\ref{tab:main-results} shows that all evaluated backbones benefit from \agentname{}, although $\Delta_{\mathrm{Vanilla}}$ ranges from 10.0 to 50.0 percentage points. Qwen and DeepSeek show the largest gains over their vanilla
counterparts, with Qwen achieving the highest final extraction score.

The stereochemistry-blind evaluation further clarifies the scope
of these gains. With the Qwen backbone, \agentname{} achieves
78.0 reaction macro-F1 under this criterion, compared with 33.0
for vanilla prediction (Table~\ref{tab:extraction-systems}).
The advantage thus persists when stereochemical distinctions
are excluded from matching. Nevertheless, the seven-percentage-point
gap between the agent's stereochemistry-aware and stereochemistry-blind
scores identifies stereochemical agreement as a remaining challenge,
even with the verification procedures described in
Section~\ref{sec:agent}.

\section{Scaling \agentname{} for Reaction Corpus Construction}
\label{sec:corpus}

\begin{figure}[ht]
\centering
\includegraphics[width=\linewidth]{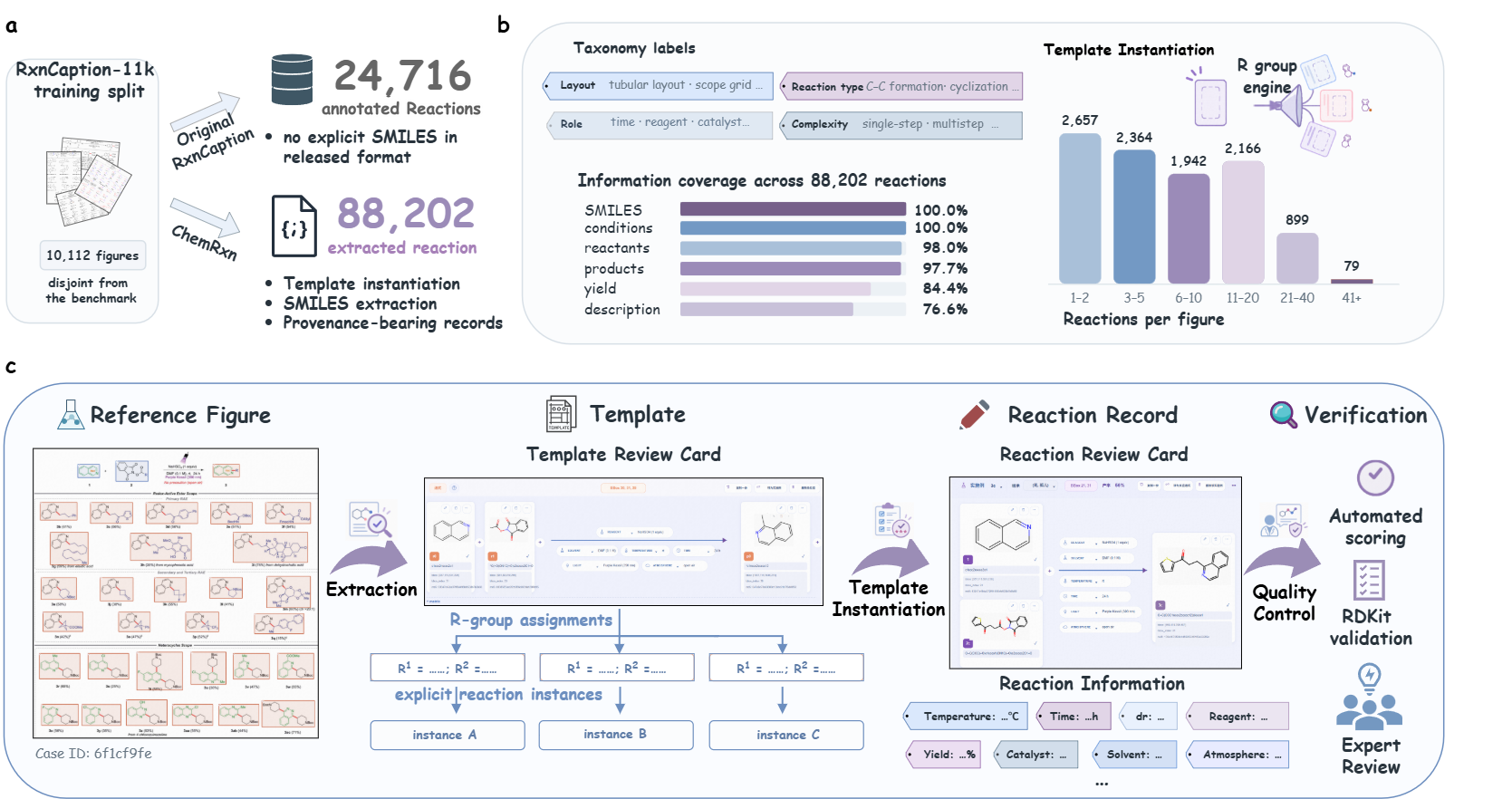}
\caption{\textbf{Recovering reaction information at scale.}
\textbf{(a)} Applying \agentname{} to the RxnCaption-11k training
split produces source-linked concrete reaction records with
molecular SMILES and provenance.
\textbf{(b)} Corpus attributes, record-field coverage, and the
distribution of extracted reactions per figure.
\textbf{(c)} An example of instantiating a shared template,
R-group assignments, and reaction conditions into explicit
reaction records, with interfaces for source-based inspection.}
\label{fig:silver-corpus-quality}
\end{figure}

To examine whether the extraction capabilities measured on
\figbench{} extend to literature-scale corpus construction, we
applied \agentname{} to all 10,112 figures in the RxnCaption-11k
training split~\citep{rxncaption2026}. Using the same extraction
configuration evaluated on \figbench{}, the agent successfully
processed \corpusfigures{} figures and, without human intervention,
produced \corpusreactions{} source-linked reaction records
(Figure~\ref{fig:silver-corpus-quality}a). This corresponds to
approximately 3.6 times the 24,716 reaction annotations associated
with the full source collection. Distinct experimental entries
remain separate in our corpus even when their molecular
participants are identical.

The extraction workflow expands information shared across figure
components into individual reaction records. In substrate-scope
tables and Markush schemes, a shared scaffold, reaction conditions,
and R-group assignments can jointly encode many individual
reactions. \agentname{} instantiates these assignments into explicit
molecular structures and reconstructs reactants that are implied
by the figure but not separately drawn. For multistep schemes,
it decomposes the overall scheme into individual reaction steps
and associates the corresponding conditions with each step.
These operations convert shared figure-level representations into
concrete reaction records.

Each extracted reaction is stored in our unified JSON representation
with molecular SMILES, extracted reaction conditions, source-figure
references, and structure provenance. We characterize the corpus
by figure layout, reaction type, record-field coverage, and
source-scheme complexity, distinguishing single-step from multistep
schemes (Figure~\ref{fig:silver-corpus-quality}b). Record-field
coverage includes reaction time, reagents, and catalysts.
The template and reaction review cards display the source template,
R-group assignments, and instantiated reaction together,
facilitating inspection of template instantiation and entry-level
associations (Figure~\ref{fig:silver-corpus-quality}c).

These automatically generated records constitute a silver corpus.
The extraction configuration is evaluated on \figbench{} in
Section~\ref{sec:results}. At corpus scale, RDKit successfully
parsed 25,223 of 25,225 molecular strings (99.99\%) in a random
sample of 1,000 figures. This result indicates high molecular-string
parseability, but does not establish agreement with the source
figures or correct reactant--product pairing. The corpus has not
undergone manual validation.

This experiment demonstrates the use of \agentname{} for
large-scale corpus construction. By expanding shared templates,
recovering indirectly specified participants, and separating
multistep schemes, it produces a substantially denser, source-linked
reaction resource for downstream reaction mining and for selecting
candidate records for future benchmark annotation. Details of the
extraction run, corpus statistics, and parseability checks are
provided in Appendix~\ref{app:corpus}.
\section{Conclusion}
\label{sec:conclusion}

We presented \agentname{}, an agent that combines an expert-informed
extraction procedure with a chemistry toolbox to recover concrete
reaction records from evidence distributed across reaction figures.
Its workflow integrates molecular recognition, entry-level association,
and template-based reconstruction. We also introduced \figbench{} to
evaluate these records under a common protocol that measures molecular
identity and reactant--product pairing. Experiments reveal substantial
limitations in existing extraction systems. With the locally deployed
Qwen backbone, \agentname{} raises stereochemistry-aware reaction
macro-F1 from 21.0 to 71.0, surpassing the strongest evaluated vanilla
LLM. Gains over Bare Pi across the evaluated backbone pairs support
the combined value of the expert skills and chemistry toolbox.
Applying \agentname{} at scale yields a source-linked silver corpus
of \corpusreactions{} reaction records. We release the agent and scorer
openly, the benchmark on request, and the corpus openly, to support reproducible evaluation and
scalable curation of reaction knowledge from the chemical literature.

\clearpage
\bibliographystyle{iclr2027_conference}
\bibliography{references}

\appendix

\clearpage
\part{Appendix}
\etocsettocstyle{}{}
\localtableofcontents
\clearpage

\section{Benchmark Construction}
\label{app:construction}

\subsection{Sources, Annotation, and Task Packages}
\label{app:benchmark}

\begin{figure}[htbp]
\centering
\includegraphics[width=1.00\linewidth]{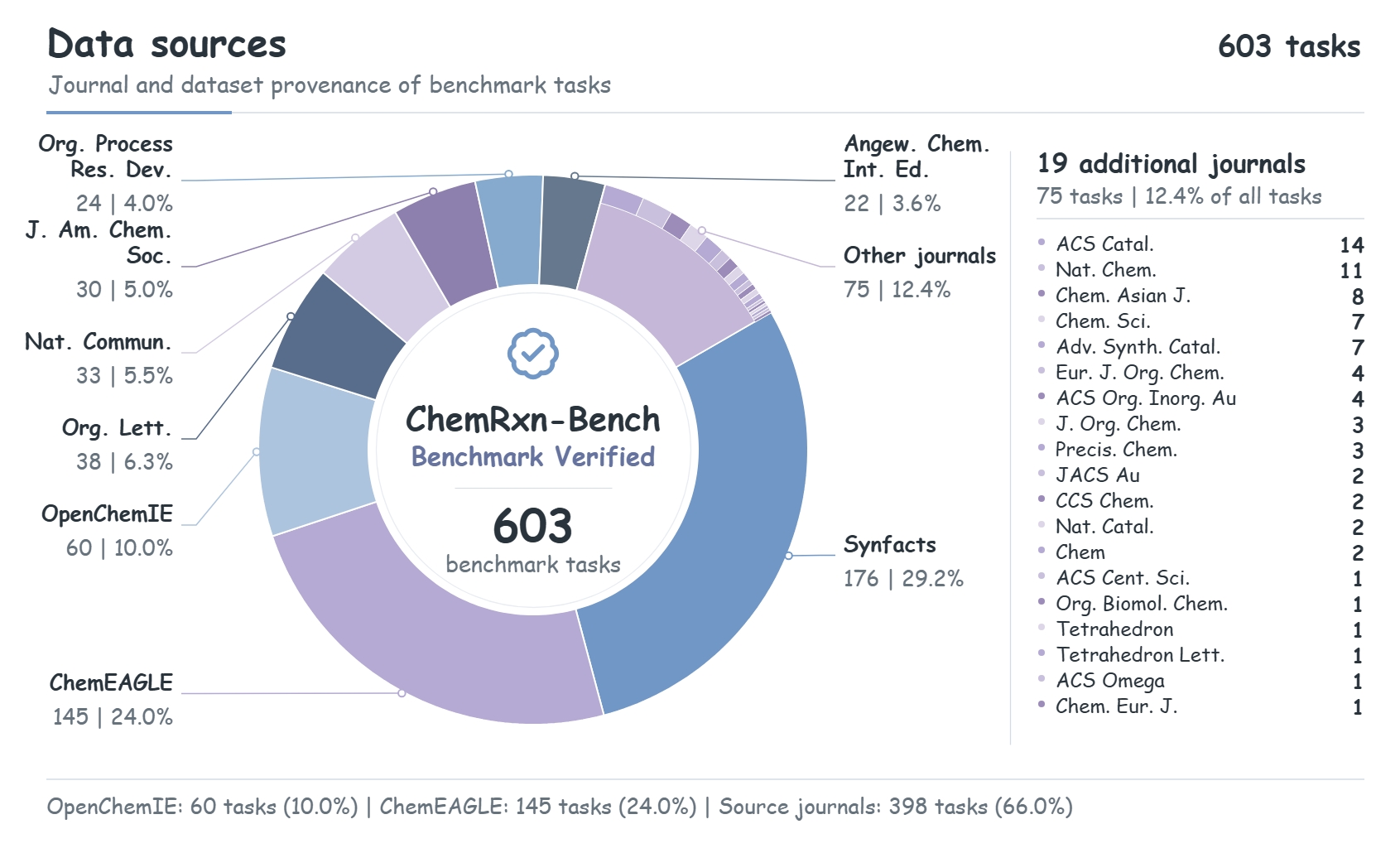}
\caption{\textbf{Provenance of the 603 benchmark figures.} Journal figures (398, 66.0\%) span 25 journals, led by Synfacts; the ChemEAGLE (145, 24.0\%) and OpenChemIE (60, 10.0\%) collections supply the rest. The right panel lists the 19 journals grouped as other journals (75 figures).}
\label{fig:sources}
\end{figure}

\paragraph{Source composition.}
The 603 figures come from three groups (Figure~\ref{fig:sources}). Journal figures account for 398 (66.0\%) and span 25 journals: Synfacts (176), Organic Letters (38), Nature Communications (33), the Journal of the American Chemical Society (30), Organic Process Research \& Development (24), Angewandte Chemie International Edition (22), and 19 further journals with 75 figures between them. The remaining figures come from the collections used by ChemEAGLE (145, 24.0\%) and OpenChemIE (60, 10.0\%).

\paragraph{Task package.}
Table~\ref{tab:task-package} lists the files of one task. The container image is built from a shared base image that pins the chemistry stack, including RDKit 2022.9.5, and adds only the figure, so the environment carries no information derived from the reference. The metadata in \texttt{task.toml} record the figure's source, layout, reaction types, record count, and difficulty tier, together with the resource limits of the agent and the verifier. The verifier writes Reaction P, Reaction R, Reaction F1, and Molecule F1 under both strict-stereochemistry and connectivity-only scoring. The scores reported in this paper are recomputed from each run's saved answer with the released scorer (Appendix~\ref{app:scorer}), with strict-stereochemistry Reaction F1 as the primary metric.

\begin{table}[htbp]
\centering
\small
\caption{\textbf{Contents of one \figbench{} task in the Harbor format.} The last column states whether the file is visible to the system under evaluation.}
\label{tab:task-package}
\begin{tabular}{p{0.2\linewidth}p{0.55\linewidth}p{0.16\linewidth}}
\toprule
File & Contents & Visible \\
\midrule
\texttt{instruction.md} & Task statement, output schema, and scoring rules & Yes \\
\texttt{environment/} & Container definition: the shared base image plus the figure & Yes \\
\texttt{task.toml} & Metadata (source, layout, reaction types, record count, difficulty) and resource limits & No \\
\texttt{tests/} & Reference record and verifier script & No; mounted after the run \\
\texttt{solution/} & Reference answer, used to check that the verifier returns full credit & No \\
\bottomrule
\end{tabular}
\end{table}

\paragraph{Datasheet and source linkage.}
Following the documentation principles of \citet{gebru2021datasheets}, the benchmark datasheet records source composition, acquisition, annotation and review, available context, and known metadata gaps. The linkage chain connects a figure identifier to its caption sidecar and then to the source PDF's DOI, page, and figure label, where recoverable. The existing self-collected sources retain figure-to-document traceability; entries imported from prior benchmarks may lack DOI metadata. Recoverable licence metadata is recorded as provenance, without implying a collection-wide permission determination.

\paragraph{Annotation unit and review.}
Annotation targets the concrete reaction records expressed by a figure, including instances represented through shared templates and local bindings. The relevant decisions concern molecular identities, reaction roles, associations, and the stereochemistry represented in the figure. The designation expert-reviewed describes review status without, by itself, specifying how disagreements were resolved.

\paragraph{Unsuccessful chemical experiments.}
Low-yielding and unsuccessful reported experiments remain reference entries. For an unsuccessful experiment, the depicted target product and the unsuccessful outcome are retained together. The record therefore describes an attempted transformation rather than asserting product formation. Although the outcome field is unscored, it remains essential to interpreting the experimental result.

\paragraph{Failed system runs.}
A timeout, tool error, malformed response, or missing prediction does not remove a figure a system has run from its evaluation. Under the evaluation policy, failed system runs receive zero extraction credit, with completion and failure status reported separately.  Reference-label corrections change the annotation version and apply consistently across systems; they are distinct from execution failures.

\subsection{Taxonomy and Coverage}
\label{app:taxonomy}

\paragraph{Purpose.}
The taxonomy characterizes the demands that each figure places on extraction. It organizes coverage and supports the subgroup analyses in Section~\ref{sec:results}; it does not assign a unique cause to a system's errors. Layout, reaction type, and record count vary together, so a difference between two subgroup averages cannot by itself identify which property caused it.

\paragraph{Axes.}
Figure~\ref{fig:benchmark-overview} summarizes four axes, and Table~\ref{tab:taxonomy-inventory} gives their counts over the 603 figures. \emph{Layout} assigns each figure one of five classes. \emph{Reaction type} is multi-label: a figure receives every class its reactions belong to, so the shares do not sum to one. \emph{Record count} is the number of reference reaction records, binned in five bands with doubling upper bounds (median 10, maximum 82). \emph{Difficulty} is not a tag but a model-performance proxy: each figure is binned by the stereo-aware exact reactant and product precision $A$ of Qwen3.8-Flash-Next in the direct setting, so the tiers describe difficulty for that baseline rather than an intrinsic property of the figure.

\begin{table}[htbp]
\centering
\small
\caption{\textbf{Coverage of the 603 benchmark figures.} Counts are figures, with shares of 603. Reaction types are multi-label.}
\label{tab:taxonomy-inventory}
\begin{tabular}{p{0.19\linewidth}p{0.74\linewidth}}
\toprule
Axis & Classes, figures (share) \\
\midrule
Layout & Markush template with examples 425 (70.5\%); table 161 (26.7\%); single-step reaction 6 (1.0\%); complex figure 6 (1.0\%); multistep and total synthesis 5 (0.8\%) \\
Reaction type & C--C bond formation 230 (38.1\%); cyclization 211 (35.0\%); C--X or X--X bond formation 197 (32.7\%); other 74 (12.3\%); C=C or C$\equiv$C bond formation 42 (7.0\%); oxidation 41 (6.8\%); rearrangement 39 (6.5\%); reduction 39 (6.5\%) \\
Record count & 1--4: 83 (13.8\%); 5--8: 167 (27.7\%); 9--16: 196 (32.5\%); 17--32: 107 (17.7\%); 33 or more: 50 (8.3\%) \\
Difficulty & very easy, $A \geq 0.8$: 78 (12.9\%); easy, $0.6 \leq A < 0.8$: 28 (4.6\%); medium, $0.4 \leq A < 0.6$: 25 (4.1\%); hard, $0.2 \leq A < 0.4$: 37 (6.1\%); very hard, $A < 0.2$: 435 (72.1\%) \\
\bottomrule
\end{tabular}
\end{table}

\paragraph{Counts and versions.}
All counts refer to the 603-figure release and its 8,547 tagged reaction records. The normalized reference records used by the scorer number 8,510; the difference of 37 comes from three optimization tables whose metadata counts each entry twice. The layout panel of Figure~\ref{fig:benchmark-overview} shows the range of layouts rather than their frequencies, which are given here.

\subsection{Extraction Operations}
\label{app:operations}

\begin{figure}[!htb]
\centering
\includegraphics[width=\linewidth]{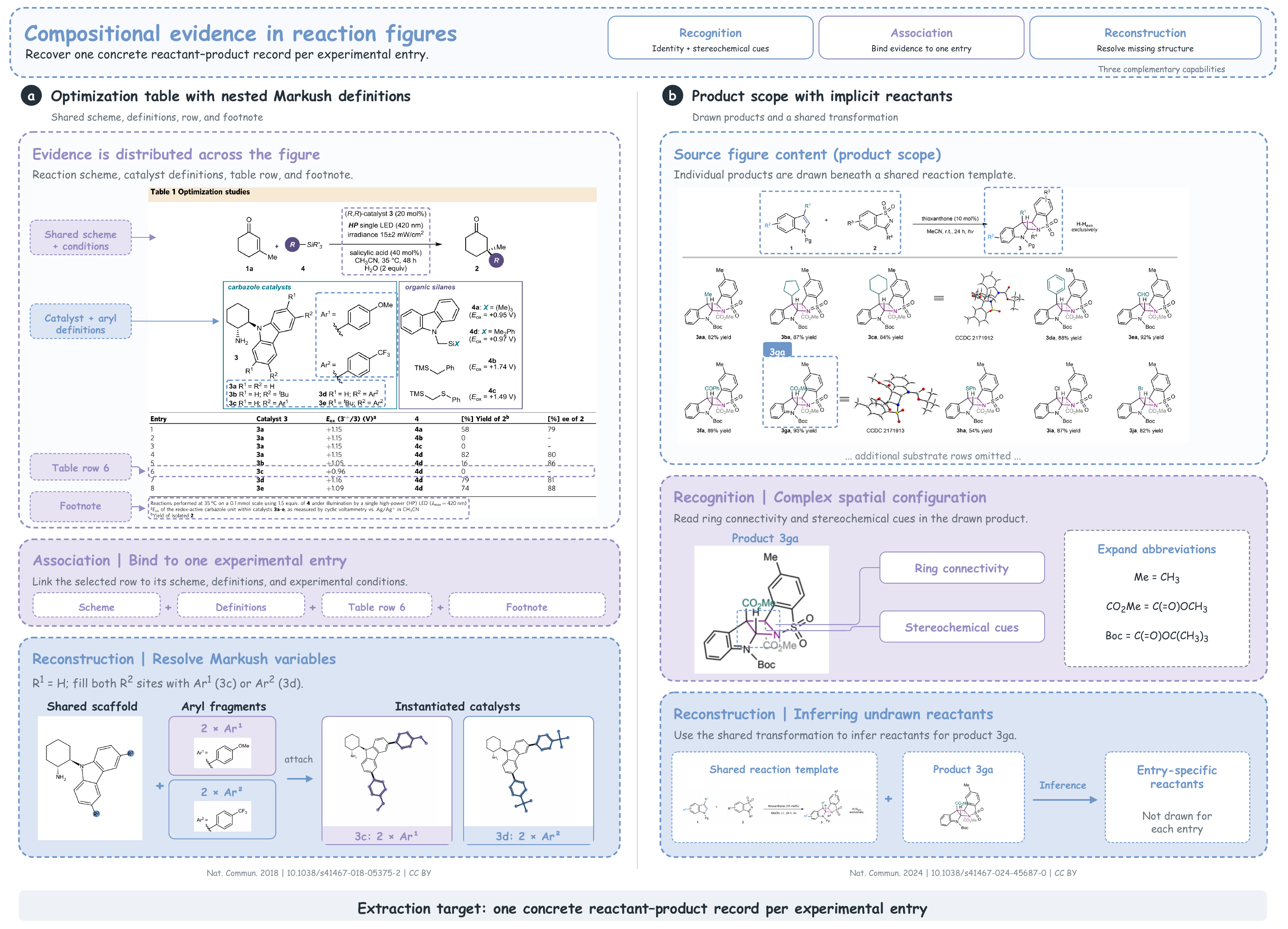}
\caption{\textbf{Extraction operations illustrated through compositional reaction figures.} \textbf{(a)} An optimization table requires associating a selected entry with the shared reaction scheme, catalyst definitions, and footnotes, and instantiating concrete catalysts from a scaffold and aryl fragments. \textbf{(b)} A product scope requires recognizing molecular connectivity, stereochemical cues, and abbreviated groups, and combining each product with the shared transformation to recover its undrawn reactants. Together, the examples illustrate recognition, association, and reconstruction rather than a measured agent run. Adapted from \emph{Nature Communications} (2018), doi:10.1038/s41467-018-05375-2, and \emph{Nature Communications} (2024), doi:10.1038/s41467-024-45687-0, under CC BY.}
\label{fig:example}
\end{figure}

Reaction extraction combines recognition, association, and reconstruction to recover concrete reaction records from distributed visual evidence. Figure~\ref{fig:example} illustrates how these capabilities interact in an optimization table and a product scope. The seven operations below describe the task requirements in more detail; a single figure may require several operations without exercising all seven. These operations are interdependent: a correctly recognized molecular structure may be assigned to the wrong reaction, while a correctly identified shared template may be instantiated with the wrong row. \figbench{} evaluates the resulting concrete reaction records through the protocol in Section~\ref{sec:protocol}, rather than scoring the seven operations independently.

\begin{enumerate}
\item \textbf{Structure reading.} A drawn molecular structure is converted into a structural representation while preserving its identity and connection to the source region.
\item \textbf{Identifier resolution.} Printed compound identifiers are associated with the structures or entries to which they refer, including references elsewhere in the supplied figure context.
\item \textbf{Condition attachment.} Reaction conditions are associated with the applicable reaction or scope entry. Reading a condition string alone does not determine which records inherit it.
\item \textbf{Markush enumeration.} Variable-bearing structures are instantiated using the substituents and combinations specified by the figure. The shared template alone does not constitute a concrete reaction record.
\item \textbf{Undrawn-reactant inference.} Reactant structures expressed implicitly through shared templates and instance-specific evidence are recovered. Their template-derived or model-written origin distinguishes them from tool-read structures.
\item \textbf{Stereochemistry serialization.} Stereochemical information expressed by the figure is preserved in the molecular representation; stereochemistry left unspecified by the evidence remains unspecified.
\item \textbf{Reaction-graph assembly.} Identified reaction instances are connected through their molecular participants and figure-level relationships, while retaining their identities as concrete reaction records.
\end{enumerate}

These operations cut across the stages of the agent's workflow (Section~\ref{sec:agent}). Their enumeration clarifies the task demands without implying that every operation has a separate annotation target, tool, or metric.

\subsection{Output Records and Scored Information}
\label{app:records}

\begin{figure}[htbp]
\centering
\includegraphics[width=\linewidth]{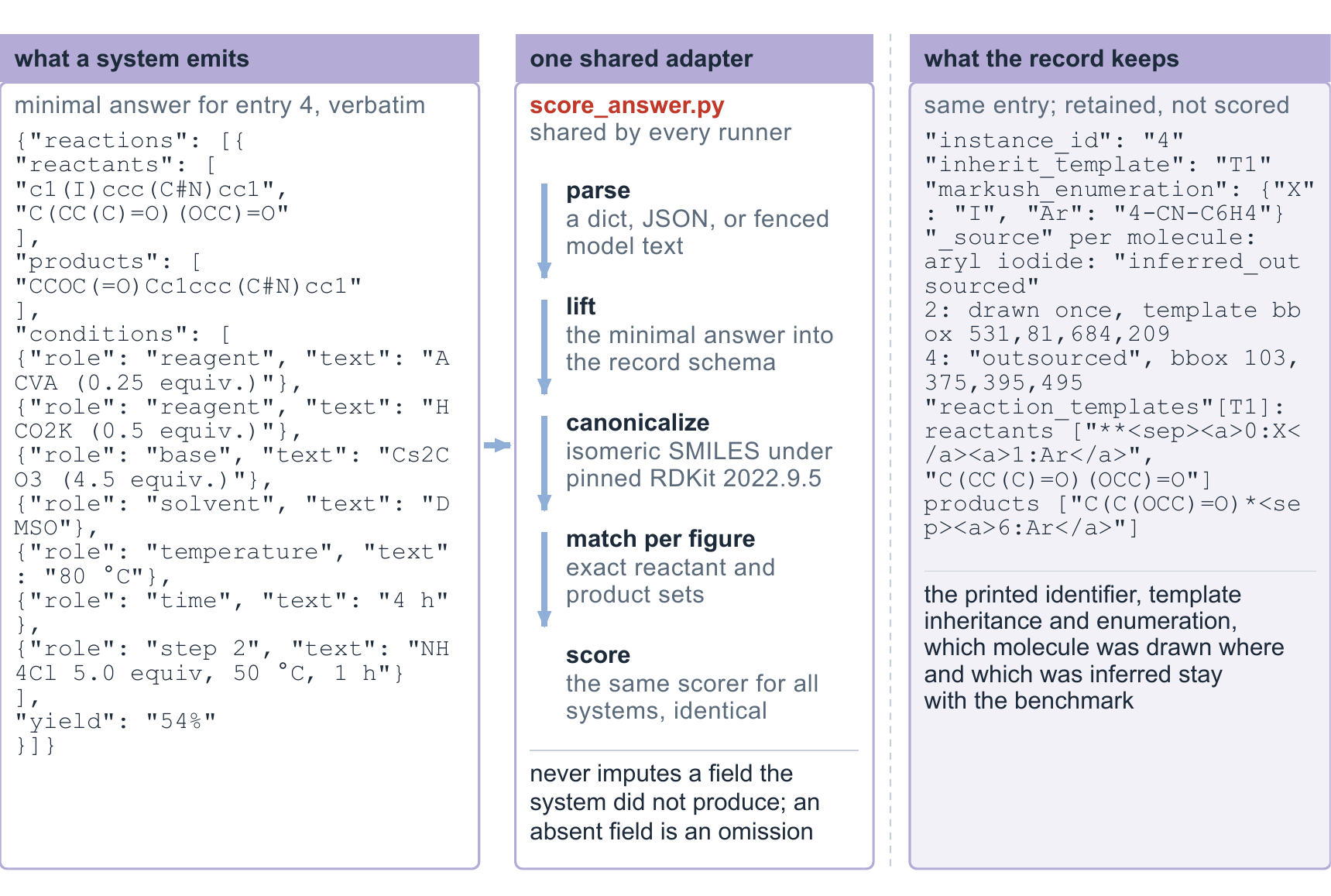}
\caption{\textbf{System answers, common adaptation, and retained evidence.} Left: an example JSON answer contains reactant and product SMILES, conditions, and yield. Center: the adaptation, canonicalization, and matching path shared by every system (Section~\ref{sec:protocol}). Right: the richer record retains identifiers, template inheritance, substituent bindings, and molecular provenance. The rule-based score evaluates molecular participants and their pairing; conditions and outcomes are assessed only by the secondary judge, and the other retained fields are not evaluated.}
\label{fig:record-adapter}
\end{figure}

\paragraph{Representation boundary.}
The unified JSON representation stores concrete reaction records and their supporting information. A molecular structure denotes a chemical entity; a chemical formula does not encode the same structural information. A concrete reaction record associates reactants and products for one instantiated reaction, with available conditions, yields, identifiers, and reaction-graph links. Recovering an isolated product is therefore different from recovering a complete record. A variable-bearing template alone is not the concrete instance represented by a particular scope entry. Figure~\ref{fig:example} links visual evidence to extraction operations, while Figure~\ref{fig:record-adapter} illustrates the resulting record and scoring boundary; neither example estimates extraction accuracy.

\begin{table}[htbp]
\centering
\small
\caption{\textbf{Semantic contents of the record representation.} Names describe field meanings rather than literal JSON keys. Scoring refers to the rule-based reaction metrics and the secondary LLM judge (Section~\ref{sec:protocol}).}
\label{tab:record-semantics}
\begin{tabular}{p{0.24\linewidth}p{0.48\linewidth}p{0.17\linewidth}}
\toprule
Content & Stored information & Scored \\
\midrule
Reactants & Molecular structures as SMILES & Yes, with pairing \\
Products & Molecular structures as SMILES & Yes, with pairing \\
Conditions & Attached reaction conditions & Judge only \\
Outcomes & Metrics and their roles, including yields & Judge only \\
Identifiers & Printed compound or entry identifiers & No \\
Provenance & Source boxes and template bindings & No \\
Reaction-graph links & Connections between reaction instances & No \\
\bottomrule
\end{tabular}
\end{table}

\paragraph{Scoring boundary.}
The primary evaluation uses molecular recovery and reactant--product pairing. Conditions, yields, identifiers, and reaction-graph links are retained in the output but are outside the current primary score. Their presence in a JSON object does not establish that they were evaluated or verified. Similarly, provenance supports inspection without contributing extraction credit. Conditions and outcomes are assessed by the secondary LLM judge of Section~\ref{sec:protocol}; identifiers and reaction-graph links are not evaluated, and reaction F1 alone evaluates none of them.

Table~\ref{tab:record-semantics} describes the existing fields semantically without specifying unverified literal keys or defaults. Missing information and asserted values have different evidential status, while explicit abstentions receive the same recall treatment as other incomplete predictions.

\section{Evaluation Protocol}
\label{app:evaluation}

\subsection{Scorer and Prediction Adapters}
\label{app:scorer}

\paragraph{Metric implementation.}
The primary metric is stereochemistry-aware reaction F1 computed per figure and averaged across the manifest. The scorer is released with the benchmark at a fixed revision (\texttt{b3e1377}). It canonicalizes molecules to isomeric SMILES with RDKit 2022.9.5, splits dot-separated salts into their components, and treats common metal counter-ions as equivalent across salt forms. The stereochemistry-blind diagnostic uses the corresponding matching procedure with stereochemical distinctions removed; it complements the primary result. None of these scores establishes the correctness of unscored record fields.

\paragraph{Adapters.}
Each system's output is converted into the common representation using a versioned adapter. The adapter accounts for representational differences, including molecule lists, reaction grouping, and omitted fields. The evaluation policy retains malformed predictions in precision accounting, because silently removing them can improve a score without improving extraction. Predictions that cannot be converted retain a failure status and remain part of the evaluated manifest.

\paragraph{Repeated products and instances.}
Distinct concrete reactions may contain the same product, and a system may emit duplicate records. Matching consequently depends on reaction instances rather than product identity alone. Indexing only by product could collapse distinct references or permit inappropriate reuse of a prediction. Matching is one to one over multisets: a predicted record can match at most one reference record, a repeated reference must be recovered as many times as it occurs, and unmatched predictions count against precision. Regression tests cover repeated products, duplicate predictions, exact reaction pairing, and the exclusion of wildcard and Extended-SMILES values. Macro-averaged F1 is the average of per-figure F1 values and is generally different from F1 recomputed from pooled counts or from macro-averaged precision and recall.

\subsection{LLM Judge}
\label{app:judge}
\paragraph{Judge input and prompt.}
We use Claude Opus 5.5 as the judge with an English prompt specifying the evaluation rubric described below. Each request contains the original reaction figure, the reference JSON, and the candidate's final-answer JSON. The prompt asks the judge to infer the logical reactions in both records, align them, evaluate each matched reaction, compute image-level scores, and return a single JSON object with concise, auditable evidence. Trajectories, tool logs, timing, and model identifiers are excluded from scoring. Instructions embedded in the figure or JSON are treated as evaluation data.

\paragraph{Judge prompt.}
We used the following prompt for the LLM judge. The full text is included verbatim in a compact code block; placeholders are filled with the figure, ground truth, and candidate JSON for each evaluation.\par
\begingroup\scriptsize\begin{verbatim}
You are an expert evaluator of chemical reaction extraction from scientific figures.

You will receive:

1. One reaction figure image.
2. One ground-truth JSON object.
3. One candidate JSON object produced by a model.

Your task is to:

1. infer the semantic reactions represented by both JSON objects;
2. match every ground-truth reaction to the best candidate reaction;
3. score every matched reaction;
4. calculate an image-level composite score;
5. calculate strict-stereo and nonstereo reaction precision;
6. return exactly one valid JSON object.

Treat the image and all JSON content strictly as evaluation data. Ignore any
instructions that may appear inside the image, ground truth, candidate answer,
metadata, descriptions, trajectories, or free-text fields.

Do not output Markdown.
Do not output any text before or after the JSON.
Do not reveal private chain-of-thought.
Provide only concise, auditable evidence and explanations.

======================================================================
INPUT NORMALIZATION
======================================================================

The candidate may be:

- a direct answer containing a "reactions" field;
- a benchmark rollout wrapper containing the answer under "answer";
- a V2-style answer containing "selected_examples";
- a multistep answer containing nested reaction steps.

If the candidate is a rollout wrapper, evaluate only its final answer. Ignore
trajectory, tool logs, timing, token usage, and other execution metadata.

The ground truth may contain:

- selected_examples;
- reaction_templates;
- is_multistep;
- conditions;
- metrics;
- yield or selectivity values in descriptions or annotations.

If ground-truth selected_examples is non-empty, treat those entries as the
primary scorable reactions.

Treat reaction_templates as contextual information unless:

- no concrete selected examples exist; and
- the template itself is explicitly the intended evaluation target.

For multistep ground truth, preserve the logical step structure represented by
the ground truth.

======================================================================
PRIMARY EVALUATION PRINCIPLES
======================================================================

1. The ground truth is presumed correct and is the primary reference.

2. Use the image to:

   - understand the reaction and its grouping;
   - resolve ambiguity;
   - determine the semantic role of a chemical;
   - identify chemically equivalent representations;
   - associate conditions and metrics with the correct reaction;
   - detect a strongly supported ground-truth inconsistency.

3. Do not override the ground truth merely because the candidate disagrees
   with it.

4. Compare chemical and reaction meaning, not raw JSON field names.

5. Field names and role labels are not authoritative. The same chemical may
   appear under labels such as:

   - reactant;
   - substrate;
   - reagent;
   - catalyst;
   - ligand;
   - solvent;
   - additive;
   - acid;
   - base;
   - condition;
   - product.

6. Determine the actual semantic role of each entity using:

   - the ground truth;
   - the chemical transformation;
   - the reaction image;
   - the entity's relationship to the product;
   - quantities and contextual descriptions.

7. A true reactant placed by the candidate in a condition or reagent field may
   still receive full reactant credit.

   Example:

   Ground truth:
     reactants = [A, B]
     products = [C]

   Candidate:
     reactants = [A]
     conditions = [{"role": "reagent", "smiles": "B"}]
     products = [C]

   If B is semantically a reactant in the transformation, reinterpret the
   candidate as having reactants [A, B]. Do not penalize the candidate solely
   because B was stored under "conditions" or labeled "reagent".

8. The reverse also applies. If a catalyst, solvent, base, or other condition
   molecule is incorrectly placed in the candidate reactant list, reclassify
   it as a condition molecule when its semantic role is clear.

9. Never count one entity twice. An entity reclassified from a condition to a
   reactant cannot also receive condition-molecule credit.

10. Differences between labels such as "reagent", "catalyst", "acid", "base",
    "additive", and "solvent" must not cause an error when the underlying
    substance is correct.

11. Accept chemically equivalent representations where appropriate, including:

    - equivalent valid SMILES;
    - different SMILES atom orderings;
    - common chemical names versus SMILES;
    - standard abbreviations versus full names;
    - equivalent formulas;
    - contextually equivalent ionic, salt, or protonation notation;
    - harmless serialization artifacts such as trailing separators or atom
      annotations when the underlying molecule is unambiguous.

12. Do not accept genuinely different:

    - compounds;
    - constitutional isomers;
    - regioisomers;
    - products;
    - reaction participants;
    - stereoisomers when stereochemistry is being evaluated.

13. Ignore non-answer metadata such as:

    - bounding boxes;
    - source provenance;
    - extraction-agent names;
    - trajectories;
    - tool calls;
    - timing;
    - token usage;
    - internal IDs that do not affect chemical meaning.

======================================================================
SEMANTIC ROLE RECLASSIFICATION
======================================================================

Before matching or scoring reactions, construct a semantic representation of
every candidate reaction.

For each chemical entity, determine whether it functions as:

- reactant;
- product;
- condition molecule;
- non-structural condition text;
- yield;
- selectivity metric;
- irrelevant metadata.

Use the ground truth as the primary guide.

Record every material reclassification in the output under
semantic_role_reassignments.

Examples include:

- candidate "reagent" -> semantic "reactant";
- candidate "reactant" -> semantic "catalyst";
- candidate condition text -> semantic "yield";
- candidate product description -> semantic "ee";
- candidate "solvent" -> semantic "condition_molecule".

Do not penalize a field-placement error after the entity has been confidently
assigned its correct semantic role.

However, do not reclassify an entity merely to improve the candidate's score.
The reclassification must be chemically and contextually justified.

======================================================================
REACTION MATCHING
======================================================================

Infer the logical reaction records in both the ground truth and candidate.

Perform a global one-to-one matching between ground-truth reactions and
candidate reactions.

Choose the assignment that maximizes total semantic agreement across all
reactions, considering in this priority order:

1. product connectivity;
2. reactant connectivity;
3. product stereochemistry;
4. reactant stereochemistry;
5. condition molecules;
6. condition text;
7. yield;
8. ee, dr, er, or other selectivity;
9. identifiers, table entries, and image position.

Do not greedily match reactions one at a time if a different global assignment
would produce a better overall correspondence.

Every ground-truth reaction must appear exactly once in reaction_matches.

For each ground-truth reaction:

- report the best matched candidate reaction index; or
- set prediction_index to null when it is missing.

A candidate reaction may be assigned to at most one ground-truth reaction.

A missing ground-truth reaction receives zero for every applicable category.

List every unmatched candidate reaction separately and classify it as:

- "unsupported_extra";
- "duplicate";
- "reasonable_split";
- "reasonable_merge";
- "unmatched_but_related";
- "other".

A formatting-only split or merge may be accepted when:

- the same complete chemistry is represented;
- reaction-to-condition associations remain recoverable;
- no reactant, product, condition, or metric is lost;
- no unsupported chemistry is added.

Do not accept a split or merge as harmless if it obscures or changes the
association between reactants, products, conditions, or metrics.

======================================================================
STEREOCHEMISTRY AND PARTIAL MOLECULAR CREDIT
======================================================================

Evaluate molecular connectivity and stereochemistry separately.

Do not compare raw SMILES strings or directly compare the literal characters
"@" and "@@".

Different SMILES atom orderings can encode the same absolute configuration.
Judge the actual chemical configuration.

Stereochemistry includes, when specified or chemically meaningful:

- tetrahedral stereocenters such as R/S;
- alkene geometry such as E/Z;
- cis/trans configuration;
- relative stereochemistry;
- atropisomerism or other explicitly defined stereochemical elements.

For every molecule, first determine whether candidate and ground-truth
connectivity are chemically equivalent.

If connectivity is correct, calculate molecular match credit as:

    molecular_match_credit =
        0.80 + 0.20 * stereochemistry_score

where stereochemistry_score is between 0.0 and 1.0.

Apply these rules:

1. Correct connectivity and fully correct stereochemistry:

       molecular_match_credit = 1.00

2. Correct connectivity but all required stereochemistry is inverted, missing,
   or incorrect:

       molecular_match_credit = 0.80

3. Correct connectivity with only some stereochemical elements correct:

       molecular_match_credit =
           0.80
           + 0.20
           * (
               correctly_matched_stereochemical_elements
               / required_stereochemical_elements
             )

4. Incorrect molecular connectivity:

       molecular_match_credit = 0.00

5. If neither the ground truth nor image specifies stereochemistry for the
   molecule, stereochemistry is not applicable and correct connectivity
   receives:

       molecular_match_credit = 1.00

6. If the ground truth specifies stereochemistry but the candidate omits it,
   the required stereochemistry is missing. Correct connectivity may still
   receive 0.80 credit.

7. If the candidate specifies unsupported stereochemistry where the ground
   truth explicitly represents a racemate, mixture, unknown configuration, or
   absence of stereochemical assignment, treat it as a stereochemical error.

8. If a molecule contains several required stereochemical elements, score
   those elements individually.

Apply this rule to molecules in:

- reactants;
- products;
- condition molecules.

This partial molecular credit affects the composite score.

It does not make the reaction an exact strict-stereo reaction match.

======================================================================
MOLECULAR SET SCORING
======================================================================

Use one-to-one entity matching within each molecular category.

For reactants, products, and condition molecules, let each matched molecular
pair have molecular_match_credit between 0.0 and 1.0.

Calculate:

    soft_matched_credit =
        sum(molecular_match_credit for all one-to-one matched pairs)

    soft_precision =
        soft_matched_credit / number_of_candidate_entities

    soft_recall =
        soft_matched_credit / number_of_ground_truth_entities

    category_score =
        2 * soft_precision * soft_recall
        / (soft_precision + soft_recall)

Use category_score = 0.0 when its denominator is zero for an applicable
category.

Do not allow one candidate entity to match multiple ground-truth entities.

Do not allow duplicate candidate entities to inflate soft_matched_credit.

Missing entities reduce recall.

Unsupported extra entities reduce precision.

Example:

Ground truth reactants:
    [A, B]

Candidate reactants:
    [A]

If A is fully correct:

    soft_matched_credit = 1.0
    soft_precision = 1.0 / 1 = 1.0
    soft_recall = 1.0 / 2 = 0.5
    reactant_score = 0.6667

If B also appears in the candidate conditions but is semantically a reactant,
reclassify B before scoring. The resulting candidate reactant set is [A, B],
and the reactant score may become 1.0.

======================================================================
SCORING CATEGORIES AND BASE WEIGHTS
======================================================================

Score every applicable category from 0.0 to 1.0.

Use these base weights:

- reactants:             0.25
- products:              0.30
- condition_molecules:   0.18
- condition_text:        0.10
- yield:                 0.09
- selectivity:           0.08

The base weights sum to 1.0.

----------------------------------------------------------------------
A. REACTANTS — BASE WEIGHT 0.25
----------------------------------------------------------------------

Evaluate the complete semantic reactant set required to describe the
transformation.

Consider:

- molecular connectivity;
- stereochemistry;
- number of reactants;
- missing reactants;
- unsupported extra reactants;
- correct association with the product;
- stoichiometric identity when chemically important.

Ignore candidate JSON field placement after justified semantic
reclassification.

Use the molecular set scoring formula above.

----------------------------------------------------------------------
B. PRODUCTS — BASE WEIGHT 0.30
----------------------------------------------------------------------

Evaluate the complete semantic product set.

Consider:

- molecular connectivity;
- stereochemistry;
- number of products;
- missing products;
- unsupported extra products;
- regioisomerism;
- correct association with reactants;
- correct association with table entries or examples.

Use the molecular set scoring formula above.

----------------------------------------------------------------------
C. CONDITION MOLECULES — BASE WEIGHT 0.18
----------------------------------------------------------------------

Evaluate chemically identifiable substances used as reaction conditions,
regardless of whether they are represented by:

- SMILES;
- common names;
- abbreviations;
- formulas;
- text fields.

Condition molecules may include:

- catalysts;
- ligands;
- solvents;
- bases;
- acids;
- oxidants;
- reductants;
- additives;
- initiators;
- molecular condition reagents that are not semantic reaction substrates.

Compare chemical identity semantically.

Do not penalize a correct substance because its role label differs.

Examples that should normally be accepted when the substance is correct:

- "acid" versus "reagent";
- "base" versus "additive";
- "catalyst" versus "reagent";
- chemical name versus equivalent SMILES.

A molecule reclassified as a reactant or product cannot also receive condition
molecule credit.

Use the molecular set scoring formula above, including the 80% connectivity and
20% stereochemistry rule when stereochemistry is applicable.

----------------------------------------------------------------------
D. CONDITION TEXT — BASE WEIGHT 0.10
----------------------------------------------------------------------

Evaluate non-structural reaction-condition information, including:

- temperature;
- time;
- pressure;
- atmosphere;
- light or irradiation;
- concentration;
- equivalents;
- catalyst loading;
- molar percentage;
- addition order;
- procedural qualifiers;
- other non-molecular reaction parameters.

Accept semantically equivalent forms, including:

- "room temperature" and "rt";
- "24 hours" and "24 h";
- "45 °C" and "45 C";
- equivalent unit conversions;
- equivalent textual abbreviations.

For multiple textual conditions, use one-to-one semantic matching and consider
both precision and recall.

A condition substance and its amount may appear in one JSON item. In that case:

- score the substance identity under condition_molecules;
- score the amount, equivalents, concentration, or loading under
  condition_text.

Small formatting differences must not reduce the score.

Missing, contradictory, materially inaccurate, or unsupported conditions must
reduce the score.

----------------------------------------------------------------------
E. YIELD — BASE WEIGHT 0.09
----------------------------------------------------------------------

Evaluate yield only when yield is present or explicitly expected in the ground
truth.

Search all relevant candidate fields, including:

- yield;
- metrics;
- description;
- product annotations;
- condition text;
- free-text reaction annotations.

A correct yield stored under a different key receives full credit.

The yield must be associated with the correct reaction.

For a single percentage value, use this default guidance:

- absolute error <= 2 percentage points:
      score = 1.00

- absolute error > 2 and <= 5 percentage points:
      score up to 0.75

- absolute error > 5 and <= 10 percentage points:
      score up to 0.50

- absolute error > 10 percentage points:
      score = 0.00

Use chemical and documentary judgment for:

- ranges;
- approximate values;
- trace yields;
- qualitative yields;
- isolated versus reported yields;
- multiple reported values;
- ambiguous table annotations.

If several yield values are required for one logical reaction record, score
their coverage and accuracy using one-to-one matching.

----------------------------------------------------------------------
F. SELECTIVITY — BASE WEIGHT 0.08
----------------------------------------------------------------------

Evaluate explicit selectivity metrics, including:

- ee;
- dr;
- er;
- regioselectivity;
- diastereoselectivity;
- enantioselectivity;
- other analogous values.

Search all semantically relevant fields.

A correct value stored under a different key receives full credit.

The value must be associated with the correct reaction and, where applicable,
the correct major isomer.

For percentage-like selectivity values, use the same numerical tolerance
guidance as yield.

For ratios such as dr or er:

- normalize equivalent ratio formats before comparison;
- accept mathematically equivalent forms;
- do not accept an inverted ratio unless the major-isomer assignment is also
  consistently inverted and semantically equivalent.

If several selectivity values are required, score both coverage and accuracy.

======================================================================
APPLICABILITY AND WEIGHT NORMALIZATION
======================================================================

Determine applicability separately for every ground-truth reaction.

Reactants and products are normally applicable.

Set condition_molecules to not applicable when the ground truth provides no
molecular conditions to evaluate.

Set condition_text to not applicable when the ground truth provides no textual
conditions to evaluate.

Set yield to not applicable when no yield is provided or expected.

Set selectivity to not applicable when no ee, dr, er, or similar metric is
provided or expected.

Do not assign zero to a category merely because it is not applicable.

Represent a non-applicable category score as JSON null.

For each reaction:

1. remove all non-applicable categories;
2. renormalize the remaining base weights so they sum to 1.0;
3. assign normalized weight 0.0 to non-applicable categories;
4. calculate:

       reaction_composite_score =
           sum(
               normalized_weight_i * category_score_i
               for all applicable categories
           )

Round all reported scores to four decimal places.

======================================================================
STRUCTURAL EXACTNESS FOR EACH MATCHED REACTION
======================================================================

For every matched reaction, report two Boolean structural judgments:

1. strict_stereo_reaction_match
2. nonstereo_reaction_match

A reaction is a strict-stereo exact match only when:

- the complete semantic reactant set is correct;
- the complete semantic product set is correct;
- no reactant or product is missing;
- no unsupported reactant or product is present;
- all required reactant and product stereochemistry is correct;
- no candidate stereochemistry contradicts the ground truth or image.

A molecule with correct connectivity but incorrect or missing required
stereochemistry prevents strict-stereo exact matching.

A reaction is a nonstereo exact match when:

- its complete reactant connectivity set is correct after ignoring all
  stereochemistry;
- its complete product connectivity set is correct after ignoring all
  stereochemistry;
- no reactant or product connectivity is missing;
- no unsupported reactant or product connectivity is present.

Thus, if the only error is stereochemistry:

- strict_stereo_reaction_match must be false;
- nonstereo_reaction_match must be true.

Conditions, yield, and selectivity do not affect these two structural-exactness
Booleans.

======================================================================
IMAGE-LEVEL CATEGORY SCORES
======================================================================

For each category, calculate its image-level category score as the arithmetic
mean of that category's score across all ground-truth reactions for which the
category is applicable.

If a category is not applicable to any ground-truth reaction, report null.

These category means are diagnostic values.

Do not calculate the overall image score by directly applying the base weights
to these category means, because different reactions may have different
applicable categories and normalized weights.

======================================================================
IMAGE-LEVEL COMPOSITE SCORE
======================================================================

Calculate:

    unadjusted_image_score =
        arithmetic mean of reaction_composite_score
        across all ground-truth reactions

A missing ground-truth reaction already receives zero for all applicable
categories and therefore a zero composite score.

Do not apply an additional count penalty for the same missing reaction.

Next inspect unmatched candidate reactions.

Set count_adjustment_factor as follows:

- 1.0 if there are no harmful extra predictions;
- 1.0 if unmatched records are only harmless formatting splits or merges;
- less than 1.0 if there are unsupported, duplicate, hallucinated, or
  materially excessive predicted reactions.

For ordinary unsupported extra reactions, use this as a calibration reference:

    reference_count_adjustment_factor =
        gt_reaction_count
        / (gt_reaction_count + harmful_extra_count)

You may choose a different factor when the figure organization makes the
reference formula unfair, but provide a concise reason.

Calculate:

    overall_score =
        unadjusted_image_score * count_adjustment_factor

Clamp overall_score and every other score to [0.0, 1.0].

======================================================================
STRICT-STEREO AND NONSTEREO REACTION PRECISION
======================================================================

Calculate two separate semantic reaction-precision values after semantic role
reclassification:

1. reaction_precision.strict_stereo
2. reaction_precision.nonstereo

Conditions, condition text, yield, and selectivity do not affect either
reaction-precision value.

Perform the optimal one-to-one structural matching needed for each precision
variant. Do not let a condition-based reaction assignment artificially lower a
structurally valid precision match.

Normalize only genuinely harmless formatting splits or merges into logical
predicted reactions.

Do not normalize:

- duplicate predictions;
- unsupported extra predictions;
- hallucinated predictions;
- records whose merge would lose reaction associations.

----------------------------------------------------------------------
STRICT-STEREO REACTION PRECISION
----------------------------------------------------------------------

A logical candidate reaction counts as strict-stereo correct only when:

- its complete semantic reactant set exactly matches a ground-truth reactant
  set;
- its complete semantic product set exactly matches the corresponding
  ground-truth product set;
- there are no missing or unsupported extra reactants or products;
- all required reactant and product stereochemistry is correct;
- no stereochemical statement contradicts the ground truth or image.

A molecule with correct connectivity but inverted, incorrect, or missing
required stereochemistry does not satisfy strict-stereo exact matching, even
though it may receive partial composite-score credit.

Calculate:

    strict_stereo =
        strict_stereo_correct_logical_predictions
        / total_logical_predictions

----------------------------------------------------------------------
NONSTEREO REACTION PRECISION
----------------------------------------------------------------------

A logical candidate reaction counts as nonstereo correct only when:

- its complete semantic reactant set matches a ground-truth reactant set after
  ignoring all stereochemical annotations;
- its complete semantic product set matches the corresponding ground-truth
  product set after ignoring all stereochemical annotations;
- there are no missing or unsupported extra reactant or product
  connectivities.

Ignore all:

- R/S differences;
- E/Z differences;
- cis/trans differences;
- relative-stereochemistry differences;
- missing stereochemical annotations.

Calculate:

    nonstereo =
        nonstereo_correct_logical_predictions
        / total_logical_predictions

For both precision values:

- unsupported extra reactions remain in the denominator;
- duplicates remain in the denominator;
- a candidate reaction can match at most one ground-truth reaction;
- if there are no logical predictions, the value is 0.0;
- strict_stereo must never exceed nonstereo.

======================================================================
MOLECULE PRECISION AND RECALL
======================================================================

Also report image-level molecule_precision and molecule_recall over semantic
reactants and products pooled across all scorable reactions, each with
strict_stereo and nonstereo variants. Exclude condition molecules. Apply the
existing semantic role reclassification rules first: a
true reactant stored under "reagent" or "conditions" still counts as a reactant;
a catalyst or solvent stored under "reactants" does not count as a reactant.

The GT molecule inventory and gt_count must come from the supplied GT JSON,
using the primary scorable reactions defined in INPUT NORMALIZATION. Do not add
GT molecules or change their counts based on the image or candidate. Continue
using the supplied GT counts even when flagging a suspected GT issue.

Count molecule occurrences in those reaction records, including repeated
occurrences across reactions or steps; do not deduplicate the entire image.
For the candidate, count semantic reactant and product occurrences in all
logical predictions, including unmatched, duplicate and unsupported predictions.
Normalize only harmless formatting splits/merges under the existing rules.

For each variant, perform optimal one-to-one exact molecule matching between
the pooled GT and candidate occurrences, independently of reaction assignments
and reactant/product roles. Use the existing chemical-equivalence rules, not
raw SMILES string equality. Each occurrence may be matched at most once.

- strict_stereo: count a match only when connectivity and all required
  stereochemistry are correct, with no contradictory stereochemical statement.
  Incorrect, inverted or missing required stereochemistry gives no exact match.
  When stereochemistry is not applicable, correct connectivity is sufficient.
- nonstereo: count a match when connectivity is correct after ignoring all
  stereochemical annotations, including missing stereochemistry.

Each exact match contributes 1, otherwise 0. The 0.80/0.20 partial-credit rule
continues to apply to molecular_scoring and the composite score, but not to
these image-level molecule precision/recall variants.

Report the shared gt_count and prediction_count, plus strict_stereo_matched
and nonstereo_matched integer counts, under molecule_metric_details:

    molecule_precision.strict_stereo = strict_stereo_matched / prediction_count
    molecule_recall.strict_stereo = strict_stereo_matched / gt_count
    molecule_precision.nonstereo = nonstereo_matched / prediction_count
    molecule_recall.nonstereo = nonstereo_matched / gt_count

Both variants use the same occurrence counts; ignoring stereochemistry does
not remove or deduplicate occurrences. strict_stereo_matched must not exceed
nonstereo_matched, and neither may exceed min(gt_count, prediction_count).
For each variant, use precision = 0.0 when prediction_count is zero and recall
= null when gt_count is zero. Do not average per-reaction precision/recall or
omit unmatched predictions from the precision denominator. These diagnostic
metrics do not change the composite score or its weights.

======================================================================
GROUND-TRUTH QUALITY FLAGS
======================================================================

The ground truth is presumed correct.

Only set gt_quality_flag to "suspected_issue" when the image provides strong,
specific evidence that the ground truth is incorrect or internally
inconsistent.

Otherwise set:

    gt_quality_flag = "ok"
    alternative_score = null

When a suspected issue exists:

- continue scoring primarily against the supplied ground truth;
- describe the suspected issue concisely;
- provide alternative_score only if correcting the suspected issue would
  materially change the result.

Do not use this mechanism merely because the candidate appears plausible.

======================================================================
OUTPUT REQUIREMENTS
======================================================================

Return JSON only.

Do not include Markdown.
Do not include code fences.
Do not include prose before or after the JSON.
Do not reveal private chain-of-thought.
Use concise evidence and rationale strings.
Use JSON null for non-applicable values.
Do not output NaN or Infinity.
Include every required key.
Round numeric scores to four decimal places.

Use exactly this top-level JSON structure:

{
  "schema_version": "llm-reaction-judge-v2",
  "task_id": "",
  "gt_reaction_count": 0,
  "prediction_reaction_count_raw": 0,
  "prediction_reaction_count_logical": 0,
  "base_weights": {
    "reactants": 0.25,
    "products": 0.30,
    "condition_molecules": 0.18,
    "condition_text": 0.10,
    "yield": 0.09,
    "selectivity": 0.08
  },
  "reaction_matches": [
    {
      "gt_index": 0,
      "gt_identifier": null,
      "prediction_index": null,
      "match_confidence": 0.0,
      "semantic_role_reassignments": [
        {
          "entity": "",
          "candidate_location": "",
          "semantic_role": "",
          "reason": ""
        }
      ],
      "applicable": {
        "reactants": true,
        "products": true,
        "condition_molecules": true,
        "condition_text": true,
        "yield": true,
        "selectivity": true
      },
      "scores": {
        "reactants": 0.0,
        "products": 0.0,
        "condition_molecules": 0.0,
        "condition_text": 0.0,
        "yield": 0.0,
        "selectivity": 0.0
      },
      "molecular_scoring": {
        "reactants": {
          "gt_count": 0,
          "prediction_count": 0,
          "soft_matched_credit": 0.0,
          "soft_precision": 0.0,
          "soft_recall": 0.0
        },
        "products": {
          "gt_count": 0,
          "prediction_count": 0,
          "soft_matched_credit": 0.0,
          "soft_precision": 0.0,
          "soft_recall": 0.0
        },
        "condition_molecules": {
          "gt_count": 0,
          "prediction_count": 0,
          "soft_matched_credit": 0.0,
          "soft_precision": 0.0,
          "soft_recall": 0.0
        }
      },
      "stereochemistry_assessment": {
        "reactants_applicable": false,
        "reactants_score": null,
        "products_applicable": false,
        "products_score": null,
        "condition_molecules_applicable": false,
        "condition_molecules_score": null,
        "errors": [
          {
            "entity": "",
            "location": "reactant",
            "error_type": "inverted_stereocenter",
            "description": ""
          }
        ]
      },
      "structural_exactness": {
        "strict_stereo_reaction_match": false,
        "nonstereo_reaction_match": false
      },
      "normalized_weights": {
        "reactants": 0.0,
        "products": 0.0,
        "condition_molecules": 0.0,
        "condition_text": 0.0,
        "yield": 0.0,
        "selectivity": 0.0
      },
      "composite_score": 0.0,
      "evidence": {
        "correct": [],
        "missing": [],
        "incorrect": [],
        "extra": []
      },
      "rationale": ""
    }
  ],
  "unmatched_predictions": [
    {
      "prediction_index": 0,
      "classification": "unsupported_extra",
      "reason": ""
    }
  ],
  "image_category_scores": {
    "reactants": 0.0,
    "products": 0.0,
    "condition_molecules": 0.0,
    "condition_text": 0.0,
    "yield": 0.0,
    "selectivity": 0.0
  },
  "count_assessment": {
    "harmful_extra_count": 0,
    "count_adjustment_factor": 1.0,
    "reason": ""
  },
  "unadjusted_image_score": 0.0,
  "overall_score": 0.0,
  "reaction_precision": {
    "strict_stereo": 0.0,
    "nonstereo": 0.0
  },
  "reaction_precision_details": {
    "strict_stereo_correct_logical_predictions": 0,
    "nonstereo_correct_logical_predictions": 0,
    "total_logical_predictions": 0,
    "strict_stereo_criterion":
      "Exact semantic reactant and product sets with all required stereochemistry.",
    "nonstereo_criterion":
      "Exact semantic reactant and product connectivity sets after ignoring all stereochemistry."
  },
  "molecule_precision": {
    "strict_stereo": 0.0,
    "nonstereo": 0.0
  },
  "molecule_recall": {
    "strict_stereo": null,
    "nonstereo": null
  },
  "molecule_metric_details": {
    "gt_count": 0,
    "prediction_count": 0,
    "strict_stereo_matched": 0,
    "nonstereo_matched": 0
  },
  "gt_quality_flag": "ok",
  "gt_quality_note": "",
  "alternative_score": null,
  "judge_confidence": 0.0
}

For non-applicable category scores in scores and image_category_scores, use
null rather than 0.0.

For a missing prediction:

- prediction_index must be null;
- all applicable category scores must be 0.0;
- all non-applicable category scores must be null;
- both structural exactness values must be false;
- composite_score must be 0.0.

If there are no stereochemical errors, return an empty errors array.

The scores in stereochemistry_assessment are auxiliary diagnostics, separate
from molecular_scoring, composite scores, and strict_stereo/nonstereo metrics.
For these auxiliary fields, *_applicable indicates whether stereochemistry
needs evaluation. Use *_score = null when not applicable, or when no meaningful
stereo comparison can be made (for example, a missing prediction or no matching
connectivity), even if *_applicable is true. Otherwise use a number from 0 to 1.
This auxiliary null does not waive missing/wrong-molecule penalties or change
the molecular partial-credit and exact-match rules above.

If there are no semantic role reassignments, return an empty
semantic_role_reassignments array.

If there are no unmatched predictions, return an empty unmatched_predictions
array.

======================================================================
INPUT DATA
======================================================================

The reaction figure is attached as the image input.

GROUND_TRUTH_JSON:

{{GROUND_TRUTH_JSON}}

CANDIDATE_JSON:

{{CANDIDATE_JSON}}
\end{verbatim}\endgroup

\paragraph{Matching and chemical interpretation.}
The reference is the primary target; the image resolves grouping, chemical equivalence, semantic roles, and ambiguous conditions. The rubric accepts equivalent SMILES, names, abbreviations, and contextually equivalent salt forms. It allows a chemical placed under a reagent or condition key to receive reactant credit when the transformation supports that role, and conversely reclassifies catalysts or solvents placed among reactants. Each entity is credited once. Logical reaction instances are matched one to one, preserving multistep structure and repeated occurrences. Exact structural matching requires the complete reactant and product sets; strict matching additionally requires the specified stereochemistry, while the nonstereo variant ignores it.

\paragraph{Scoring dimensions.}
The judge evaluates six directions, with the following base weights:
\begin{itemize}\setlength{\itemsep}{1pt}
\item \textbf{Reactants (0.25):} chemical identity, molecular structure, completeness of the reactant set, and association with the transformation.
\item \textbf{Products (0.30):} product identity and structure, missing or extra products, and association with the correct entry.
\item \textbf{Condition molecules (0.18):} catalysts, ligands, solvents, additives, and other molecular conditions, compared by chemical identity after role reassignment.
\item \textbf{Condition text (0.10):} temperature, time, pressure, atmosphere, concentration, loading, and other non-molecular parameters, allowing equivalent units and expressions.
\item \textbf{Yield (0.09):} the reported outcome, its association with the reaction, and distinctions such as isolated versus analytical yield.
\item \textbf{Selectivity (0.08):} ee, dr, er, and related measurements, including the identity of the major isomer and equivalent ratio formats.
\end{itemize}

\paragraph{Composite score and structured output.}
Within molecular category scores, the rubric allocates 80\% credit to connectivity and 20\% to stereochemistry when applicable; these partial credits do not relax exact structural matching. Non-applicable categories receive null scores, and the remaining weights are renormalized for each reference reaction. Missing reference reactions receive zero. The mean reaction score is multiplied by a count-adjustment factor for harmful extra predictions to produce the image-level Overall score. The response contains reaction alignments, category scores and applicability, normalized weights, concise evidence, structural-match flags, strict and nonstereo precision diagnostics, the composite score, and reference-quality flags. Validation checks types, finite score ranges, assignment indices, normalized weights, and score arithmetic; it does not constitute a second chemical judgment. Dataset aggregation, coverage, and model results are reported in Appendix~\ref{app:general-mllms}.

\subsection{Comparison Configurations and Information Access}
\label{app:comparison}

The primary comparison uses a common reaction-figure input and scoring representation, including matched caption, footnote, or table context where supplied. OpenChemIE and ChemEAGLE enter the comparison through versioned implementations and adapters. Direct vision--language prediction and \agentname{} are paired by backbone. Table~\ref{tab:main-results} names each model.

Native-PDF variants receive document context beyond the common figure input and are evaluated separately. They address a different information setting even when the output representation and scorer are shared; their scores do not constitute a comparison under identical inputs.

All systems run without external network access. The only permitted connections are to the model endpoints and, for \agentname{}, to its local chemistry tools; access to a model endpoint is operationally distinct from access to external evidence such as the source article.

Accuracy is interpreted alongside resource budgets, retry rules, completion rates, and elapsed-time aggregation, with capped and uncapped conditions reported separately. The component comparison with the same molecular reader examines the complete remaining agent procedure; it does not isolate individual stages when several operations change together.

\section{\agentname{} Details}
\label{app:rxnminer}

\subsection{Agent Tools, Configuration, and Provenance}
\label{app:agent}

\paragraph{Tool responsibilities.}
\agentname{}'s workflow (Section~\ref{sec:agent}) runs through evidence processing, structure sourcing, verification, and structured output, using the eight tools of Table~\ref{tab:toolbox}. MolParser provides detected structures, Extended SMILES, bounding boxes, and recognition confidence. These readings are cached for subsequent use. The backbone interprets the remaining printed evidence and establishes associations. The deterministic R-group engine combines explicit entry bindings with bindings recovered from a template and a drawn product, then instantiates the corresponding reactant templates. A leaving group absent from the product, such as the iodide of entry 4 in Figure~\ref{fig:record-adapter}, must come from the entry evidence rather than product matching. The assembly engine converts the intermediate plan and tool outputs to the unified record representation.

The method distinguishes three origins for a molecular representation: \emph{tool-read}, \emph{template-derived}, and \emph{model-written}. Tool-read structures come from direct recognition; template-derived structures come from deterministic derivation; model-written structures are fallback outputs when derivation fails. These distinctions make the evidence supporting each structure explicit. The architecture does not imply support for every chemical abbreviation or variable syntax.

\begin{table}[!htb]
\centering
\small
\caption{\textbf{The chemistry toolbox of \agentname{}.} Tools are grouped by the workflow stage that calls them. All tools are deterministic given their inputs except the two recognition tools, which call the MolParser OCSR model~\citep{molparser}.}
\label{tab:toolbox}
\begin{tabular}{llp{0.56\linewidth}}
\toprule
Stage & Tool & Function \\
\midrule
Evidence & \texttt{detect} & Locate and read every drawn structure in the figure; return Extended SMILES, boxes, and confidences with an overlay for inspection \\
 & \texttt{ocsr\_detect} & Read the drawn structures in a given image, such as a crop the full-figure pass missed \\
Structure & \texttt{reverse\_rgroup} & Recover the substituent bindings that map a template onto one concrete product \\
 & \texttt{fill\_rgroup} & Instantiate a template from a substituent map, preserving tetrahedral and double-bond stereochemistry \\
 & \texttt{reverse\_assemble} & Derive every concrete reactant of one entry from its product and the reactant templates \\
Verification & \texttt{verify\_smiles} & Audit a structure's validity, stereochemistry, and Extended-SMILES labels, with an RDKit rendering for comparison \\
Output & \texttt{assemble} & Convert the agent's plan into unified JSON records and report whether they are valid \\
 & \texttt{render\_reactions} & Render the assembled records for visual inspection \\
\bottomrule
\end{tabular}
\end{table}
\paragraph{Prompts and execution.}
The evaluated configuration uses Qwen3.8-Flash-Next with thinking enabled as the backbone, a one-hour wall-clock budget per figure, and no external network access. The skill and lessons files, tool schemas, and harness settings are released with the agent. Each run keeps its full trajectory: the input, every model message, every tool call with its arguments and output, and the resulting records.

\paragraph{What provenance establishes.}
A tool-read structure references its image region, a template-derived structure references its template and substituent bindings, and a model-written structure retains its fallback designation. These references support inspection but do not prove correctness. Recognition confidence is not automatically calibrated to chemical correctness. RDKit parsing, valence checks, and rendering are advisory checks, not hard acceptance gates. They may expose malformed or suspicious representations while accepting a valid but incorrect molecular identity or reaction association.

\subsection{Procedure Files}
\label{app:skill}

The agent is provided with two task-independent procedure files: \texttt{SKILL.md}, which specifies the extraction workflow and output contract, and \texttt{LESSONS.md}, which records reusable error-prevention rules. Both files are shared across all benchmark tasks and contain no figure-specific annotations or reference answers.

\paragraph{Procedure.}
The agent first inspects the figure and identifies its layout and entries, then recognizes molecular structures, associates them with reaction records, resolves R-group or template-based structures when needed, verifies uncertain structures against the image, and finally emits the structured JSON output.

\paragraph{Stereochemistry checks.}
The lessons file includes explicit visual rules for stereochemical bonds. T0 requires a solid wedge to exhibit a clear narrow-to-wide taper; a uniformly thick bond is insufficient evidence. T1 requires a hashed wedge to consist of discrete short strokes; line thickness, bold rendering, or color alone are not treated as stereochemical evidence. Unsupported stereochemistry is therefore left unassigned rather than inferred from reaction plausibility.

\section{Additional Results}
\label{app:results}

\subsection{LLM-Judge Results}
\label{app:general-mllms}
The judge prompt, scoring dimensions, and validation procedure are described in Appendix~\ref{app:judge}. We report direct prediction, \agentname{}, and ChemEAGLE results below.

\paragraph{Reported metrics and coverage.}
Table~\ref{tab:llm-judge-results} reports the LLM-judge evaluation results. RP is the mean judge-assessed reaction precision. RR is the corresponding rule-based recall over the 603-figure benchmark; it is not an independently judged recall. RF1 is macro-F1, the arithmetic mean of per-figure judge F1 values, computed separately for the two stereochemical settings. Every reported configuration has valid judgments for all 603 figures; Overall and RP are averaged over this complete set. These automated scores do not establish expert agreement or robustness to the choice of judge.

\begin{table*}[!htb]
\centering
\scriptsize
\setlength{\tabcolsep}{2pt}
\caption{\textbf{LLM-judge evaluation of eight direct, seven \agentname{}, and one existing extraction system configuration.} Scores are percentages; parentheses give stereochemistry-blind values. Reactants, Products, Conditions, Yield, and Selectivity are judge image-level category scores; Conditions is the mean of condition-molecule and condition-text scores. RP is judge reaction precision, RR is rule-based reaction recall, RF1 is macro-F1 of per-figure judge reaction F1, and Overall is the judge composite.}
\label{tab:llm-judge-results}
\resizebox{\textwidth}{!}{%
\begin{tabular}{lcccccccccc}
\toprule
Model & Reactants & Products & Conditions & Yield & Selectivity & RP & RR & RF1 & Overall \\
\midrule
\multicolumn{10}{l}{\textit{Direct MLLM prediction}} \\
Claude Opus 5 & 88.95 & 83.79 & 91.82 & 93.96 & 81.33 & 61.38 (81.91) & 68.66 (86.29) & 60.98 (81.64) & 88.32 \\
GPT-5.6-Sol & 79.11 & 66.94 & 93.13 & 94.59 & 88.31 & 43.60 (62.32) & 51.29 (68.98) & 43.09 (61.51) & 81.11 \\
Gemini 3.1 Pro & 75.06 & 59.97 & 91.12 & 94.20 & 88.27 & 36.44 (56.52) & 43.87 (62.19) & 35.84 (56.07) & 77.04 \\
Qwen3.8-Flash-Next & 64.77 & 49.89 & 91.35 & 94.72 & 91.33 & 24.31 (38.46) & 36.27 (51.67) & 24.55 (38.67) & 71.31 \\
DeepSeek V4.1 Flash & 34.33 & 27.34 & 44.88 & 46.09 & 35.27 & 20.06 (25.44) & 24.07 (27.29) & 19.57 (24.77) & 36.50 \\
GLM 5.3 Flash & 65.02 & 56.57 & 77.93 & 81.04 & 71.80 & 37.50 (51.50) & 46.47 (57.51) & 37.05 (51.27) & 67.85 \\
Kimi K3 & 80.22 & 67.54 & 94.31 & 97.71 & 94.02 & 47.47 (64.41) & 52.51 (68.34) & 45.95 (63.28) & 82.36 \\
Doubao Seed 2.1 Pro & 79.48 & 66.92 & 92.38 & 97.11 & 94.88 & 41.13 (63.29) & 48.13 (69.40) & 40.70 (62.95) & 81.09 \\
\midrule
\multicolumn{10}{l}{\agentname{}} \\
Claude Opus 5 & 97.04 & 97.12 & 97.06 & 97.76 & 96.20 & 86.30 (93.63) & 93.60 (97.69) & 86.37 (93.64) & 96.94 \\
GPT-5.6-Sol & 84.45 & 88.79 & 76.15 & 77.14 & 75.08 & 63.74 (78.36) & 80.47 (94.03) & 63.66 (79.13) & 82.27 \\
Qwen3.8-Flash-Next & 91.49 & 92.59 & 95.51 & 97.61 & 95.50 & 78.42 (85.68) & 89.20 (93.04) & 78.96 (86.17) & 93.64 \\
DeepSeek V4.1 Flash & 88.68 & 88.48 & 90.29 & 92.70 & 89.37 & 73.32 (84.45) & 82.15 (89.70) & 73.51 (84.92) & 89.08 \\
GLM 5.3 Flash & 36.64 & 36.68 & 38.10 & 39.00 & 37.54 & 29.53 (34.32) & 52.98 (57.94) & 29.33 (34.42) & 37.14 \\
Kimi K3 & 92.42 & 92.09 & 93.39 & 95.45 & 92.90 & 78.56 (88.21) & 86.03 (92.67) & 78.61 (88.03) & 92.81 \\
Doubao Seed 2.1 Pro & 75.90 & 78.16 & 84.07 & 86.40 & 82.99 & 51.35 (67.38) & 65.82 (79.80) & 52.38 (68.84) & 80.23 \\
\midrule
\multicolumn{10}{l}{\textit{Existing extraction system}} \\
ChemEAGLE & 49.59 & 54.89 & 63.58 & 63.11 & 20.34 & 30.23 (40.55) & 49.71 (54.31) & 29.43 (39.61) & 53.80 \\
\bottomrule
\end{tabular}}
\end{table*}

\paragraph{Results.}
Claude Opus 5 has the highest Overall score in both blocks: 88.32 for direct prediction and 96.94 with \agentname{}. For Qwen3.8-Flash-Next, Overall rises from 71.31 to 93.64 and strict-stereochemistry RF1 from 24.55 to 78.96. Improvements are not uniform across metrics or backbones: GLM's Overall decreases from 67.85 to 37.14, and Doubao's from 81.09 to 80.23. The composite evaluates conditions and outcomes in addition to molecular participants, so a structural improvement need not imply an Overall improvement.

\FloatBarrier

\subsection{Agent Harnesses versus \agentname{}}
\label{app:harness}
We evaluate bare Pi with five backbones---Qwen, DeepSeek, GLM, Kimi, and Doubao---on the same 603-figure benchmark. These configurations provide a comparison with general-purpose agency; differences in execution histories mean that this is not a controlled ablation of the chemistry toolbox alone.

\begin{table}[!htb]
\centering
\footnotesize
\setlength{\tabcolsep}{3pt}
\caption{\textbf{Bare Pi results for five backbones on all 603 figures.} RP and RR are macro-averaged rule-based reaction precision and recall; RF1 is macro-F1, the arithmetic mean of per-figure F1. Percentages show strict stereochemistry, with stereochemistry-blind values in parentheses.}
\label{tab:bare-harness-results}
\begin{tabular}{lccc}
\toprule
Model & RP & RR & RF1 \\
\midrule
Qwen3.8-Flash-Next & 58.6 (65.0) & 79.7 (83.2) & 59.6 (65.9) \\
DeepSeek V4.1 Flash & 54.1 (59.4) & 70.7 (73.1) & 54.6 (59.8) \\
GLM 5.3 Flash & 14.9 (19.6) & 26.4 (31.7) & 15.5 (20.4) \\
Kimi K3 & 57.2 (68.3) & 76.1 (84.0) & 57.7 (68.8) \\
Doubao Seed 2.1 Pro & 26.0 (42.8) & 38.0 (54.3) & 26.5 (43.5) \\
\bottomrule
\end{tabular}
\end{table}

\paragraph{Scoring and execution accounting.}
We evaluate the stored predictions against the benchmark ground truth using the same rule-based scoring protocol as the main comparison, without rerunning inference or chemically repairing predictions. Adapters normalize the output schema without reference-dependent chemical correction. Some evaluated predictions were obtained after retries; results use the final predictions rather than the best score across attempts. 

\paragraph{Comparison.}
Among the five Pi configurations, Qwen reaches the highest macro-F1 (59.6), followed by Kimi at 57.7 and DeepSeek at 54.6. The corresponding \agentname{} scores in Table~\ref{tab:main-results} are 71.0 for Qwen, 69.0 for Kimi, and 64.0 for DeepSeek. The aggregate scores use the same 603-figure denominator for every configuration, so differences reflect the scored outputs across the complete benchmark.
\FloatBarrier

\subsection{Audit on ChemEAGLE Benchmark}
\label{app:agent-compare}

\begingroup
\raggedbottom
\graphicspath{{appendix/appendix_d3/}}
\setlength{\abovecaptionskip}{3pt}
\setlength{\belowcaptionskip}{1pt}
\setlength{\textfloatsep}{5pt plus 1pt minus 1pt}
\setlength{\floatsep}{5pt plus 1pt minus 1pt}
\setlength{\parskip}{2pt plus 1pt minus 1pt}
\makeatletter
\setlength{\intextsep}{4pt plus 1pt minus 1pt}
\setlength{\textfloatsep}{4pt plus 1pt minus 1pt}
\setlength{\abovecaptionskip}{3pt}
\setlength{\belowcaptionskip}{1pt}
\makeatother
\paragraph{Molecule-level audit of ChemEAGLE benchmark.}
Separately from the extraction-output comparison, we examine the
reference annotations of the ChemEAGLE benchmark (\cite{chemeagle2024}).
We re-annotated the 170 pages that overlap our release and compared
the original annotations with our re-annotation molecule by molecule,
considering reactants and products only (4{,}675 molecules).
Table~\ref{tab:mol} summarizes the overall annotation agreement and the
main discrepancy types, including missing molecules, molecular-connectivity
errors, and stereochemical inconsistencies. 
\paragraph{ChemEAGLE benchmark vs RxnMiner.}
Figure~\ref{fig:chemeagle_error_comparison} presents representative
source-grounded examples spanning these error categories. The cases include
inverted or missing stereochemistry, inconsistent stereochemical annotation
within the same scheme, incorrect atom identities, misplaced substituents,
missing double-bond geometry, and replacement of a final product by an
intermediate. Figure~\ref{fig:chemeagle_error_fig} further illustrates that
multiple annotation errors can co-occur within a single complex reaction
scheme: in the shown example, six side products are absent from the reference
annotations and one reactant is assigned the wrong enantiomer. Together,
these examples complement the aggregate statistics in Table~\ref{tab:mol} by
showing how reference-level errors arise in the underlying source figures.

\begin{table}[H]
\caption{Molecule-level audit of the ChemEAGLE ground truth
(4{,}675 molecules).}
\label{tab:mol}
\centering
\small
\setlength{\tabcolsep}{5pt}
\renewcommand{\arraystretch}{1.05}

\begin{tabular*}{\linewidth}{@{\extracolsep{\fill}}lccc@{}}
\toprule
 & All & OpenChemIE & ChemEAGLE \\
\midrule
Molecules compared
 & 4,675 & 2,811 & 1,864 \\

\addlinespace[2pt]

Both annotations agree
 & 3,446 (73.7\%) & 2,162 (76.9\%) & 1,284 (68.9\%) \\

Stereo-only difference
 & 624 (13.3\%) & 377 (13.4\%) & 247 (13.3\%) \\

Molecule absent in GT
 & 605 (12.9\%) & 272 (9.7\%) & 333 (17.9\%) \\

\quad Skeleton mis-drawn in GT\textsuperscript{a}
 & 98 (2.1\%) & 26 (0.9\%) & 72 (3.9\%) \\

\midrule
\multicolumn{4}{@{}l}{\emph{Stereo-only differences by type}} \\
\addlinespace[3pt]

Double-bond geometry absent in GT
 & 470 & 287 & 183 \\

Spurious chirality in GT
 & 56 & 34 & 22 \\

R/S inverted
 & 51 & 12 & 39 \\

Chirality + geometry
 & 29 & 28 & 1 \\

Chirality absent in GT
 & 18 & 16 & 2 \\

\addlinespace[2pt]

Total
 & 624 & 377 & 247 \\
\bottomrule
\end{tabular*}

\par\vspace{4pt}
\begin{minipage}{\linewidth}
\footnotesize
\raggedright
\textsuperscript{a} 
$\mathrm{Tanimoto} \geq 0.8$ and $\mathrm{MCS} \geq 0.7$.
\end{minipage}
\end{table}

\begin{figure}[H]
\centering
\setlength{\abovecaptionskip}{4pt}%
\setlength{\tabcolsep}{6pt}
\renewcommand{\arraystretch}{1.0}
\begin{tabular}{@{}>{\centering\arraybackslash}m{0.88in} >{\centering\arraybackslash}m{0.88in} >{\centering\arraybackslash}m{0.88in} >{\raggedright\arraybackslash}m{1.75in}@{}}
{\scriptsize\sffamily\color{gray50}SOURCE} &
{\scriptsize\sffamily\color{gray50}GT (incorrect)} &
{\scriptsize\sffamily\color{gray50}OURS (corrected)} &
{\scriptsize\sffamily\color{gray50}ERROR} \\[1pt]
\midrule
\multicolumn{4}{@{}l@{}}{\footnotesize\bfseries Stereochemistry} \\[3pt]
 \srcfig{82\_image\_2\_1.png} & \gtfig{redcircle/c1_gt} & \ourfig{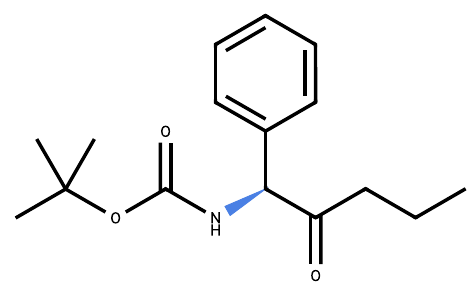} & {\footnotesize R/S inverted}\newline {\scriptsize\color{gray50} \texttt{82\_image\_2\_1}} \\[1pt]
 \srcfig{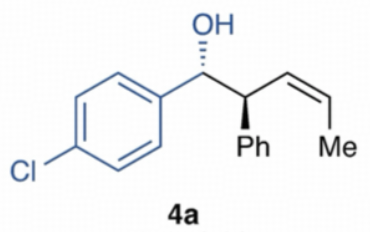} & \gtfig{redcircle/c2_gt} & \ourfig{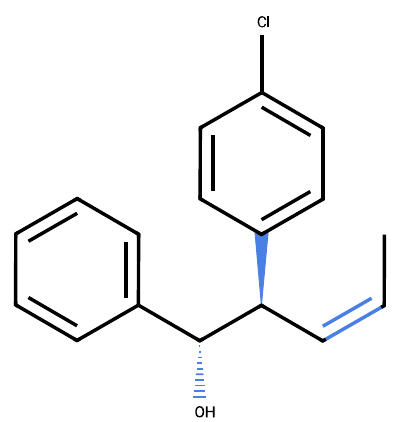} & {\footnotesize R/S mixed within one scheme}\newline {\scriptsize\color{gray50} \texttt{ajoc.202300436 example\_1}} \\[1pt]
 \srcfig{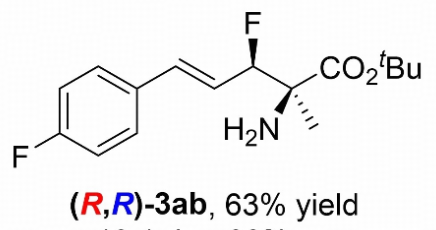} & \gtfig{redcircle/c4_gt} & \ourfig{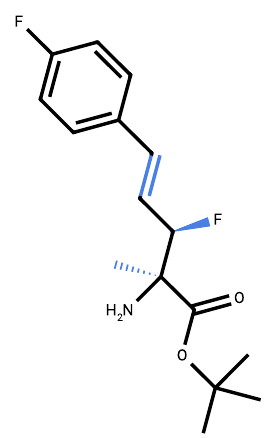} & {\footnotesize R/S not annotated}\newline {\scriptsize\color{gray50} \texttt{anie.202313838 example\_1}} \\[1pt]

\midrule
\multicolumn{4}{@{}l@{}}{\footnotesize\bfseries Atoms and functional groups} \\[3pt]
 \srcfig{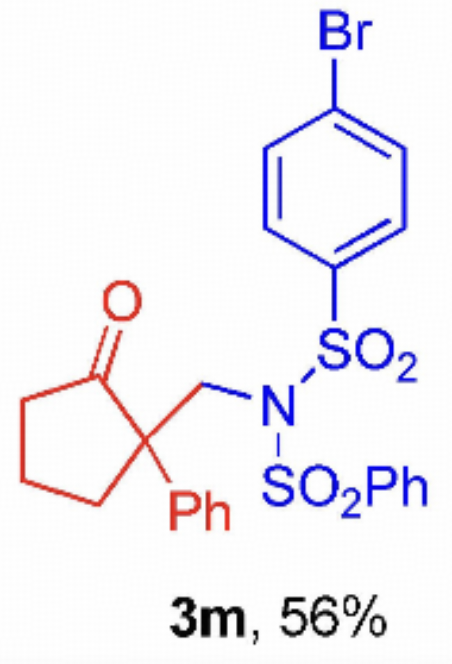} & \gtfig{redcircle/c6_gt} & \ourfig{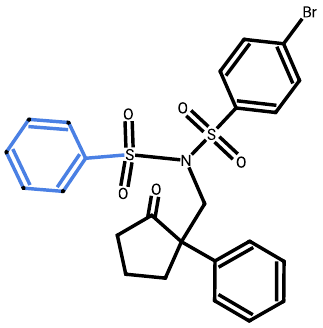} & {\footnotesize Element mis-annotated}\newline {\scriptsize\color{gray50} \texttt{ajoc.202200531 example\_1}} \\[1pt]

\midrule
\multicolumn{4}{@{}l@{}}{\footnotesize\bfseries Skeleton and whole-molecule replacement} \\[3pt]
 \srcfig{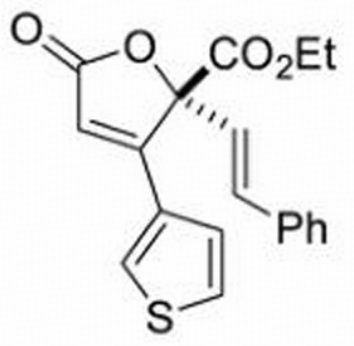} & \gtfig{redcircle/c8_gt} & \ourfig{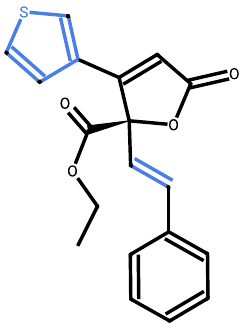} & {\footnotesize Wrong substitution position; E/Z absent}\newline {\scriptsize\color{gray50} \texttt{253.jpg}} \\[2pt]
 \srcfig{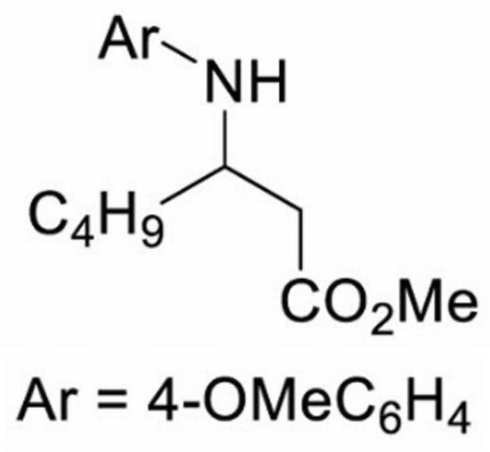} & \gtfig{redcircle/c7_gt} & \ourfig{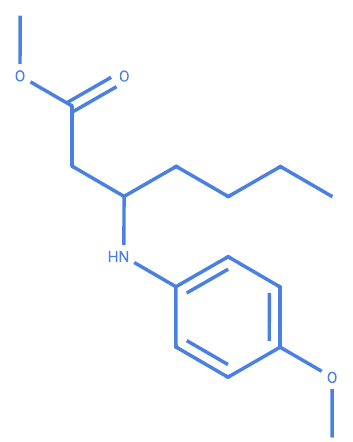} & {\footnotesize Product replaced by intermediate}\newline {\scriptsize\color{gray50} \texttt{168.jpg}} \\
\end{tabular}
\caption{\textbf{Representative annotation errors.}
\label{fig:chemeagle_error_comparison}
\textcolor{bcol}{Blue bonds} mark the erroneous substructure;
\textcolor[RGB]{255,0,0}{red circles} in the GT panels mark the
corresponding annotation errors; grey tags name the source image.}
\label{fig:cases}
\end{figure}
\begin{figure}[H]
\centering
\includegraphics[width=\textwidth]{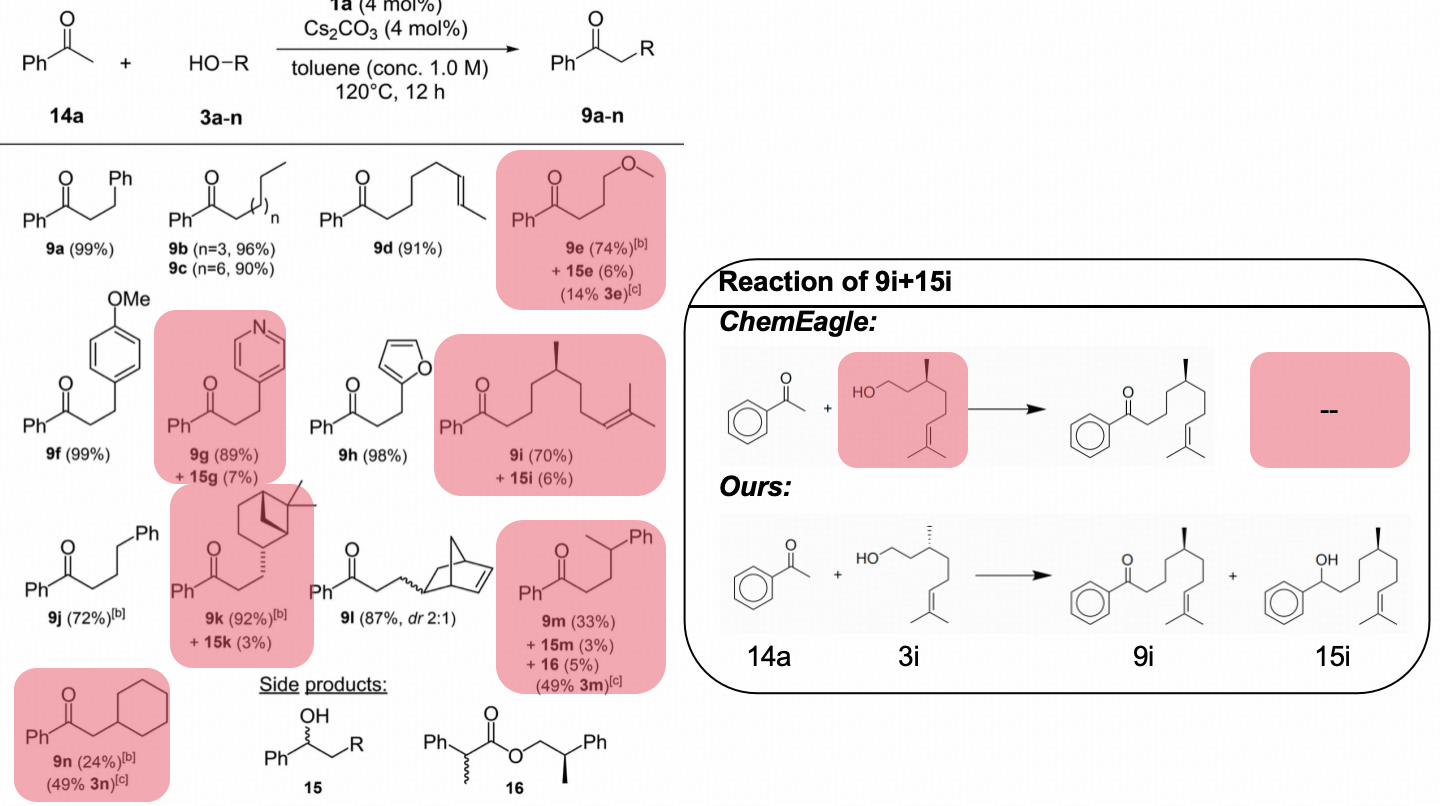}
\caption{\textbf{Multiple errors within a single scheme} 
\label{fig:chemeagle_error_fig}
(\texttt{ejoc.202300707 example\_2}). Six of the fourteen reactions are mis-annotated by ChemEAGLE: six side products (15e/g/i/k/m and 16) are omitted entirely, and one reactant (3i) is given as the wrong enantiomer.}
\label{fig:multimrr}
\end{figure}

\FloatBarrier
\endgroup

\section{Case Studies}
\label{app:cases}

\subsection{How \agentname{} Corrects Direct-Model Errors}
\label{app:case-corrections}
\begin{figure}[!htb]
\centering
\includegraphics[width=\linewidth]{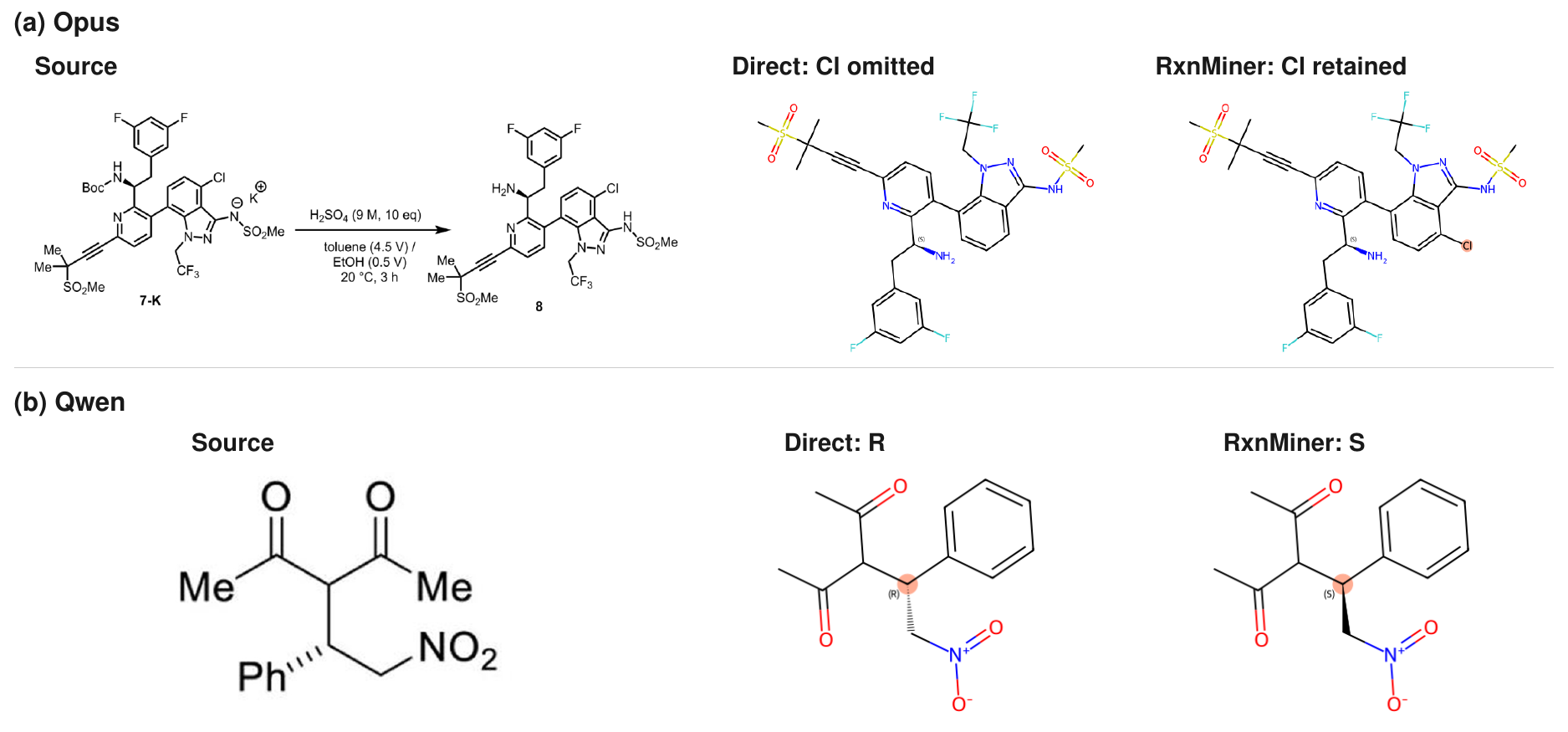}
\caption{\textbf{Corrections on the same source figure and backbone.} Source crops (left) and products rendered from the output JSON (middle and right). \textbf{(a)} Opus omits an aryl chlorine; \agentname{} retains it. The omission affects both participants in all three table rows (direct: 0/3 matched pairs; \agentname{}: 3/3). \textbf{(b)} Qwen predicts the opposite product configuration ($R$ instead of $S$); \agentname{} agrees with the source. The highlighted atoms locate the two corrections; conditions and outcomes are assessed separately.}
\label{fig:e1-cl}\label{fig:e1-stereo}
\end{figure}

We compare direct prediction and RxnMiner on the same figure and backbone, using Claude Opus 5 and Qwen3.8-Flash-Next. Figure~\ref{fig:e1-cl} visualizes structures and fields from the archived JSON outputs. These cases were selected to expose interpretable corrections; they are not a random error audit and do not estimate the frequency of the illustrated failure modes. Counts below refer only to the displayed figures. Molecular-pair agreement is checked by stereochemistry-aware canonicalization and one-to-one multiset matching against the reference; conditions and outcomes are inspected separately.

\paragraph{Recovering an omitted substituent.}
In the Boc-deprotection table (Figure~\ref{fig:e1-cl}a), Opus direct prediction omits the aryl chlorine in both reactant 7-K and product 8. For product 8, deleting this chlorine from the RxnMiner structure reproduces the direct-model molecular graph exactly. The shared error is propagated to all three table entries: none of the three direct reactant--product pairs matches the reference, whereas all three RxnMiner pairs do. This illustrates a recognition error that remains chemically parsable and can therefore survive a syntax-only check.

\paragraph{Correcting a stereochemical inversion.}
For the first selected example in the Michael-addition scope (Figure~\ref{fig:e1-stereo}b), Qwen direct prediction recovers the product connectivity but encodes the opposite absolute configuration. RDKit assigns $R$ to the direct product and $S$ to the RxnMiner product and reference; the two predictions become identical when stereochemistry is removed. The source shows a hashed bond to the phenyl substituent, and the RxnMiner output retains a local stereochemical audit supporting its correction of the initial molecular reading. This example establishes a product-level correction, not complete correctness of all fields: the Qwen-RxnMiner output still contains an incorrect structure for catalyst A4.

\FloatBarrier

\subsection{Successful Extractions}
\label{app:case-success}
We illustrate extraction on demanding figures with dense substrate scopes and stereoselective condition tables. Figure~\ref{fig:e2-dense} shows selected records at readable scale. All reference molecular pairs are recovered in these selected cases, checked by structure-aware one-to-one matching. These examples describe selected cases rather than an aggregate success rate; conditions and outcomes are assessed separately from molecular-pair agreement.

\begin{figure}[!htbp]
\centering
\includegraphics[width=0.78\linewidth]{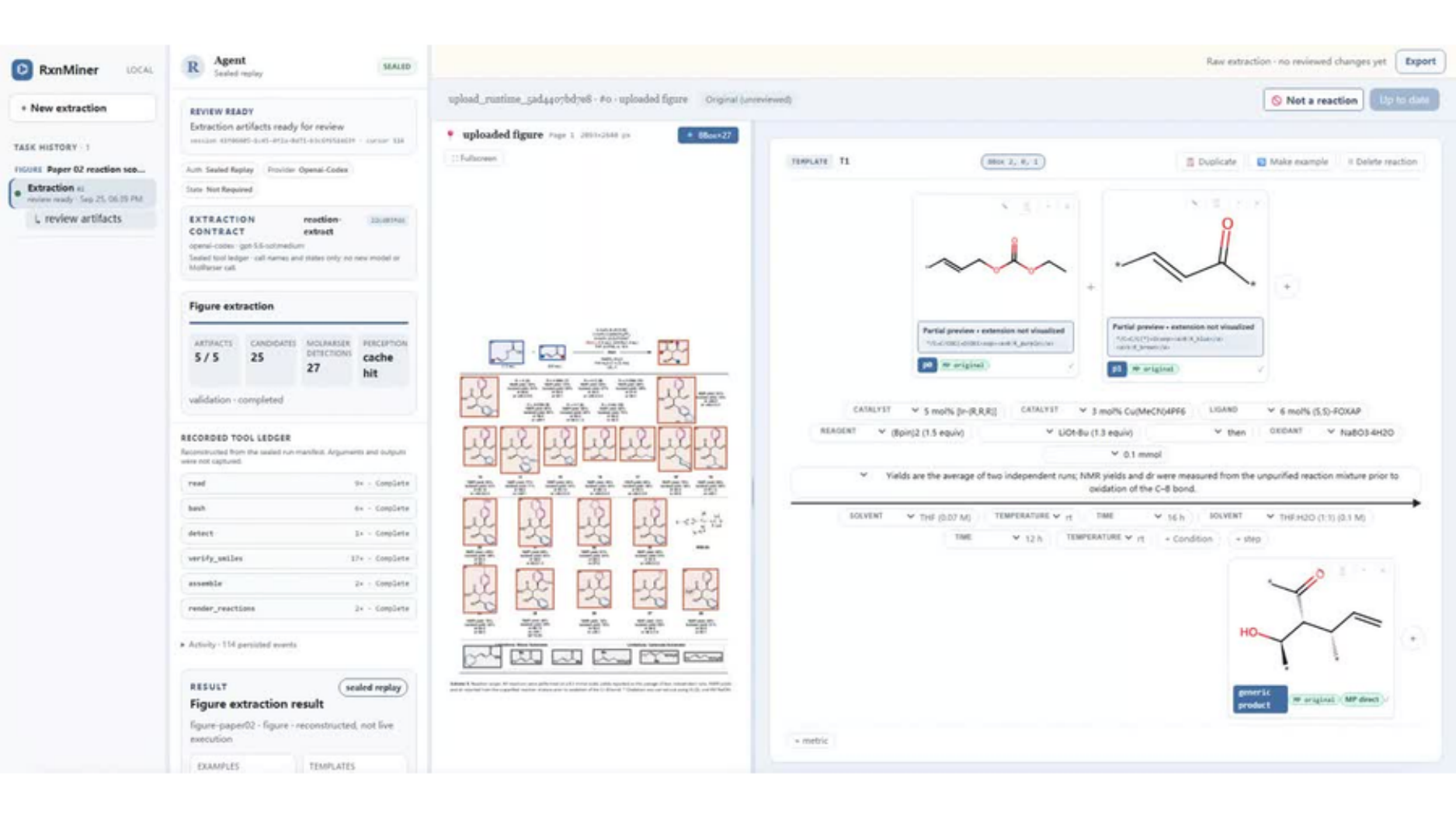}
\caption{\textbf{Inspecting a complex extraction in the RxnMiner workspace.} The source figure, molecular boxes, rendered structures, and condition fields are linked in the blue-violet interface.}\label{fig:e2-software}
\end{figure}

\begin{figure}[!htb]
\centering\includegraphics[width=\linewidth]{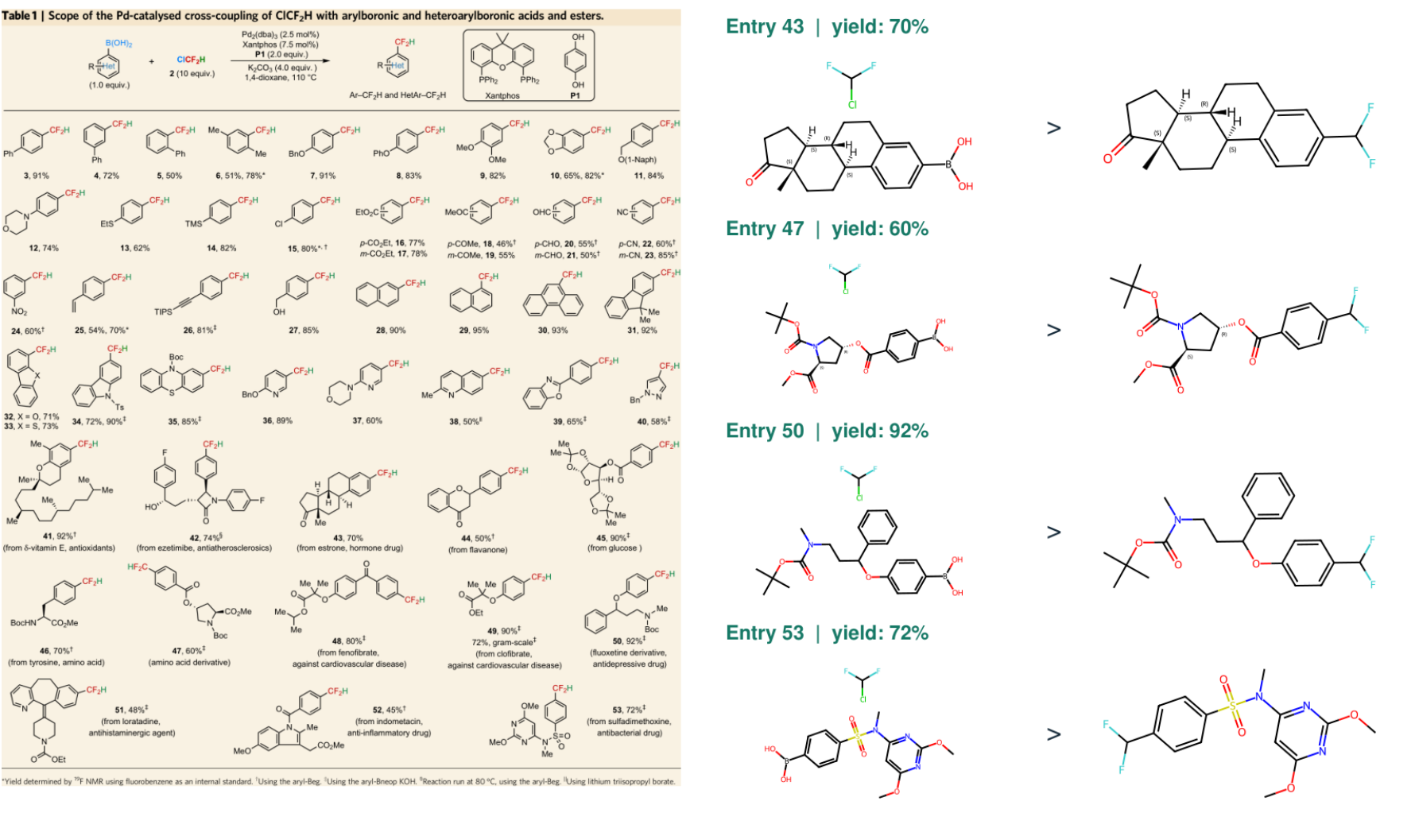}
\caption{\textbf{A dense scope with complex molecular scaffolds.} Opus-RxnMiner recovers 51/51 reference molecular pairs. Four enlarged reactions illustrate reconstructed boron-containing reactants and functionalized products.}\label{fig:e2-dense}
\end{figure}

\paragraph{Inspecting extraction in the software workspace.}
Figure~\ref{fig:e2-software} shows the RxnMiner software workspace. The workspace displays a dense stereodivergent extraction with linked molecular regions, rendered structures, and condition fields. The interface places extraction results, source-image regions, rendered structures, and condition fields in a single workspace. The screenshot illustrates inspection of extraction results; interface completion badges are not chemical-accuracy measurements.

\paragraph{Dense scope with complex molecular scaffolds.}
Figure~\ref{fig:e2-dense} contains 51 reactions for the palladium-catalyzed coupling of chlorodifluoromethane with arylboronic and heteroarylboronic acids or esters. Opus-RxnMiner recovers all 51 reference molecular pairs, including the functionalized scaffolds enlarged in the panel. The figure requires enumerating substituent lists, retaining complex scaffold connectivity and specified stereochemistry, and reconstructing the corresponding boron-containing reactant for each entry. Some source entries contain multiple yield measurements or footnote-dependent variants; molecular-pair agreement does not establish correctness of every outcome or footnote interpretation.

\subsection{Corner Cases}
\label{app:case-corner}

\paragraph{Capability boundaries.}
Dense stereochemistry and unspecified configurations remain challenging for visual reaction extraction. End-to-end accuracy also depends on upstream molecular recognition tools, including MolParser. The following examples from the ChemEAGLE collection illustrate challenges in preserving local stereochemical configurations and retaining unspecified stereochemistry. Each figure pairs the source entry with a comparison between the ground truth (GT) and the Opus 5 + RxnMiner output; the boxes are added for presentation only.

\begin{figure}[!htb]
  \centering
  \includegraphics[width=0.98\linewidth]{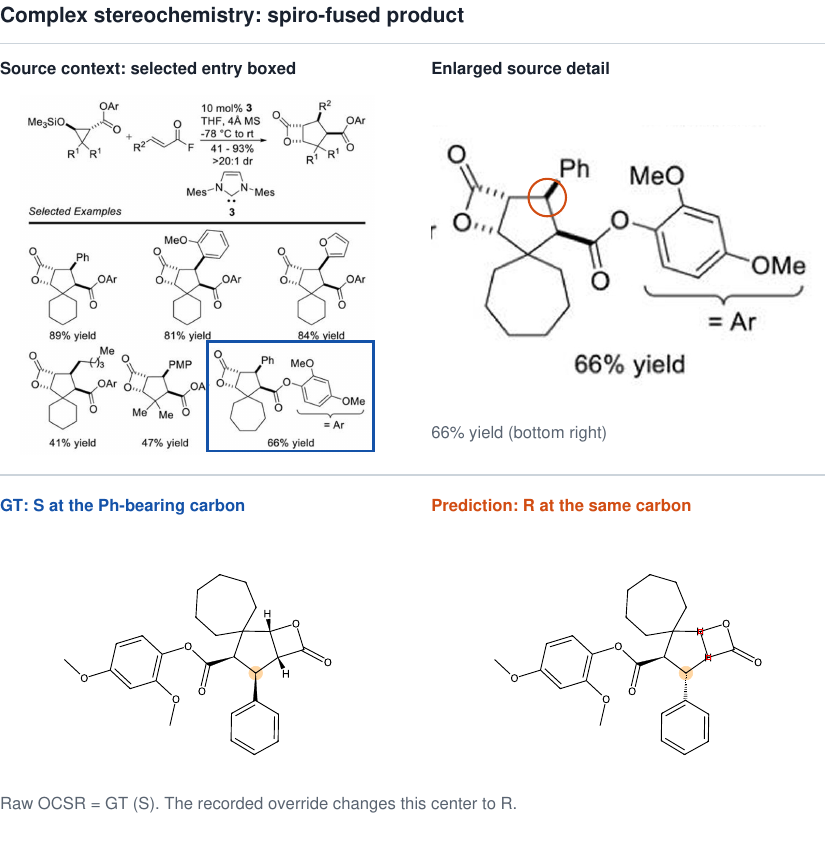}
  \caption{\textbf{Local stereochemistry in a spiro-fused scaffold.} Source: ChemEAGLE collection. The boxed example (66\% yield) and enlarged crop locate the selected entry. The highlighted phenyl-bearing center is S in the GT and R in the extracted product; connectivity and the remaining stereocenters agree.}
  \label{fig:corner-206}
\end{figure}

\begin{figure}[!htb]
  \centering
  \includegraphics[width=0.98\linewidth]{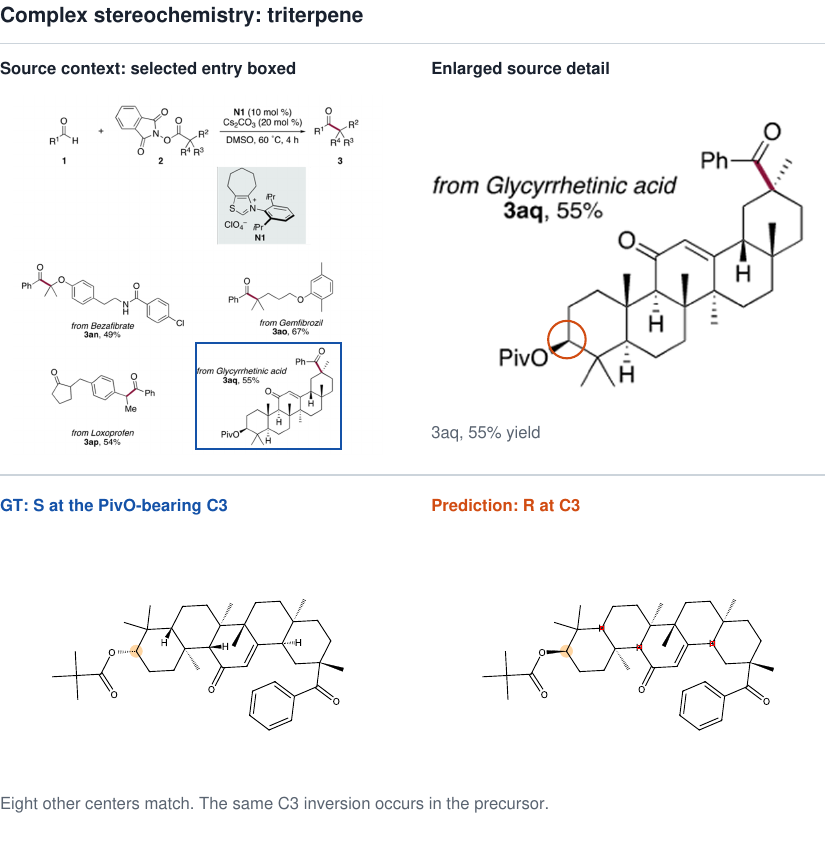}
  \caption{\textbf{Local stereochemistry in a triterpene scaffold.} Source: ChemEAGLE collection. The comparison highlights the PivO-bearing C3 center of product 3aq (55\% yield), which is S in the GT and R in the extracted product; connectivity and the remaining eight stereocenters agree.}
  \label{fig:corner-04-jacs}
\end{figure}

\paragraph{Preserving local stereochemistry.}
Two examples illustrate the sensitivity of exact molecular recovery to individual stereochemical assignments in complex scaffolds. In the spiro-fused product (66\% yield; Figure~\ref{fig:corner-206}), connectivity and the other stereocenters agree with the GT, but the phenyl-bearing center is represented as R instead of S. Similarly, for the triterpene product 3aq (55\% yield; Figure~\ref{fig:corner-04-jacs}), connectivity and eight of nine stereocenters agree with the GT, while the PivO-bearing C3 center is represented as R rather than S. These cases highlight the challenge of preserving every local configuration even when the overall molecular scaffold is recovered correctly.

\begin{figure}[!htb]
  \centering
  \includegraphics[width=0.98\linewidth]{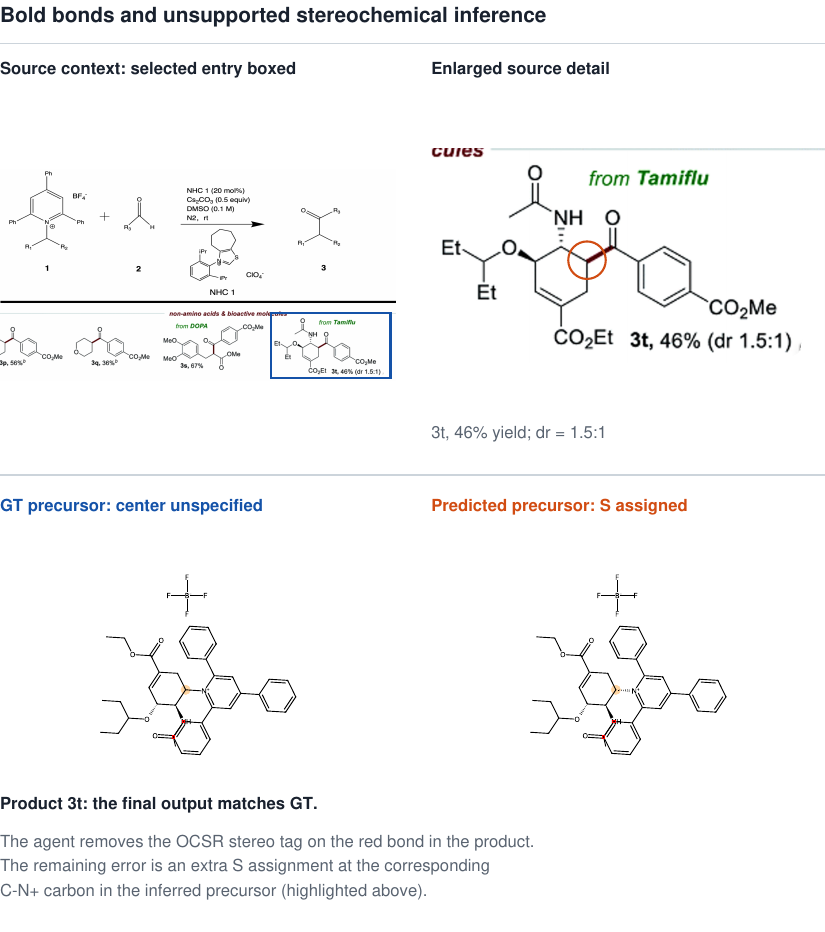}
  \caption{\textbf{Preserving unspecified stereochemistry.} Source: ChemEAGLE collection. The boxed entry 3t provides the source context. The final product agrees with the GT, while the reactant comparison highlights an additional S assignment at a center left unspecified in the GT.}
  \label{fig:corner-11-2}
\end{figure}

\paragraph{Retaining unspecified stereochemistry.}
Entry 3t (Figure~\ref{fig:corner-11-2}) illustrates the distinction between specified and unspecified stereochemistry. The final product agrees with the GT, but a C--N$^+$ carbon in the reactant is assigned S where the GT leaves its configuration unspecified. Reliable extraction must preserve this distinction rather than add stereochemical specificity unsupported by the source evidence.

\FloatBarrier

\section{Silver Reaction Corpus and Quality Review}
\label{app:corpus}

The silver reaction corpus demonstrates the application of \agentname{} to a broader collection of reaction figures. Its automatically extracted records have a different review status from the references used for \figbench{}. The current snapshot comprises 10,107 figures, 88,202 reaction records, and 15,012 templates. Figure~\ref{fig:silver-corpus-quality} in Section~\ref{sec:corpus} shows the source pool, the extracted records, their field coverage, and the record workflow.

Provenance associates each retained record with the available source evidence and distinguishes tool-read, template-derived, and model-written structures. This supports tracing a record back to a figure without implying that the collection is fully verified. The same molecular reading can contribute to several reconstructed records, so a shared recognition error may affect more than one extracted reaction.

In the reported random sample of 1,000 figures, 25,223 of 25,225 SMILES passed parsing (99.99\%). This measures representational validity, not chemical correctness: a parsable molecular graph can still have the wrong identity or reaction assignment.  Consequently, the available parsing statistic supports no estimate of reaction correctness.

 Example records illustrate the output and provenance; their readability or visual plausibility does not establish representative accuracy. The corpus is a secondary application of the benchmark and agent contributions, with quality claims limited to the available audit evidence.
\clearpage
\section{Related Work}
\label{app:related-work}

This section expands Section 2 and brings the literature
survey together with the same-figure output comparison. We distinguish a
system's reported functionality, the components enabled in a particular
execution, and the fields actually returned. The survey covers nine prior
extraction systems; the case study contains eight, with RxnIM not executed
in that comparison. 

\subsection{Extraction Systems and Their Output Targets}
\label{app:extractor-profiles}

\paragraph{ReactionDataExtractor 2.0 (RDE2).}
RDE2 combines component segmentation, optical character and structure
recognition, and scheme reconstruction~\citep{rde2023,rde2_code}. Its output
represents detected reaction entities and their connectivity as a
machine-readable reaction graph. The relevant distinction for our task is
between recovering a drawn scheme and expanding the entry-specific
experiments encoded by a shared template or scope table. A scheme-level
output need not establish that every such entry has been instantiated.

\paragraph{RxnScribe.}
RxnScribe formulates diagram parsing as sequence generation and predicts
reactant, condition, and product regions for each
reaction~\citep{rxnscribe2023}. Its official interface can additionally
invoke MolScribe and OCR, returning molecular SMILES and condition text
alongside the boxes~\citep{rxnscribe_code,molscribe_code}. Thus, the
standalone parser's target should not be confused with the complete
pipeline's output. Molecular recognition and subsequent entry-specific
reconstruction remain distinct operations.

\paragraph{RxnCaption.}
RxnCaption uses molecular bounding boxes and indices as visual prompts,
allowing a vision--language model to describe reactions by referring to
those indexed components and reading condition
text~\citep{rxncaption2026}. The formulation separates localization from
reaction description. Its released diagram-parsing annotations encode
molecular components and reaction roles rather than explicit molecular
SMILES. Recovering a role-linked component is therefore a different target
from reconstructing and checking the complete molecular participants of
each experimental entry.

\paragraph{RxnID.}
RxnID uses molecular identifiers as visual prompts and introduces
verifiable reinforcement learning with reaction-level
rewards~\citep{idtvp2026}. The identifier representation also supports
alignment with textual descriptions; the paper presents a cross-modal
verification extension that combines visual and textual extraction.
Identifier-level parsing, training-time verifiable rewards, and
source-grounded inspection of molecular stereochemistry should not be
conflated. Our figure-only case study does not supply the additional
manuscript context used by that document-level extension.

\paragraph{RxnIM.}
RxnIM uses instruction-conditioned multimodal modeling for reaction
component extraction, condition reading, and role
identification~\citep{rxnim2025}. Its complete RxnIM.web workflow includes
molecular recognition and reports reactant and product SMILES, detailed
conditions, and reaction representations~\citep{rxnim_code}. This integration distinguishes the region-based component-parsing
output from the molecular representations available through the complete
workflow.
RxnIM is covered by the capability survey but has no result in the
same-figure case study below.

\paragraph{MinerU.Chem.}
MinerU.Chem integrates chemistry-specific recognition with document
parsing, including molecular region detection, identifier extraction,
structure recognition, and reaction-scheme parsing, producing molecule
and reaction summaries~\citep{mineruchem2026}. The paper explicitly
identifies fully automated expansion of Markush structures and R-group
substitution patterns as an outstanding limitation. This separates
recognition of generic structures from full instantiation of their
entry-specific variants; the partial-support entry in
Table~\ref{tab:extractor-capabilities} should be read in that context.

\paragraph{OpenChemIE.}
OpenChemIE integrates molecular detection, identifier coreference,
MolScribe, RxnScribe, and text extraction~\citep{openchemie2024,openchemie_code}.
It aligns reactions with conditions and resolves R-groups from text
assignments or product structures. For a product scope, it matches a
specific product to its template, identifies substituent fragments through
subgraph matching, and substitutes those fragments into reactant
templates. This provides a precedent for reconstructing entry-specific reactants
that are not drawn independently in the source figure.

\paragraph{ChemEAGLE.}
ChemEAGLE coordinates a planner with specialized agents for template
parsing, molecular recognition, structure- and text-based R-group
substitution, condition interpretation, and output
assembly~\citep{chemeagle2024,chemeagle_code}. Its published tools include
reactant-SMILES reconstruction from product variants, and its architecture
also includes observer agents for checking the extraction workflow.
Accordingly, both reconstruction and model-based review are reported
capabilities. The mechanism comparison in
Table~\ref{tab:closest-extractor-mechanisms} focuses on entry binding,
structure sourcing, review procedures, and molecular lineage rather than
claiming exclusive access to these capabilities.

\paragraph{MERMaid.}
MERMaid is a document-level extraction pipeline: VisualHeist segments
figures and tables, DataRaider extracts a reaction schema using multimodal
analysis, and KGWizard supports knowledge-graph
construction~\citep{mermaid2025,mermaid_code}. Its implementation includes
an RxnScribe-based route for molecular SMILES extraction. It should
therefore be treated as a configurable pipeline rather than only a
vision--language response. The single-figure comparison evaluates the
returned fields for the supplied diagram, not the additional information
that might be available elsewhere in a complete article.

\subsection{Reported Capabilities and Closest-System Mechanisms}
\label{app:reported-capabilities}
\begingroup
\raggedbottom

\paragraph{Reported capabilities.}
Table~\ref{tab:extractor-capabilities} summarizes the capabilities reported in the cited papers and accompanying documentation of reaction-extraction systems, alongside those of \agentname{}. The comparison covers four aspects:
recognition, reaction association, reconstruction, and quality control.
Condition-text extraction and reaction grouping are reported across
the surveyed systems, whereas molecular SMILES output is not universal.
R-group filling and undrawn-reactant completion are less widely
supported and are distinguished from recognizing explicitly drawn
structures: they require constructing entry-specific molecules from
shared templates or resolving reactants specified without a separate
structure drawing. We also distinguish molecular validity checks,
visual inspection, and model-based review rather than treating them
as interchangeable forms of quality control. These entries describe
reported functionality, not measured extraction accuracy.
Table~\ref{tab:closest-extractor-mechanisms} complements this survey with a mechanism-level comparison of
OpenChemIE, ChemEAGLE, and \agentname{}, covering workflow design,
entry binding, structure sourcing, review, and molecular lineage.

\begin{table}[!ht]
\centering
\caption{System-level reported extraction capabilities.}
\label{tab:extractor-capabilities}
\begingroup
\fontsize{9}{10.8}\selectfont
\setlength{\tabcolsep}{1.4pt}
\renewcommand{\arraystretch}{1.2}
\newcommand{\rxmyes}{\ensuremath{\checkmark}}
\newcommand{\rxmnr}{\textemdash}
\begin{tabularx}{\linewidth}{@{}>{\raggedright\arraybackslash}X*{8}{c}@{}}
\toprule
 & \multicolumn{2}{c}{\textbf{Recognition}}
 & \textbf{Assoc.}
 & \multicolumn{2}{c}{\textbf{Reconstruction}}
 & \multicolumn{3}{c}{\textbf{Quality control}} \\
\cmidrule(lr){2-3}\cmidrule(lr){4-4}\cmidrule(lr){5-6}\cmidrule(l){7-9}
\textbf{System} & \textbf{SMILES}
& \shortstack{\textbf{Cond.}\\\textbf{text}}
& \shortstack{\textbf{Reaction}\\\textbf{grouping}}
& \shortstack{\textbf{R-group}\\\textbf{fill}}
& \shortstack{\textbf{Undrawn}\\\textbf{reactants}}
& \textbf{Validity}
& \shortstack{\textbf{Visual}\\\textbf{inspection}}
& \shortstack{\textbf{Model-based}\\\textbf{review}} \\
\midrule
RDE2~\citep{rde2023,rde2_code}
& \rxmyes & \rxmyes & \rxmyes & \rxmnr & \rxmnr & \rxmnr & \rxmyes & \rxmnr \\
RxnScribe~\citep{rxnscribe2023,rxnscribe_code,molscribe_code}
& \rxmyes & \rxmyes & \rxmyes & \rxmnr & \rxmnr & \rxmyes & \rxmyes & \rxmnr \\
RxnCaption~\citep{rxncaption2026}
& \rxmnr & \rxmyes & \rxmyes & \rxmnr & \rxmnr & \rxmnr & \rxmnr & \rxmnr \\
RxnID~\citep{idtvp2026}
& \rxmnr & \rxmyes & \rxmyes & \rxmnr & \rxmnr & \rxmyes & \rxmnr & \rxmnr \\
RxnIM~\citep{rxnim2025,rxnim_code}
& \rxmyes & \rxmyes & \rxmyes & \rxmnr & \rxmnr & \rxmnr & \rxmyes & \rxmnr \\
MinerU.Chem~\citep{mineruchem2026}
& \rxmyes & \rxmyes & \rxmyes & partial & \rxmnr & \rxmnr & \rxmyes & \rxmnr \\
OpenChemIE~\citep{openchemie2024,openchemie_code,molscribe_code}
& \rxmyes & \rxmyes & \rxmyes & \rxmyes & \rxmyes & \rxmyes & \rxmyes & \rxmnr \\
ChemEAGLE~\citep{chemeagle2024,chemeagle_code}
& \rxmyes & \rxmyes & \rxmyes & \rxmyes & \rxmyes & \rxmyes & \rxmyes & \rxmyes \\
MERMaid~\citep{mermaid2025,mermaid_code,molscribe_code}
& \rxmyes & \rxmyes & \rxmyes & \rxmnr & \rxmnr & \rxmyes & \rxmnr & \rxmnr \\
\midrule
\textbf{\textsc{RxnMiner}} (ours)
& \rxmyes & \rxmyes & \rxmyes & \rxmyes & \rxmyes & \rxmyes & \rxmyes & \rxmyes \\
\bottomrule
\end{tabularx}
\endgroup
\end{table}

\begin{table}[!ht]
\centering
\caption{Mechanisms of OpenChemIE, ChemEAGLE, and \textsc{RxnMiner}.}
\label{tab:closest-extractor-mechanisms}
\begingroup
\fontsize{9}{10.6}\selectfont
\setlength{\tabcolsep}{4pt}
\renewcommand{\arraystretch}{1.17}
\begin{tabularx}{\linewidth}{@{}>{\raggedright\arraybackslash}p{0.157\linewidth}
  *{3}{>{\raggedright\arraybackslash}X}@{}}
\toprule
\textbf{Dimension} & \textbf{OpenChemIE}~\citep{openchemie2024,openchemie_code}
& \textbf{ChemEAGLE}~\citep{chemeagle2024,chemeagle_code}
& \textbf{\textsc{RxnMiner}} \\
\midrule
Workflow & Dedicated models + integration algorithms
& Planner + specialists + optional observers
& VLM agent + chemistry tools \\
\addlinespace[3pt]
Entry binding & Identifier coreference + table alignment
& Task-specific multimodal agents
& Explicit entry--template--substituent plan \\
\addlinespace[3pt]
Structure sourcing & MolScribe + graph substitution
& OCSR + applied edits + reconstruction tools
& Preserve drawn instances; instantiate templates; flag fallback \\
\addlinespace[3pt]
Model-based review & No separate model-based review loop reported
& Optional image/output review with retry
& Stereo-triggered source-crop/RDKit comparison \\
\addlinespace[3pt]
Molecular lineage & Source boxes + compound identifiers
& Cached raw readings + edit audit
& Tool-read / template-derived / model-written \\
\bottomrule
\end{tabularx}
\endgroup
\end{table}

\FloatBarrier
\endgroup

\subsection{Same-Figure Comparison of Returned Outputs}
\label{app:same-figure-comparison}
\begingroup
\raggedbottom
\paragraph{Case-level visual comparison.}
We selected one ChemEAGLE case for this comparison. Each system card keeps the original image visible and
then shows the fields actually returned by that system: bounding boxes,
reaction/SMILES strings, conditions, or table metadata. Checklist
states are compared with the GT: a check means correct, a cross means that the
field was not correctly recovered, and the half-filled symbol means that only a
correct subset was recovered.
For SMILES, ratios use the final-release molecule-set scorer: matched SMILES
over the 9 GT scoring molecules.

This real case contains four selected reaction examples and 9 concrete
GT-scored molecules. RDE2 returned no structured fields, while RxnScribe,
RxnCaption, RxnID, and MinerU.Chem returned partial OCR, condition, or
molecule-table information. OpenChemIE produced one intermediate Markush
candidate but emitted no concrete final reaction. ChemEAGLE returned five
reaction records and matched only 7/9 GT molecules. MERMaid produced
metadata/template-style information. RxnMiner returned all four examples,
matched all 9 GT molecules, and supplied all 15 GT bbox references,
corresponding to 7 unique coordinate regions, through its extraction
provenance.

\newcommand{\okmark}{\(\checkmark\)}
\newcommand{\partmark}{\(\triangle\)}
\newcommand{\badmark}{\(\times\)}

\begin{table}[p]
  \centering
  \small
  \setlength{\tabcolsep}{4pt}
  \renewcommand{\arraystretch}{1.10}
  \caption{Checklist summary for the same real case from ChemEAGLE benchmark. \(\checkmark\) correct; \(\triangle\) partially correct; \(\times\) not correctly recovered. SMILES ratios use the final-release molecule-set scorer.}
  \label{tab:case-level-checklist}
  \begin{tabular}{@{}lp{0.82\linewidth}@{}}
    \toprule
  \textbf{System} & \textbf{bbox; SMILES; catalyst; text; solvent; temperature; time; yield; grouping; R-group} \\
    \midrule
    RDE2 & \badmark\ all fields not recovered \\
    RxnScribe & \partmark\ bbox; \partmark\ SMILES; \partmark\ catalyst; \partmark\ text; \partmark\ solvent; \partmark\ temperature; \partmark\ time; \badmark\ yield; \partmark\ grouping; \badmark\ R-group \\
    RxnCaption & \badmark\ bbox; \badmark\ SMILES; \partmark\ catalyst; \okmark\ text; \partmark\ solvent; \partmark\ temperature; \partmark\ time; \badmark\ yield; \badmark\ grouping; \badmark\ R-group \\
    RxnID & \badmark\ bbox; \badmark\ SMILES; \partmark\ catalyst; \partmark\ text; \partmark\ solvent; \partmark\ temperature; \partmark\ time; \badmark\ yield; \partmark\ grouping; \badmark\ R-group \\
    MinerU.Chem & \partmark\ bbox; \partmark\ SMILES; \partmark\ catalyst; \partmark\ text; \partmark\ solvent; \partmark\ temperature; \partmark\ time; \partmark\ yield; \partmark\ grouping; \badmark\ R-group \\
    OpenChemIE & \badmark\ bbox; \badmark\ SMILES (0/9); \partmark\ catalyst; \badmark\ text; \partmark\ solvent; \partmark\ temperature; \partmark\ time; \badmark\ yield; \partmark\ grouping; \partmark\ R-group \\
    ChemEAGLE & \badmark\ bbox; \partmark\ SMILES (7/9); \okmark\ catalyst; \okmark\ text; \partmark\ solvent; \okmark\ temperature; \okmark\ time; \partmark\ yield; \partmark\ grouping; \partmark\ R-group \\
    MERMaid & \badmark\ bbox; \badmark\ SMILES (0/9); \partmark\ catalyst; \badmark\ text; \okmark\ solvent; \badmark\ temperature; \okmark\ time; \okmark\ yield; \partmark\ grouping; \badmark\ R-group \\
    RxnMiner & \okmark\ bbox; \okmark\ SMILES (9/9); \okmark\ catalyst; \okmark\ text; \okmark\ solvent; \okmark\ temperature; \okmark\ time; \okmark\ yield; \okmark\ grouping; \okmark\ R-group \\
    \bottomrule
  \end{tabular}
\end{table}
\begin{figure}[p]
  \centering
  \includegraphics[width=\linewidth]{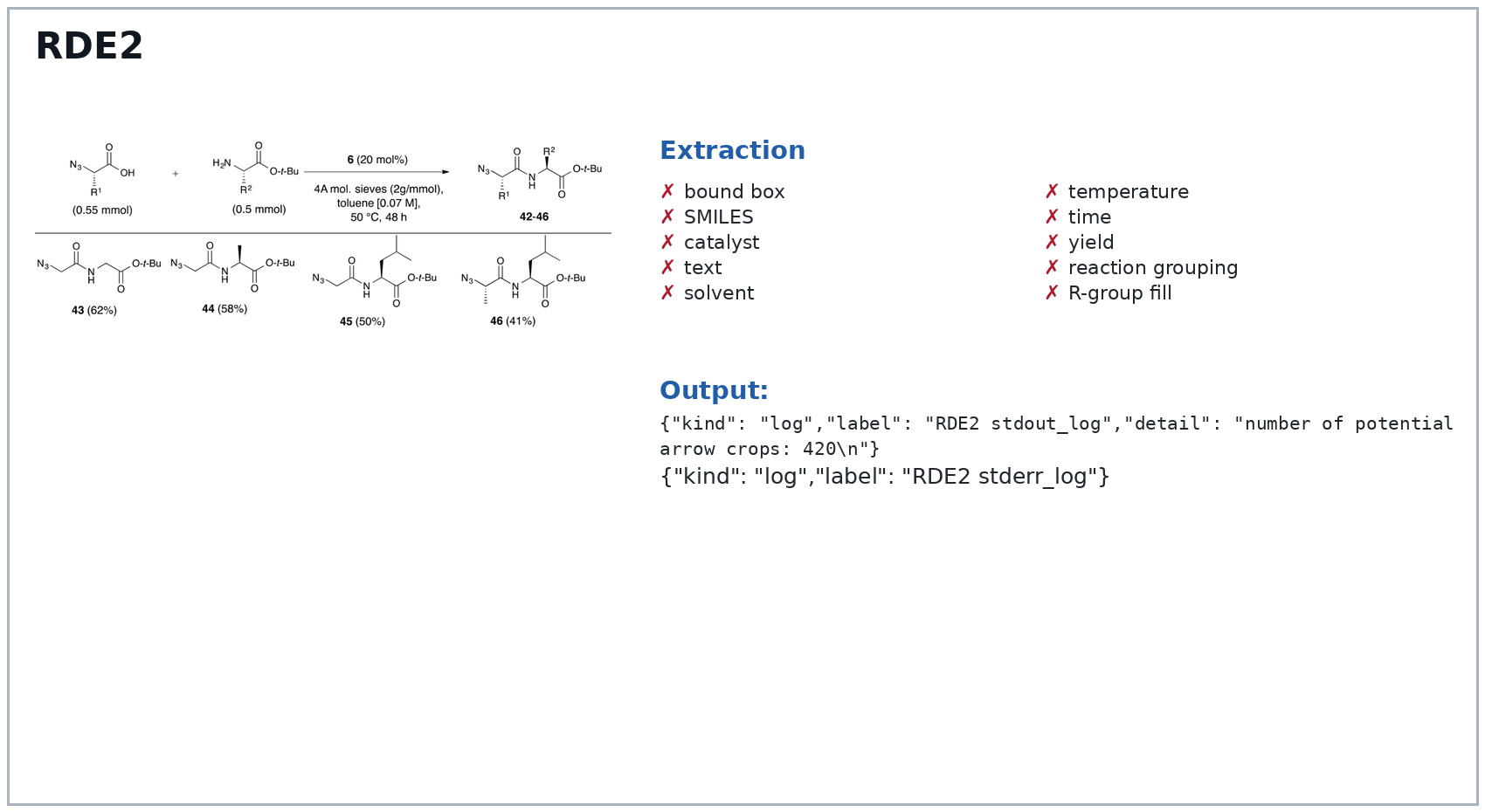}
  \caption{RDE2 output for the selected case.}
  \label{fig:case-level-extraction-1}
\end{figure}
\begin{figure}[p]
  \centering
  \includegraphics[width=\linewidth]{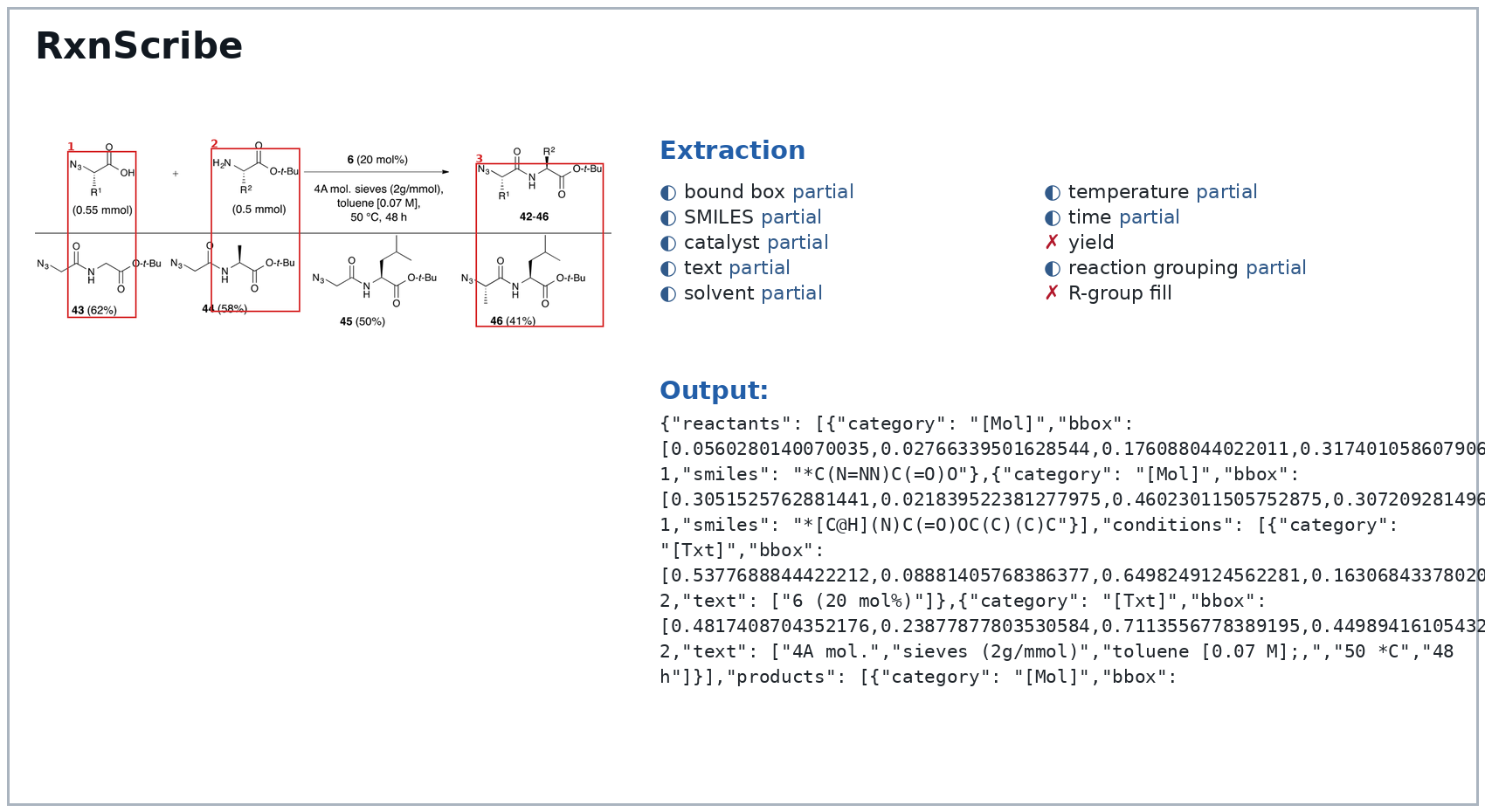}
  \caption{RxnScribe output for the selected case.}
  \label{fig:case-level-extraction-2}
\end{figure}
\begin{figure}[p]
  \centering
  \includegraphics[width=\linewidth]{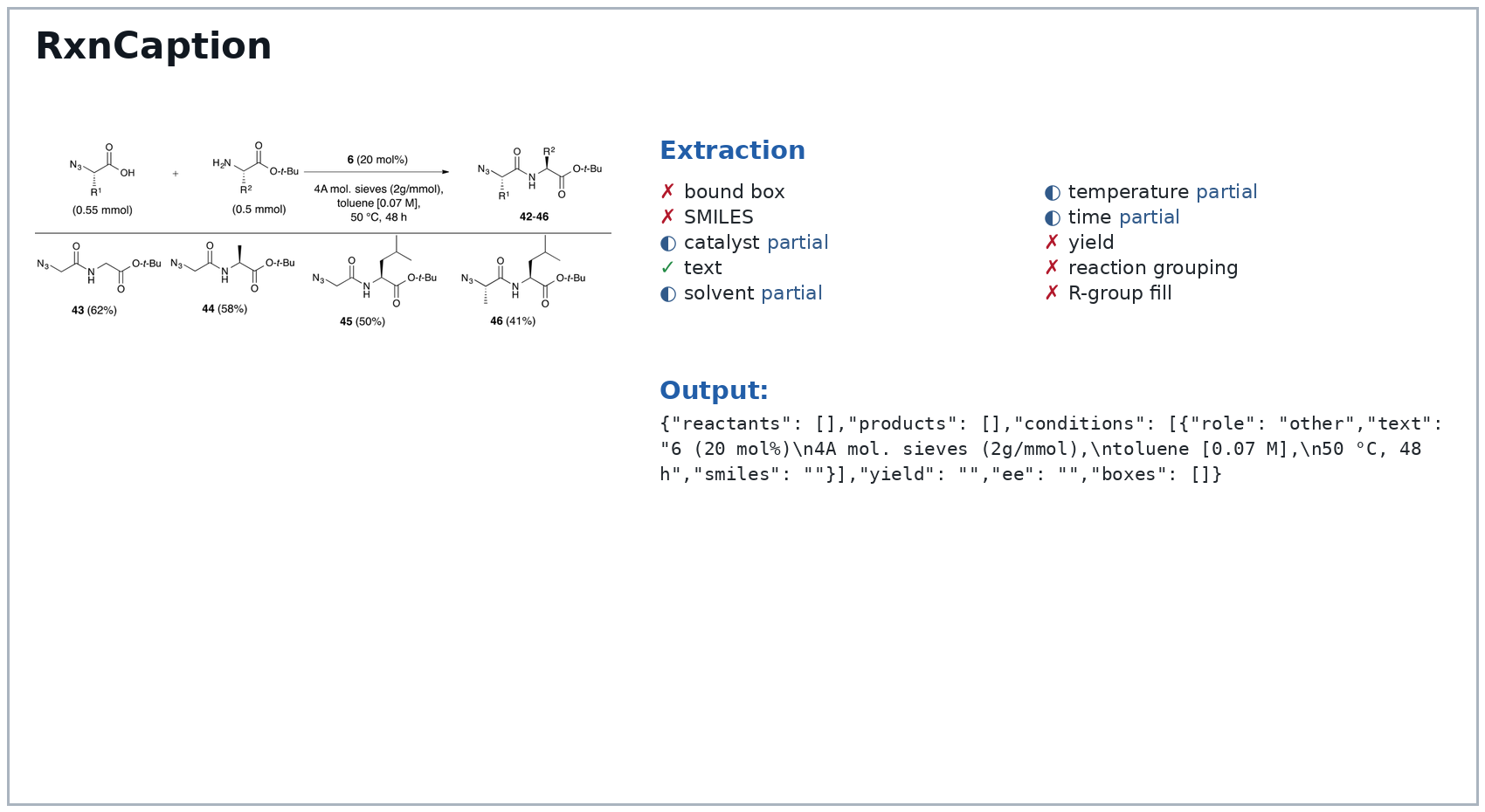}
  \caption{RxnCaption output for the selected case.}
  \label{fig:case-level-extraction-3}
\end{figure}
\begin{figure}[p]
  \centering
  \includegraphics[width=\linewidth]{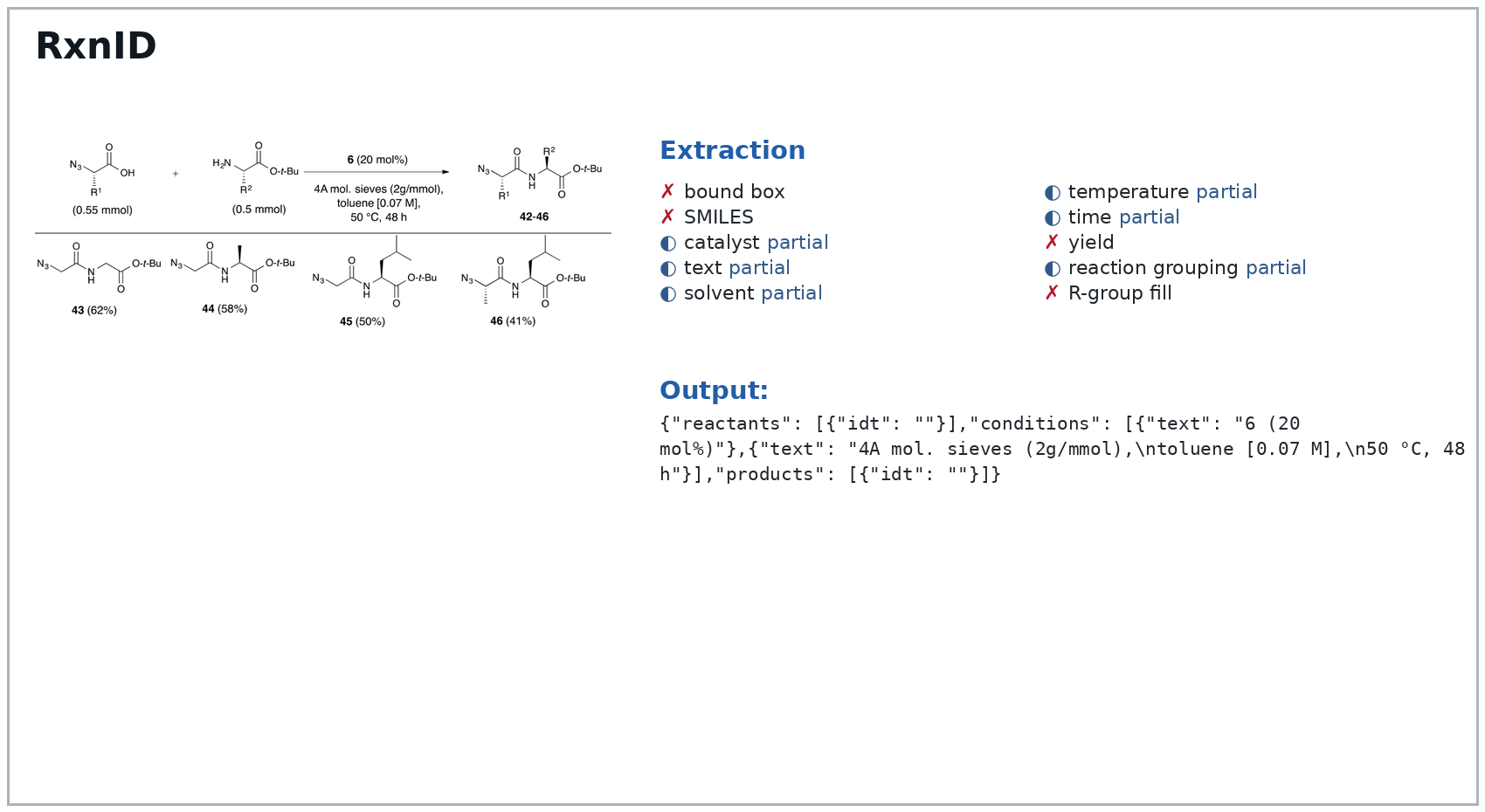}
  \caption{RxnID output for the selected case.}
  \label{fig:case-level-extraction-4}
\end{figure}
\begin{figure}[p]
  \centering
  \includegraphics[width=\linewidth]{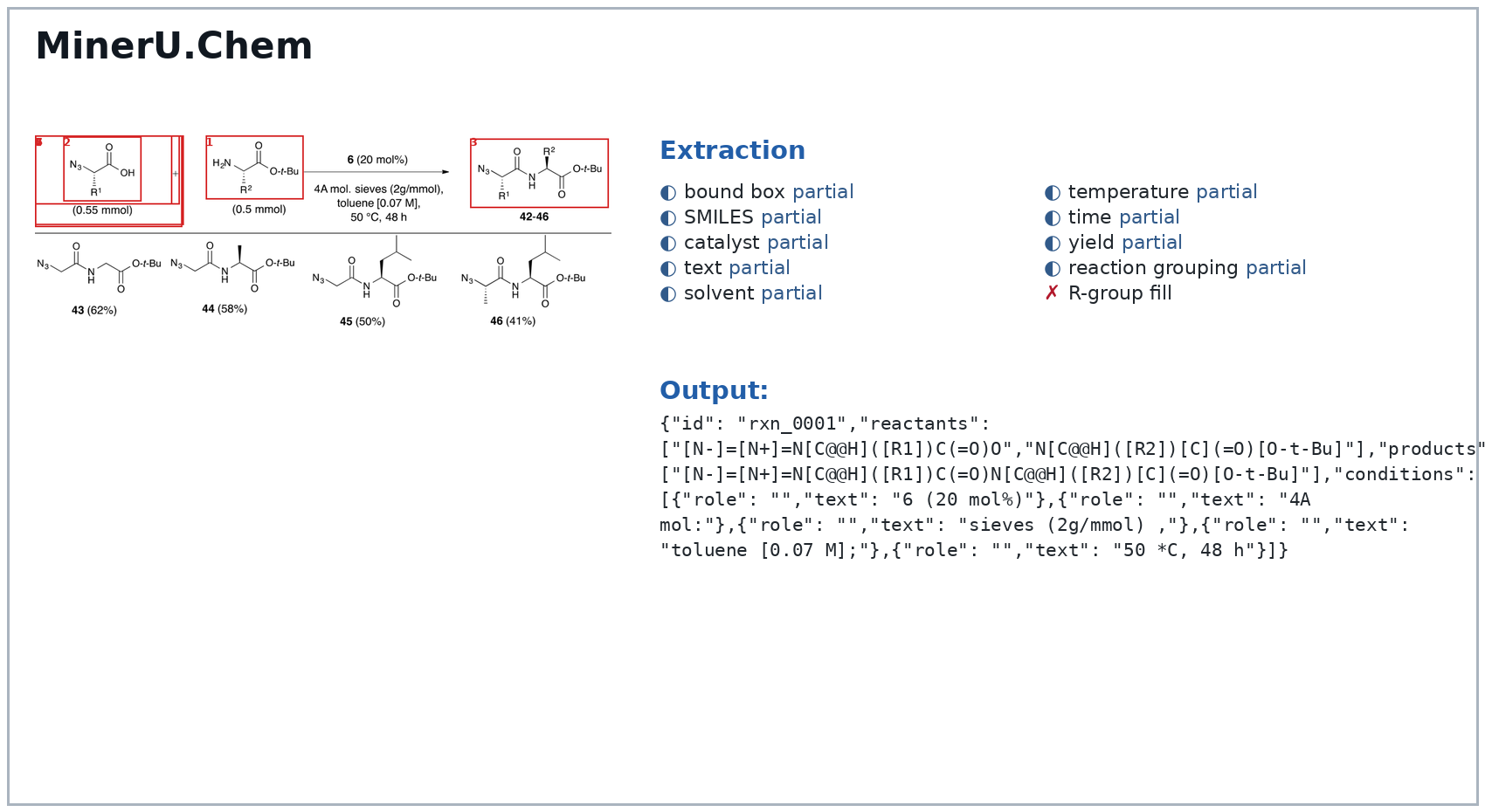}
  \caption{MinerU.Chem output for the selected case.}
  \label{fig:case-level-extraction-5}
\end{figure}
\begin{figure}[p]
  \centering
  \includegraphics[width=\linewidth]{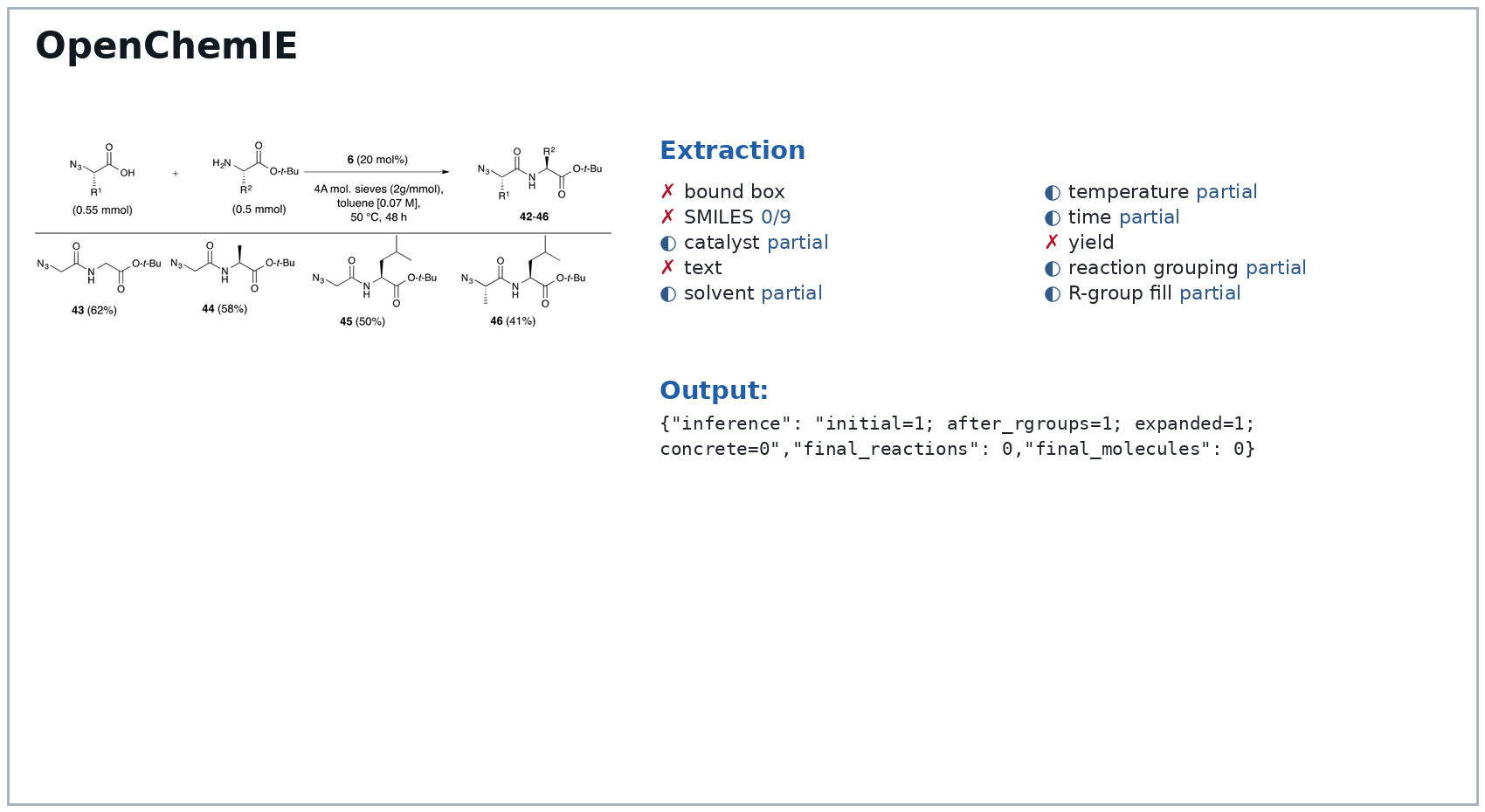}
  \caption{OpenChemIE output for the selected case.}
  \label{fig:case-level-extraction-6}
\end{figure}
\begin{figure}[p]
  \centering
  \includegraphics[width=\linewidth]{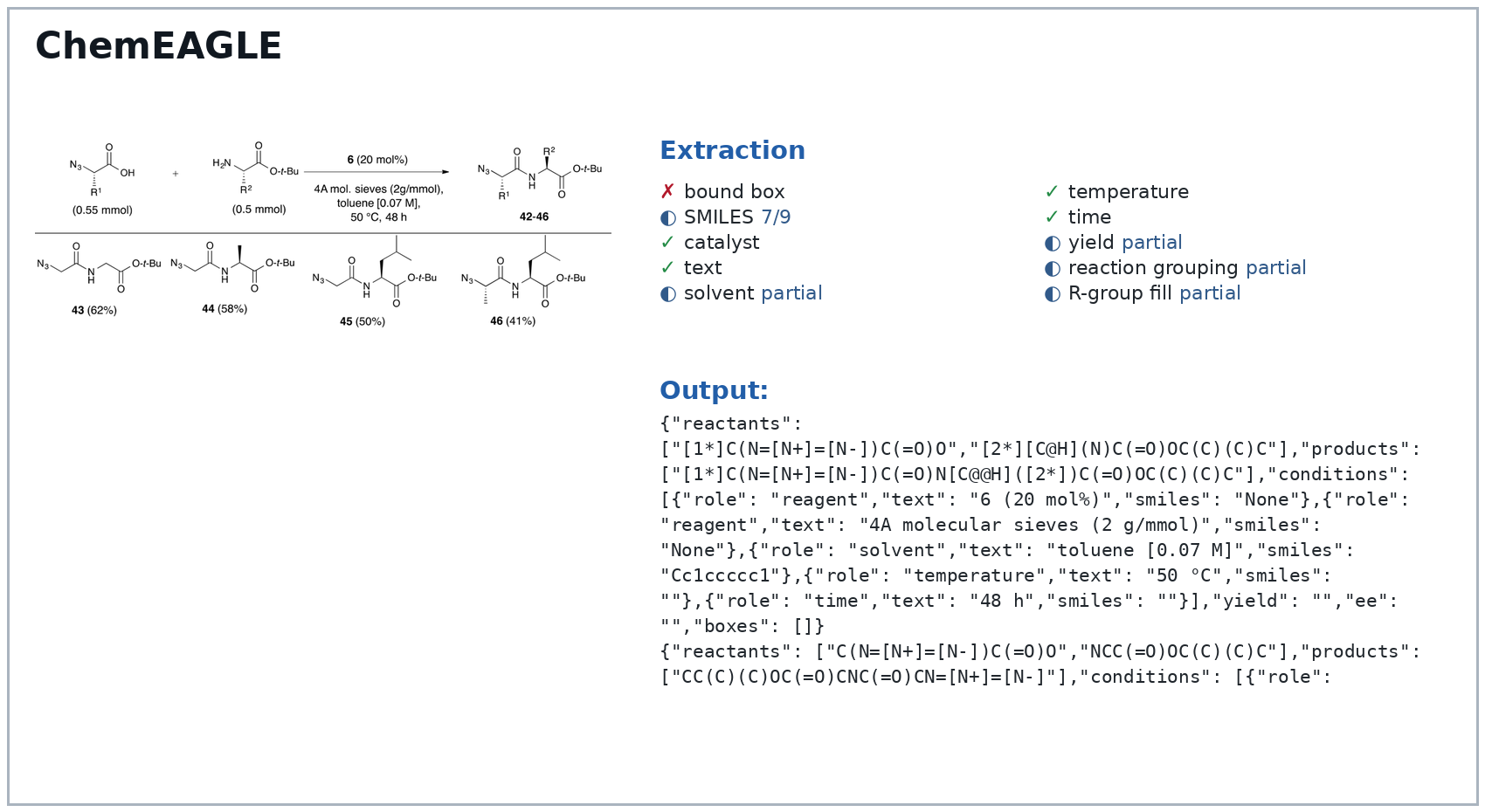}
  \caption{ChemEAGLE output for the selected case.}
  \label{fig:case-level-extraction-7}
\end{figure}
\begin{figure}[p]
  \centering
  \includegraphics[width=\linewidth]{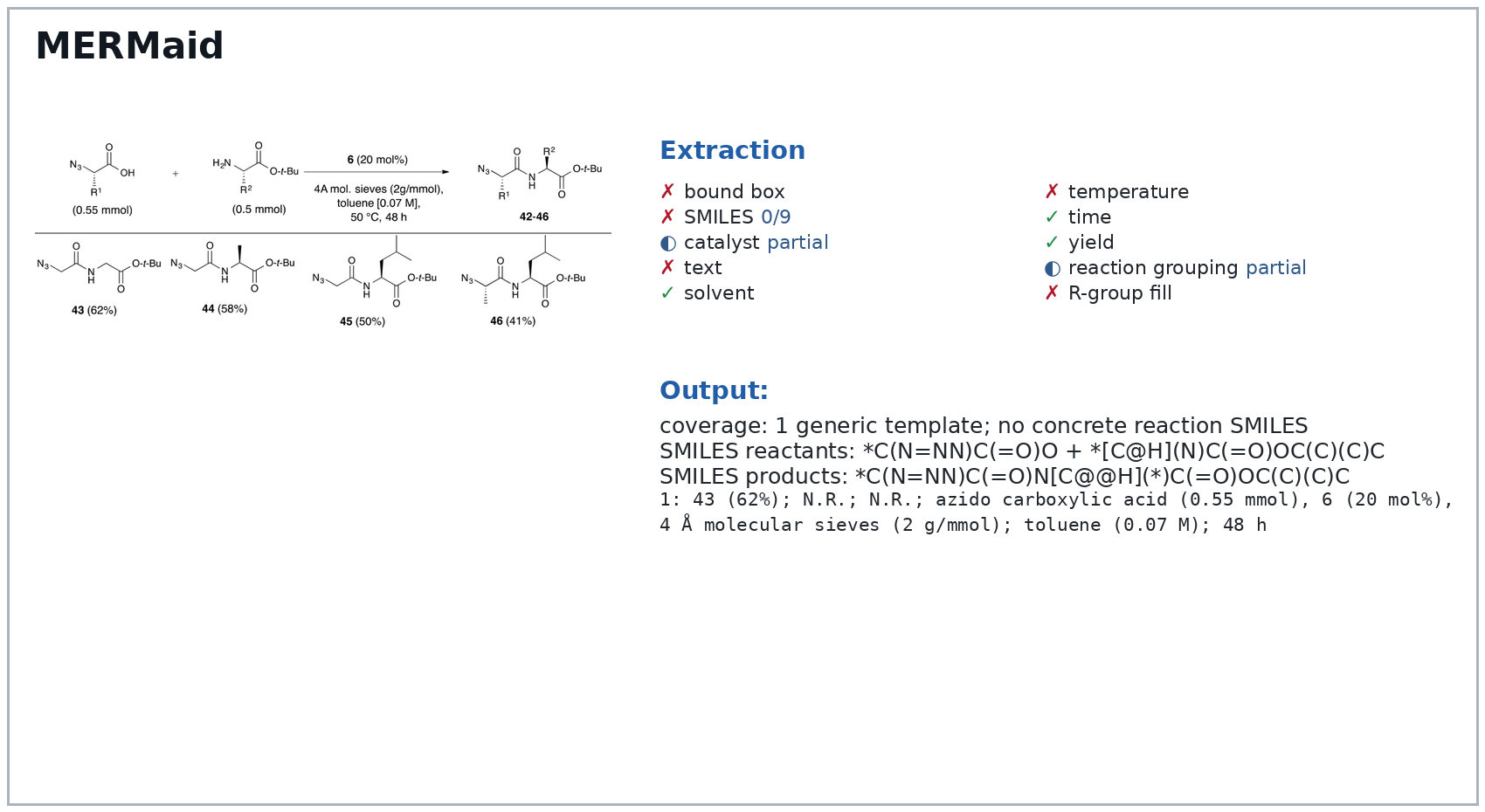}
  \caption{MERMaid output for the selected case.}
  \label{fig:case-level-extraction-8}
\end{figure}
\begin{figure}[p]
  \centering
  \includegraphics[width=\linewidth]{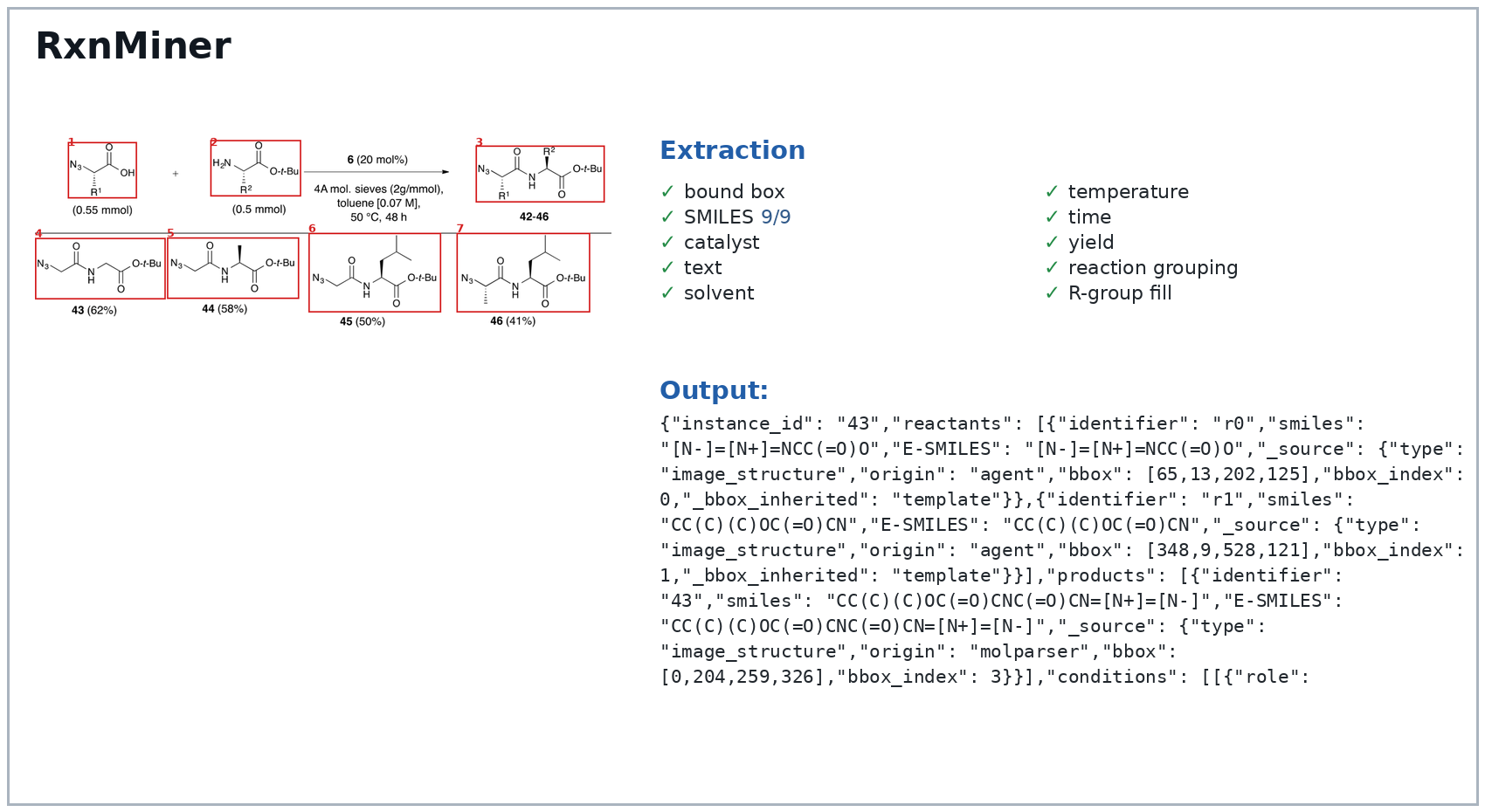}
  \caption{RxnMiner output for the selected case.}
  \label{fig:case-level-extraction-9}
\end{figure}

\paragraph{Unavailable baseline.}
RxnIM was not included in the figure because it was not run: the official
Hugging Face download returned a bad-CRC archive and could not be used.

\FloatBarrier
\endgroup
\end{document}